\documentclass[twoside]{article}

\usepackage{amsfonts}       % blackboard math symbols
\usepackage{subcaption}
\usepackage{multirow}
\usepackage{mathtools}
\usepackage{amsthm}
\usepackage{xspace}
\usepackage{enumitem}
\usepackage{doi}
\usepackage{algpseudocode}
\usepackage{orcidlink}
\usepackage[acronym]{glossaries}

\glsdisablehyper

\usepackage{siunitx}
\usepackage{tikz}

\usepackage{booktabs}
\usepackage{float} % in preamble

\newcommand{\Math}[1]{\ensuremath{#1}\xspace}

\newcommand{\ReLU}{\Math{\operatorname{ReLU}}}
\newcommand{\GELU}{\Math{\operatorname{GELU}}}
\newcommand{\Enc}{\Math{\operatorname{E}}}
\newcommand{\Dec}{\Math{\operatorname{D}}}

\newcommand{\LayerNorm}{\Math{\operatorname{LayerNorm}}}

\newacronym{ARC}{ARC}{AI2's Reasoning Challenge}
\newacronym{CE}{CE}{cross-entropy}
\newacronym{DL}{DL}{deep learning}
\newacronym{CNN}{CNN}{convolutional neural network}
\newacronym{GC}{GC}{garbled circuits}
\newacronym{HE}{HE}{homomorphic encryption}
\newacronym{FHE}{FHE}{fully homomorphic encryption}
\newacronym{KD}{KD}{knowledge distillation}
\newacronym{LLM}{LLM}{large language model}
\glsunset{LLM}
\newacronym{ML}{ML}{machine learning}
\newacronym{MPC}{MPC}{multi-party computation}
\newacronym{NAS}{NAS}{network architecture search}
\newacronym{NTP}{NTP}{next-token prediction}
\newacronym{PTA}{PTA}{post-training approximation}
\newacronym{AAT}{AAT}{approximation-aware training}
\newacronym{SDK}{SDK}{software development kit}
\newacronym{SC}{SC}{skip-connection}
\newacronym{OT}{OT}{oblivious transfer}
\newacronym{PPML}{PPML}{privacy-preserving machine learning}
\newacronym{ResNet}{ResNet}{residual net}
\newacronym{SIMD}{SIMD}{single instruction multiple data}
\newacronym{SS}{SS}{Shared Secret}
\newacronym{SOTA}{SOTA}{State-of-the-Art}
\newacronym{FFN}{FFN}{Feed-Forward Network}

\usepackage{amsmath,amsfonts,bm}

\def\eqref#1{equation~\ref{#1}}
\def\1{\bm{1}}

\DeclareMathAlphabet{\mathsfit}{\encodingdefault}{\sfdefault}{m}{sl}
\SetMathAlphabet{\mathsfit}{bold}{\encodingdefault}{\sfdefault}{bx}{n}

\newcommand{\R}{\mathbb{R}}

\newcommand{\Softmax}{\Math{\operatorname{Softmax}}}
\newcommand{\PowerSoftmax}{\Math{\operatorname{PowerSoftmax}}}

\newcommand{\norm}[1]{\left\lVert #1 \right\rVert}
\usepackage{hyperref}   
\usepackage{amsmath}
\usepackage[ruled,vlined]{algorithm2e}
\usepackage{wrapfig} 
\usepackage{subcaption}
\usepackage[accepted]{aistats2026}

\usepackage[round]{natbib}

\begin{document}



\twocolumn[

\aistatstitle{PowerSoftmax: Towards Secure LLM Inference over Encrypted Data}
% \runningauthor{Itamar Zimerman \And Allon Adir \And Ehud Aharoni \And Matan Avitan \And Moran Baruch \AND Nir Drucker \And Jenny Lerner \And Ramy Masalha \And Reut Moshe \And Omri Soceanu}

% \aistatsauthor{Itamar Zimerman \And Allon Adir \And Ehud Aharoni \And Matan Avitan \And Moran Baruch}

% \aistatsauthor{Nir Drucker \And Jenny Lerner \And Ramy Masalha \And Reut Moshe \And Omri Soceanu\AND}
\vspace{-16pt}
\aistatsauthor{Itamar Zimerman\,\orcidlink{0000-0001-8321-0609} \And Allon Adir\,\orcidlink{0000-0001-8128-6706} \And Ehud Aharoni\,\orcidlink{0000-0002-3647-1440} \And Matan Avitan \orcidlink{0009-0001-0762-9616} \AND}
\vspace{-16pt}
\aistatsauthor{Moran Baruch\,\orcidlink{0000-0003-0615-6164} \And Nir Drucker\,\orcidlink{0000-0002-7273-4797} \And Jenny Lerner\,\orcidlink{0000-0002-3998-6917} \And Ramy Masalha\,\orcidlink{0000-0002-6808-5675} \AND}
\vspace{-16pt}
\aistatsauthor{Reut Meiri\orcidlink{0009-0003-9640-448X} \And Omri Soceanu\,\orcidlink{0000-0002-7570-4366}}

%\And Nir Drucker \And Jenny Lerner \And Ramy Masalha \And Reut Moshe \And Omri Soceanu}
\aistatsaddress{IBM Research} ]
\runningauthor{Zimerman et al.}

\begin{abstract}
Modern cryptographic methods for implementing privacy-preserving LLMs such as \gls{HE} require the LLMs to have a polynomial form. Forming such a representation is challenging because transformers include non-polynomial components, such as \Softmax and layer normalization. Previous approaches have either directly approximated pre-trained models with large-degree polynomials, which are less efficient over HE, or replaced non-polynomial components with easier-to-approximate primitives before training, e.g., \Softmax with pointwise attention. The latter approach might introduce scalability challenges. We present a new HE-friendly variant of self-attention that offers a stable form for training and is easy to approximate with polynomials for secure inference. Our work introduces the first polynomial LLMs %with 32 layers and
over a billion parameters, exceeding the size of previous models by more than tenfold. The resulting models demonstrate reasoning and in-context learning (ICL) capabilities comparable to standard transformers of the same size, representing a breakthrough in the field. Finally, we provide a detailed latency breakdown for each computation over encrypted data, paving the way for further optimization, and explore the differences in inductive bias between models relying on our HE-friendly variant and standard transformers. %Our code is attached as a supplement.
\end{abstract}

\section{INTRODUCTION}

\Gls{PPML} solutions and in particular privacy-preserving \glspl{LLM} \citep{yan2024protecting,yao2024survey} aim to provide confidentiality guarantees for user data, the model owner, or both. One prominent cryptographic primitive for achieving this is \acrfull{HE}, as it allows computations to be performed on encrypted data without revealing any information to the (potentially untrusted) computing environment. Furthermore, HE enables non-interactive computations, which increases the usability of these solutions.

However, modern \gls{HE} schemes like CKKS~\citep{ckks2017} face a significant challenge of only supporting polynomial computations on encrypted data.
% when these computations have a polynomial form. 
This limitation complicates the deployment of DL models, particularly for \glspl{LLM}, which depend on non-polynomial functions like \Softmax in self-attention. To overcome this, existing approaches have adapted these non-polynomial operations into polynomial forms using techniques such as unique polynomial approximation~\citep{lee2021precise} or fine-tuning procedures~\citep{baruch2021fighting}.
While these methods have enabled the execution of FFNs, CNNs~\citep{lee2022low,baruch2023sensitive}, and small transformers~\citep{zimermanconverting} over \gls{HE}, they often struggle with stability and sensitivity issues~\citep{%zhou2019polynomial,
goyal2020improved}, preventing an effective scale-up.

We take a different approach. Rather than modifying existing transformers to fit within the constraints of \gls{HE}, we revisit the core design principles of the transformer architecture~\citep{c:22} through the lens of the CKKS constraints. Concretely, we ask:

{\centering\textit{Are there HE-friendly operators that can replicate the key design principles of self-attention?}\par}

We find a positive answer by introducing a power-based variant of self-attention that is more amenable to polynomial representation. Models with this variant maintain comparable performance to \Softmax-based transformers across several benchmarks and preserve the core design characteristics of self-attention. We also present variants that include length-agnostic approximations or improved numerical stability. The entire mechanism offers a more HE-friendly and effective transformer solution than previous approaches, enabling our method to scale efficiently to \glspl{LLM} with 32 layers and 1.4 billion parameters.

    \textbf{Our main contributions:} (i) We propose a \gls{HE}-friendly self-attention variant tailored specifically for \gls{HE} environments. This variant minimizes the usage of non-polynomial operations while maintaining the core principles of attention mechanisms. Additionally, we extend this approach by introducing a numerically stable training method and a length-agnostic computation strategy for inference. As a result, our model enables secure inference at scale and is more efficient than existing methods. (ii) We leverage this technique to develop the first polynomial \gls{LLM} that exhibits reasoning and ICL capabilities, as well as the largest polynomial model trained to date, encompassing 32 transformer layers and approximately a billion parameters, and a polynomial variant of RoBERTa (iii) We validate that our model operates precisely over \gls{HE}, % \textcolor{red}{[Ref section]}, advancing the frontend of this technology in the context of \glspl{LLM}. We also
    and also provide ablation studies and profiling of latency breakdowns over encrypted data, paving the way for further improvements. 

\vspace{-3pt}
\section{BACKGROUND}
\vspace{-2pt}

{\noindent \textbf{Homomorphic Encryption\quad}}%
HE is a form of encryption that enables processing of encrypted data without decrypting it \citep{Gentry2009}, so that the results achieved from processing data inputs while encrypted are similar to the results of applying the same computation after decryption. 
Some \gls{HE} schemes \citep{bgv,bfv1} are exact, meaning that the value of the decrypted ciphertext is exactly the result of the arithmetic operation, while some like CKKS \citep{ckks2017} are approximate and introduce a tiny amount of noise ($\eta$) to the decrypted values. Formally, an \gls{HE} scheme encryption operation $\Enc:\R_1 \rightarrow \R_2$ takes a plaintext from a ring $\R_1(+, *)$ and transforms it into a ciphertext in a ring $\R_2(\oplus, \odot)$ (and the opposite holds for decryption $\Dec:\R_2 \rightarrow \R_1$). This all occurs while also maintaining the following properties for an input $x,y \in \R_1$: (i) $\Dec(\Enc(x)) = x + \eta$, (ii) $\Dec(\Enc(x) \oplus \Enc(y)) = x + y + \eta$, and (iii) $\Dec(\Enc(x) \odot \Enc(y))  = x * y + \eta$.

\noindent \textbf{Polynomial DL Models\quad}
Deep learning models rely heavily on non-polynomial activation functions like \ReLU, sigmoid, and tanh to introduce non-linearity, which enhances model expressiveness. However, over most \gls{HE} schemes, operations must have a polynomial form.
Prior work has reported that polynomial DNNs tend to face instability as the network grows (\citet{zhou2019polynomial, goyal2020improved, chrysos2020p, gottemukkula2020polynomial}). Thus, maintaining an accurate and stable network when using polynomial approximations is challenging.

There are two primary approaches for polynomial approximation: \gls{PTA} and \gls{AAT}.
In \gls{PTA}, the approximation is applied to a pre-trained network without modifying the model architecture and parameters (\citep{lee2021precise,ao2023autofhe,ju2023neujeans, zhang2024secure,avitan2025efficient}). This approach saves the costly training process by providing a precise approximation for each computation using high-degree polynomials. 

In contrast, \gls{AAT} aims to reduce the number of required approximation polynomials in the network or to minimize their degree \citep{gilad2016cryptonets, lee2023optimizing, baruch2021fighting, baruch2023sensitive, ao2023autofhe, drucker2023efficient,zimermanconverting}. Doing so can improve both latency and precision under HE, as higher-degree polynomials increase the \textit{multiplicative depth}--the number of sequential multiplications required--leading to higher computational overhead, greater resource consumption, and an increase in the accumulated noise. 
Typically, this is achieved by modifying the network architecture. For instance, early studies substituted the \ReLU activation function with quadratic activations~\citep{gilad2016cryptonets,baruch2021fighting}. %, or replaced \LayerNorm with \BatchNorm \textcolor{red}{\citep{todo}}, which does not require approximation.

To reduce polynomials' degrees in large %-scale
models, such as ResNet152 on ImageNet and transformers, while still achieving accurate approximation, recent works (\citep{baruch2023sensitive} and \citep{zimermanconverting}) have suggested using the training process to minimize the input range to the non-polynomial layers. This is done by adding a \textbf{range-loss term} to the original loss% function
, encouraging the model to operate within a range where lower-degree %polynomial
approximations are accurate enough.

% \noindent\textbf{Polynomial Transformers}\quad
% To enjoy the non-interactive property of \gls{HE}-based solutions, this paper only considers fully polynomial models. While other secure alternatives such as \citep{TransformerPPML1, TransformerPPML3,TransformerPPML5, TransformerPPML6, TransformerPPML7} exist, they require interaction with the user to process non-polynomial layers. This involves extra communication overhead and may be susceptible to some cryptographic attacks~\citep{AkaviaVald21}. In contrast, the use of \gls{HE} enables non-interactive computation in untrusted environments without additional communication. %
% %
% In transformers, the \Softmax, \LayerNorm, and \GELU are non-polynomial operations that need to be replaced or approximated.  

\noindent\textbf{Polynomial Transformers}\quad
To enjoy the non-interactive property of \gls{HE}-based solutions, this paper only considers fully polynomial models. While other secure alternatives~\citep{TransformerPPML1, TransformerPPML3,TransformerPPML5, TransformerPPML6, TransformerPPML7} exist, they require interaction with the user to process non-polynomial layers, incurring extra communication overhead and potential vulnerability to cryptographic attacks~\citep{AkaviaVald21}. In contrast, \gls{HE} enables non-interactive computation in untrusted environments without additional communication.  To this end, the non-polynomial operations in transformers, namely \Softmax, \LayerNorm, and \GELU, need to be replaced or approximated.

The first work to present a fully polynomial transformer was by \citep{zimermanconverting}, who used the \gls{AAT} approach and substituted \Softmax with a $\operatorname{scaled-\ReLU}$ that is easier to approximate. They also used the range-loss term during training to reduce the polynomial degree required for accurate approximation of \ReLU and \LayerNorm. They demonstrated a 100M-parameter polynomial transformer pretrained on WikiText-103 for secure classification % tasks
using \gls{HE}. 

Alternatively, ~\citep{zhang2024secure} used the PTA approach. They introduced a polynomial transformer by directly approximating the numerator, denominator, and division separately, without dedicated training modifications. However, as described in Sec. {\ref{subsec:ComparisonSota}, this approach has disadvantages in terms of latency.% and scalability.

In this work, we scale up the AAT for transformers approach, by replacing \Softmax with a polynomial-friendly alternative that closely replicates its behavior. This enhancement allows us to improve model performance and scalability, enabling the deployment of $1.4$B-parameters LLMs under HE, while maintaining the model's performance. After training, we approximate the non-polynomial operations using methods detailed in App.~\ref{app:approximations}, converting the trained model into a polynomial form for secure inference.

\vspace{-6pt}
\section{PROBLEM SETTING\label{sec:problem}}
\vspace{-3pt}
We consider secure inference for \glspl{LLM} over \gls{HE}, where a semi-honest server runs inference on behalf of a data-owner whose query remains encrypted throughout. Optionally, model weights may also be encrypted, depending on the trust assumptions between the server and the model owner; our approach supports both cases transparently (see App.~\ref{app:threat} for details).

The core technical goal is a transformer that relies \emph{exclusively} on polynomial operations, enabling direct \gls{HE} deployment, while matching the performance of standard transformers at billion-parameter scale. This is challenging on two fronts: polynomial networks are prone to instability even at modest scales~\citep{zhou2019polynomial,goyal2020improved,zhang2024neural}, and \gls{HE} computational cost grows with polynomial degree, making high-degree solutions impractical. These pressures pull in opposite directions, as expressivity demands higher-degree polynomials while efficiency demands lower ones, and navigating this tension is the central challenge we address.

\vspace{-6pt}
\section{METHOD\label{sec:method}}
\vspace{-2pt}
The self-attention mechanism %in transformers
 is defined by: %
{\small
\begin{equation}\label{eq:attention}
\operatorname{Self-Attention}(Q, K, V) = \Softmax\left(\frac{QK^T}{\sqrt{d_k}}\right) V\,,
\end{equation}%
}%
which is inherently non-polynomial because it includes division and exponential operations. Furthermore, for numerical stability it is common to compute the \Softmax function using the \textit{log-sum-exp} trick,
which adds non-polynomial operations. For example, it involves calculating the maximum absolute values of each row of $QK^T$. The latter operation involves high-degree polynomials that in \gls{HE} environments may introduce significant noise. Instead of directly approximating the maximum, division, and exponential functions individually (as done in  the Nexus protocol ~\citep{zhang2024secure}), our objective is to develop a more polynomial-friendly and HE-compatible \Softmax variant for transformers. Such a mechanism not only reduces the overall computational complexity, particularly in terms of multiplication depth, but also  
supports scaling polynomial transformers to models with billions of parameters and deeper architectures. 

\vspace{-6pt}
\subsection{HE-Friendly Attention}
\vspace{-2pt}
To design an HE-friendly variant of \Softmax-based attention, we start by distilling its properties that correlate with its performance: (i) Normalization of the attention scores ensures they are bounded in $[0,1]$, with their sum equal to 1, similar to probabilities; (ii) exponential scaling of attention scores, such that it amplifies the differences between higher and lower scores; and (iii) monotonic increasing and order-preserving behavior, meaning that higher input values yield higher output values, while preserving the relative order of the input values.
Building on these properties, we introduce the following attention variant:
\vspace{-4pt}
{\small
\begin{equation}\nonumber
\operatorname{HE-Friendly~Attn}(Q, K, V) = \PowerSoftmax\left(\frac{QK^T}{\sqrt{d_k}}\right) V\, 
\end{equation}
\vspace{-3pt}
\begin{equation}\label{eq:PolyAttention}
\PowerSoftmax (x)_j{=}\frac{{x_j}^p}{\sum_i x_i^p}\,, 
\end{equation}
\vspace{-1pt}
}
where we replaced the $\Softmax(x)_j = e^{x_j} / \sum_i e^{x_i}$ function with $\PowerSoftmax$, for some positive even $p$.
Equation~\ref{eq:PolyAttention} describes a variant that satisfies (i), but does not  accurately retain properties (ii) and (iii), as the variant performs \textit{polynomial scaling} instead of \textit{exponential scaling} (both have superlinear trends), and because it is not strictly monotonic increasing. Nevertheless, for suitable values of $p$, the polynomial scaling can mimic the trends of exponential scaling relatively well, as shown in Fig.~\ref{fig:motivationPowerSoftmax}. Additionally, instead of maintaining the order and strictly increasing monotonically, our variant preserves \textit{the order of the norms} and is increasing monotonically for positive values.

\begin{figure*}[t!]
\centering
\includegraphics[width=0.96\linewidth]{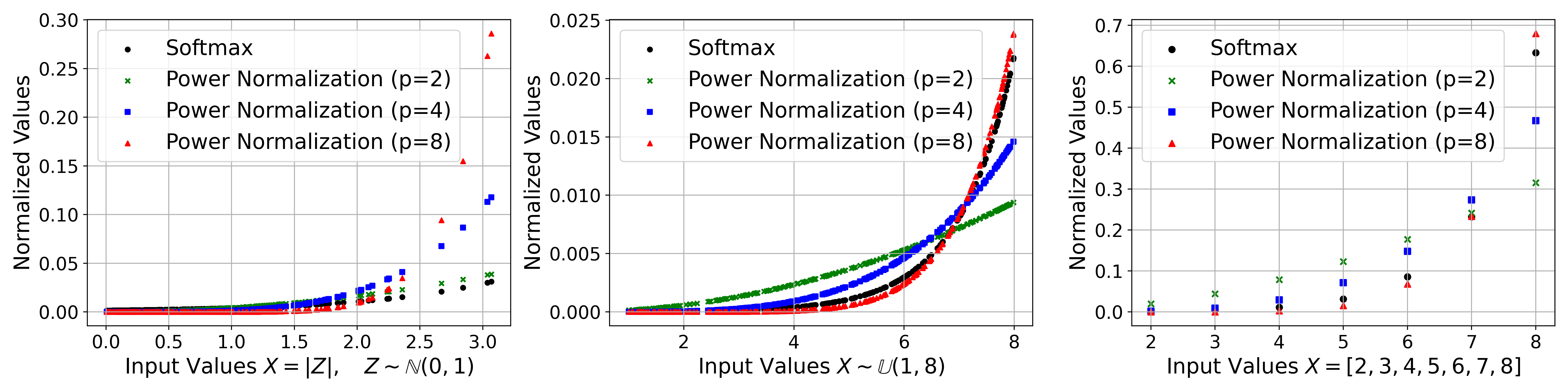}
\caption{\small\textbf{Comparison of \Softmax and \PowerSoftmax} on normally distributed values on the left, uniformly distributed values in the middle, and evenly spaced values on the right. As can be seen, the %%empirical
     scaling trends are relatively similar.}
\label{fig:motivationPowerSoftmax}
\end{figure*}

To highlight the similarities and differences between both attention mechanisms in Eqs.~\ref{eq:attention} and ~\ref{eq:PolyAttention}, we introduce a generalization of the \Softmax function within transformers, using an elementwise activation function $\sigma : \mathbb{R} \rightarrow \mathbb{R}$ followed by proportional normalization $\mathbb{N} : \mathbb{R}^L \rightarrow \mathbb{R}^L$:

{\small
\begin{equation}\label{eq:GenerelizedAttention}
\operatorname{Generalized~Self-Attn}(Q, K, V) = \mathbb{N} \left ( \sigma \left (\frac{QK^T}{\sqrt{d_k}}\right) \right) V\, 
\end{equation}
\begin{equation}
\mathbb{N}(\mathbf{x})_j = \frac{|\mathbf{x}_j|}{\norm{\mathbf{x}}_1}\,.
\end{equation}
}
In this formulation, \Softmax is obtained by setting $\sigma$ as
$\sigma_e(x)=\exp(x)$, while our variant is defined by using $\sigma_p(x)=x^p$ for $\sigma$ using a positive even $p$. {\color{black}Additionally, standard LLM practices often involve using attention masks. Thus, %for completeness,
a masked HE-friendly variant is included in App.~\ref{sec:attnMask}.

\subsection[1/epsilon2]{\texorpdfstring{$\frac{1}{{\epsilon}^2}$}{1/epsilon^2}-Lipschitz Division for \Softmax Approximation\label{subsec:lipshizVariant}}
A key challenge in approximating \Softmax or Eq.~\ref{eq:PolyAttention} with polynomials is the behavior of the inverse term $ 1/x $, which grows rapidly near zero, i.e., $ \lim_{x \to 0^+} \frac{1}{x} = \infty $. While \Softmax %($\sigma(x)=\exp(x)$)
deals with summation over strictly positive exponents, this property does not hold for \PowerSoftmax, where the denominator can potentially reach zero. To address this, we propose the \textit{$\frac{1}{{\epsilon}^2}$-Lipschitz division for \PowerSoftmax}, modifying the denominator of $\mathbb{N}$ before training as:
{\small
\begin{equation}\label{eq:LipschitzPolyAttention} 
    \dfrac{1}{{\epsilon}^2}
    \operatorname{-Lipschitz~HE-Friendly~Attn}(Q,K,V)=
\end{equation}
\begin{equation}\nonumber
     \mathbb{N}_\epsilon\left(\sigma_p\left(
    \frac{QK^T}{\sqrt{d_k}}
    \right)\right)V, \texttt{   }
    \mathbb{N}_\epsilon(\mathbf{x})_j = \frac{|\mathbf{x}_j|}{\boldsymbol{\epsilon} + \norm{\mathbf{x}}_1}\,.
\end{equation}
}

Here, $ \epsilon 
$ (e.g., $1e-3$) ensures the denominator is bounded away from zero, preventing discontinuities and ensuring $ \lim_{x \to 0^{+}} \frac{1}{x + \epsilon} = \frac{1}{\epsilon} $. This introduces a single non-polynomial division, which is $\frac{1}{{\epsilon}^2}$-Lipschitz continuity function, making the polynomial approximation more tractable.
Importantly, unlike the common use of $\epsilon$ for numerical stability in division, our approach focuses on much larger values of $\epsilon$ to reduce the multiplication depth required for approximation, making the approximation problem significantly easier for secure inference over HE.

% {\noindent\textbf{Length-Agnostic}}
\subsection{Stable Variant for Training}

By examining the $i$-th row of the unnormalized attention scores $ S_i = \left[\frac{1}{ \sqrt{d_k}} {QK^T}\right]_i $, it is clear that Eq.~\ref{eq:PolyAttention} can lead to training instability when applying \PowerSoftmax, as when $ |S_{i,j} | > 1$, ${|S_{i,j}|}^p $ can become very large, causing overflow, and when $|S_{i,j}| < 1$, $|S_{i,j}|^p$ can become very small, leading to underflow. In transformers, a similar problem occurs with the traditional \Softmax, which is mitigated using the \textit{log-sum-exp trick} to scale the values of $|S_{i}|$ within a manageable range. Inspired by this, we propose a more stable version of our \PowerSoftmax variant: %
{\small
\begin{equation}\label{eq:stableeVariant}
\operatorname{Stable~PowerSoftmax}(\mathbf{x})_j := \operatorname{PowerSoftmax}\left(\frac{\mathbf{x}}{c}\right)_j\,,
\end{equation}
}
where $c = \norm{\mathbf{x}}_\infty + \delta$ and $\delta$ is a small positive number introduced to avoid division by 0. This method leverages the fact that \PowerSoftmax is invariant to division of its input by a constant $c > 0$ (similar to \Softmax which is invariant under the subtraction% of a constant
).
By selecting $c$ such that $\forall j |S_{i,j}| < 1$, we (i) ensure that the input values stay within a range where floating-point precision is more reliable $(0 < |S_{i,j}| < 1)$, and (ii) stretch (or shrink) the values of $x$ to have a similar scale across different coordinates, preventing the loss of significant digits during division. Fig.~\ref{fig:MainFig} (middle) illustrates our HE-friendly training variant, built on top of Eqs.~\ref{eq:LipschitzPolyAttention} and~\ref{eq:stableeVariant}, compared to the original attention.

\begin{figure*}[t]
    \centering
    \includegraphics[width=0.99\linewidth]{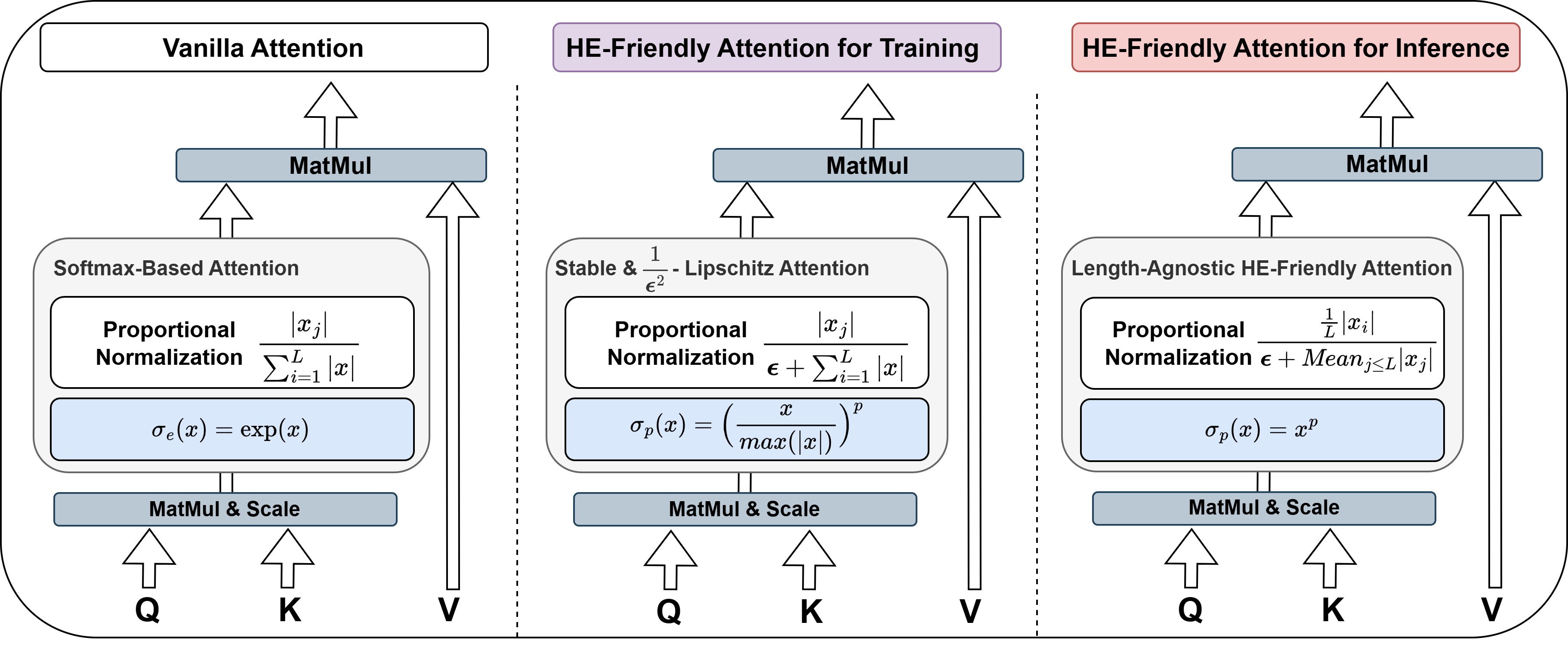}
    \caption{\small\textbf{Our Attention Variants:} (Left) the \Softmax-based attention mechanism using the generalized attention formulation (Eq.~\ref{eq:GenerelizedAttention}). (Middle) Our variant for training (purple),  builds on the stable variant from Eq.~\ref{eq:stableeVariant} and the Lipschitz division from Eq.~\ref{eq:LipschitzPolyAttention}. (Right) During secure inference with the polynomial model (red), we use a length-agnostic approximation for division, as described in Eq.~\ref{eq:LengthAgnosticAttention}.}
    \label{fig:MainFig}
\end{figure*}
 
\subsection{Length-Agnostic Range for Polynomial Division}
\label{sec:LAP}
The only non-polynomial operation in Eq.~\ref{eq:PolyAttention} is division, which can be approximated effectively in a bounded domain using the Goldschmidt algorithm \citep{goldschmidt}. However, in our attention variant, we need to approximate the function $\frac{1}{x}$, where $x$ is the sum of the scores raised to the power of $p$,
%$\left(\frac{1}{\sqrt{d_k}}QK^T\right)_j^p$, 
which is unbounded and increases linearly with the sequence length $L$. Thus, applying Goldschmidt's algorithm naively would struggle to precisely approximate division for both short and long sentences and would require relatively high-degree polynomials due to the extremely large domain range. To address this problem, we propose a length-agnostic HE-friendly attention:% variant: %
%
%
% \begin{equation}\label{eq:LengthAgnosticAttention}
%     \operatorname{Length-Agnostic~HE-Friendly~Attention}(Q,K,V) = \frac{\frac{1}{L}\left( \frac{1}{\sqrt{d_k}}QK^T\right)^p}{\textbf{Mean}_{j\leq L} \left(\frac{1}{\sqrt{d_k}}QK^T\right)_j^p}V
% \end{equation}
{\small
\begin{equation}\label{eq:LengthAgnosticAttention}
    \operatorname{Length-Agnostic~PowerSoftmax}(\mathbf{x})_j = 
\end{equation}
\begin{equation}\nonumber
    \frac{\frac{1}{L}x_j^p}{\operatorname{Mean}_{i\leq L} x_i^p}  = \frac{\Big{(}\frac{x_j}{L'}\Big{)}^p}{\operatorname{Mean}_{i\leq L} x_i^p}\,.
\end{equation}
}
This variant leverages the fact that the sequence length $L$ is not a secret, so $\frac{1}{L}$ can be computed directly without approximation (or pre-computed by the client). The resulting approximation operates on the mean of the attention scores rather than their sum. Notably, if the attention scores have mean $\mu$ and variance $\sigma^2$, the asymptotic behavior of both approaches as $L \to \infty$ can be described as follows (by the law of large numbers):
{\small
\begin{equation}
    \operatorname{Mean} \sigma_p\left(\frac{1}{\sqrt{d_k}}QK^T\right) \rightarrow \mu\, \quad 
    \sum \sigma_p \left(\frac{1}{\sqrt{d_k}}QK^T\right) \rightarrow \infty\,.
\end{equation}
}
This shows that our length-agnostic variant does not become harder to approximate as L increases, enabling a more flexible and precise polynomial approximation. Fig.~\ref{fig:MainFig} (right) compares this variant with the original attention.

\subsection{A Recipe for Polynomial LLM\label{subsec:recipe}}

Alg.~\ref{alg:alg} illustrates the entire process, which is divided into three key stages: \textbf{(i)  Architectural modification:} We begin by modifying the original transformer architecture to use an HE-friendly attention variant (Eq.~\ref{eq:stableeVariant}). This modified model is then trained from scratch using the same hyperparameters as the vanilla transformer. \textbf{(ii) Range minimization:} In the second stage, we apply a supplementary training procedure as outlined in~\citep{baruch2023sensitive} to ensure that the model operates within HE-friendly constraints. Specifically, we adjust the model's weights so that each non-polynomial component operates only within specific, restricted input domains. This is achieved by adding a regularization loss function that minimizes the range of inputs to non-polynomial layers. For activations and \LayerNorm layers, we directly apply the method from~\citep{zimermanconverting}.%, characterized by the following loss function:
    % \begin{equation}
    %     TBD \textbf{copy from previous paper and adapt the notations), perhaps move to appendix}
    % \end{equation}
    
Additionally, for the HE-friendly attention mechanism, we introduce a tailored loss term defined as:%
{
\begin{equation}
\label{eq:PWRoss}
    \mathbb{L}_{\PowerSoftmax} := \sum_{n=1}^{N_L} \max_{c \in C}%, x_i \in X} %
        \left\{ {{|z|}_{n,c}^{i}}\right\}\,,
\end{equation}
}
where we denote the number of attention layers by $N_L$ and the set of heads by $C$. Additionally, we denote the input at layer $n$ to the \PowerSoftmax layer, at head $c \in C$, when the model processes the $x_i$ example by $z_{n,c}^{i}$.
This loss serves two main purposes: First, it minimizes the upper bound of the denominator in the HE-friendly attention, making the approximation problem more tractable. Second, we observed that when the input norm to the HE-friendly attention is not too high, the stabilize factor defined in Eq.~\ref{eq:stableeVariant} can be omitted, eliminating the need for additional division approximations. \textbf{(iii) Polynomial replacement:} In the final stage, each non-polynomial layer is replaced with its polynomial approximation, resulting in a fully polynomial model. App.~\ref{app:approximations} provides further details on the %polynomial
approximations used. These approximations are designed to be highly accurate for the HE-friendly weights obtained from the previous stages.

% \smallskip
\noindent \textbf{Continual Training\quad} A significant limitation of Step 1 in Algorithm~\ref{alg:alg}, compared to \gls{PTA} methods, is the need for retraining, which can be expensive for large transformers trained on extensive datasets. To mitigate this, we propose a complementary procedure to convert standard pre-trained attention layers into \PowerSoftmax layers via a short fine-tuning step. Since both attention variants share the same trainable parameters and perform similar (though not identical) computations (as shown in Fig.~\ref{fig:motivationPowerSoftmax}), we initialize the weights of our attention variant from a vanilla pre-trained reference model. Fine-tuning the resulting model reduces the performance gap between the two variants, enabling us to take advantage of the significant computational investment made in these models.

\noindent \textbf{Wrap-up\quad} Finally, we present a unified solution that refines the attention mechanism for FHE constraints. It adopts distinct forms for training and secure inference to tackle two main challenges: ensuring stable large-scale training (via a stable variant with non-polynomial division) and enabling efficient secure inference (through Lipschitz and length-agnostic variants) by minimizing multiplication depth. Furthermore, we demonstrate how this mechanism can be continually pre-trained, thus overcoming limitations of the \gls{AAT} approach and making it truly scalable.

%%%%%%%%%%%%%%%%% TBDDDDD
{
{\small
\begin{algorithm}[t]
\SetAlgoLined
\KwIn{A vanilla transformer architecture and hyper-parameters for training.}
\KwOut{A polynomial transformer for secure inference.}
1. \textbf{Architectural modification and pre-training}: 
Modify the transformer architecture via Eqs.~\ref{eq:stableeVariant} and~\ref{eq:LipschitzPolyAttention} (stable and Lipschitz HE-friendly variant) and train the new architecture from scratch with the same hyper-parameters.\\
2. \textbf{Range minimization}: Minimize the input range to the \GELU, \LayerNorm, and \PowerSoftmax layers via the loss function defined in Eq.~\ref{eq:PWRoss}.\\
3. \textbf{Polynomial replacement}: Replace the inverse function in HE-friendly attention and the inverse square root in \LayerNorm with polynomial approximations obtained from the Goldschmidt method. Replace activations with suitable polynomial approximations (see App.~\ref{app:approximations}). Incorporate the length-agnostic approximation
(Eq.~\ref{eq:LengthAgnosticAttention}).\\
% 4. \textbf{Polynomial Evaluation}: Final evaluation using the length-agnostic approximation strategy.\\
\caption{Polynomial Transformer Construction\label{alg:alg}}
\end{algorithm}
}
}

\begin{table*}[t]
    \centering
    \small
    \caption{\small Comparison of zero-shot and five-shot results 
    %for various datasets 
    between vanilla transformer and our %polynomial-compatible 
    poly. variant across different model sizes. Original models trained on Pile%~\citep{gao2020pile}
    . Results of non-polynomial models copied from~\citep{biderman2023pythia}.
    \smallskip
    \label{tab:pile}}
    \begin{tabular}{lrrrrrrrr}
        \toprule
        & \multicolumn{4}{c}{\textbf{Zero-shot}} & \multicolumn{4}{c}{\textbf{5-shot}} \\
        \cmidrule(lr){2-5} \cmidrule(lr){6-9}
        Dataset & \multicolumn{2}{c}{1.4B} &  \multicolumn{2}{c}{70M} & \multicolumn{2}{c}{1.4B} & \multicolumn{2}{c}{70M} \\
        \cmidrule(lr){2-3} \cmidrule(lr){4-5} \cmidrule(lr){6-7} \cmidrule(lr){8-9}
        %& Org & Poly & Org & Poly & Org & Poly & Org & Poly & Org & Poly & Org & Poly \\
        & Orig. & Poly.  & Orig. & Poly. & Orig. & Poly. & Orig. & Poly. \\
        \midrule
      
        Lambada O. Acc  & 0.610 & 0.607 & 0.192 & 0.258 & 0.568 & 0.487 & 0.134 & 0.181 \\
        % ($\downarrow$) Lambada O. PPL & 6.107 & 6.351 &  138.450 & 79.380 & 7.606 & 10.800 & 304.272 & 137.326 \\
        PIQA  & 0.720 & 0.710 & 0.598 & 0.592 & 0.725 & 0.720 & 0.582 & 0.597 \\
         WinoGrande  & 0.566 & 0.562 &  0.492 & 0.503 & 0.570 & 0.568 & 0.499 & 0.505 \\
         WSC  & 0.442 & 0.395 &  0.365 & 0.365 & 0.365 & 0.548 & 0.365 & 0.452 \\
         ARC-Easy  & 0.617 & 0.602 &  0.385  & 0.420 & 0.633 & 0.613 & 0.383 & 0.387 \\
         ARC-Challenge  & 0.272 & 0.265 & 0.162 & 0.185 & 0.276 & 0.277 & 0.178 & 0.183 \\
         SciQ  & 0.865 & 0.873 &  0.606 & 0.716 & 0.926 & 0.907 & 0.598 & 0.718 \\
         LogiQA  & 0.221 & 0.217 & 0.235 & 0.210 & 0.230 & 0.222 & 0.250 & 0.238 \\ %\hline
      \bottomrule
        %\hline
        
    \end{tabular}
\end{table*}

\section{EXPERIMENTS}

We now present an empirical evaluation of our method. Sect.~\ref{subsec:polyLLMs} introduces our polynomial \glspl{LLM} and reports results on both encrypted and unencrypted data in zero-shot and fine-tuned settings. Sect.~\ref{subsec:ablations} offers a comprehensive set of ablation studies, providing empirical justifications for the key design decisions of our method, and Sect.~\ref{subsec:ComparisonSota} presents comparisons of our method and other SoTA methods in the domain. Finally, Sect.~\ref{subsec:AttnMats} compares the attention matrices generated by the standard \Softmax with those produced by our HE-friendly variant, analyzing the differences between these matrices. The experimental setup is detailed in App.~\ref{sec:experimental}.

\subsection{Polynomial LLMs\label{subsec:polyLLMs}}

We experimented with polynomial variants of a causal transformer (GPT) and a bidirectional model.

\noindent\textbf{Causal Transformer\quad} For a GPT model, we built upon the Pythia~\citep{biderman2023pythia} family of models, adapting their training procedures, evaluation methodologies, and hyperparameters. Specifically, we trained two models for \gls{NTP} on the Pile% %dataset
~\citep{gao2020pile}: a small model with 70M %parameters
% ), which was trained from scratch using Algorithm~\ref{alg:alg}, 
%(detailed in Sect.~\ref{sec:method}), 
and a large model with 1.4B parameters, using continual pretraining (Sect.~\ref{subsec:recipe}).

We evaluated these models using the popular lm-evaluation-harness framework. %\citep{eval-harness}.
Table
~\ref{tab:pile} shows that our models achieve performance comparable to the original models for 5-shot and 0-shot settings. %Full configuration appears in App.~\ref{app:pythia_params}.
These results mark a significant advancement, as no prior work has introduced polynomial LLMs with demonstrated \textbf{ICL or reasoning capabilities}. This is particularly evident on reasoning benchmarks such as the ARC%,
%PIQA dataset shows a difference of $\sim{1}\%$ between the original and the polynomial models.
%\gls{ARC}%~\citep{clark2018think}
, where our models perform competitively.

\noindent\textbf{Bidirectional Transformer\quad} 
For the bidirectional model, we tested our approach on RoBERTa~\citep{liu2019roberta}. Starting with a \Softmax-based pre-trained transformer, we applied the HE-friendly adaptation using the method described in Sect.~\ref{subsec:recipe} through continual pre-training on the OpenWebText corpus~\citep{Gokaslan2019OpenWeb}. Then, we fine-tuned our model on three datasets from the GLUE benchmark~\citep{wang2018glue} separately, adapting RoBERTa's fine-tuning process, and finally approximating the non-polynomial components. The results are depicted in Tab.~\ref{tab:glue}, and compared with the work of~\citet{zhang2024neural}. The full configuration is detailed in App.~\ref{app:roberta_params}. These results indicate a degradation of approximately 1\% compared to the original RoBERTa.

\begin{table}[H]
    \centering
    \small
    \caption{Downstream GLUE results for polynomial RoBERTa-Base. Results from \citep{zhang2024secure} are denoted by $^\diamond$.\label{tab:glue}}
    \smallskip
    \begin{tabular}{lrrr}
        \toprule
        \multicolumn{1}{l}{\multirow{2}{*}{Model}} & \multicolumn{3}{c}{Dataset} \\
        \cmidrule{2-4}
        & SST-2 & QNLI &  MNLI \\
        \midrule
        RoBERTa  & 94.80 & 92.80 & 87.60  \\
        Poly-RoBERTa   & 93.35 & 91.62 & 86.93   \\
        \midrule
        Nexus (BERT) $^\diamond$   & 92.11 & 89.90 & N.A   \\
        \bottomrule
    \end{tabular}
\end{table}

\noindent\textbf{Experiments over HE\quad} %

To demonstrate the feasibility of our method under FHE, we conducted an experiment designed to show that the accuracy of both encrypted and plaintext inference is similar. For this, we implemented polynomial Pythia 70M using HElayers 1.5.4 \citep{helayers}, and evaluated the model over 100 samples twice. We observed a maximal MSE loss of 0.005 between the encrypted and non-encrypted inference outputs, without modifying the maximal (predicted) value in 99\% of the cases. While the scale of the experiment was limited to Pythia 70M, we chose this model because the encrypted inference time takes 93 seconds per sample on a single A100 GPU. For full details of the workflow under HE, see App.~\ref{app:he-param}. Please note that we conducted experiments over 128 tokens, following previous work \citep{zhang2024secure,cho2024fast}. However, practical scenarios and LLMs often process much longer sequences with thousands of tokens. In these scenarios, the proportional cost of the \Softmax compared to other components drastically increases, as it is the only component whose complexity scales quadratically rather than linearly with sequence length. This rise in the proportional cost of \Softmax makes our method much more efficient for long-context% processing
, as other methods like \citep{zhang2024secure,cho2024fast} require computing the maximal value over each row of the attention matrix, resulting in a gap that increases with context length.

\subsection{Ablation Studies\label{subsec:ablations}}

We perform the following series of ablation studies:

{\noindent\textbf{PowerSoftmax Attention}\quad} %
We first compare \Softmax and \PowerSoftmax outside the context of \gls{HE}, showing that in addition to being an HE-friendly variant, \PowerSoftmax also exhibits similar scaling trends as \Softmax. Figures~\ref{fig:powerVssoftmaxrWikit} and \ref{fig:powerVsbaseVision} (in App.~\ref{app:additionalExpirments}) present comparative visualizations of training curves for various model sizes and datasets (including Pile, Wikitext-103, Text-8, Tiny-Imagenet, CIFAR-100 and CIFAR-10) across both NLP and vision domains, respectively. Although \Softmax generally achieves better results, it is evident that by the end of training, most of the gap between the models is reduced, and the scaling laws of the models are relatively similar.

\begin{figure*}[h]
    \centering
    \includegraphics[width=0.94\linewidth]{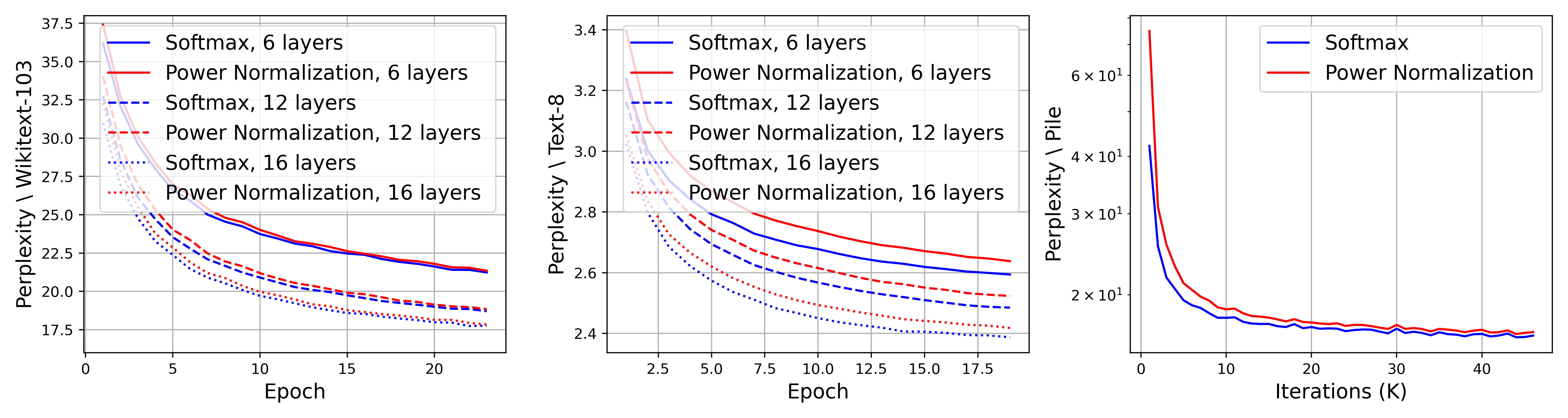}
    \caption{\small\textbf{Training curves for \gls{NTP}:} Comparison of test perplexity for transformers with \Softmax and \PowerSoftmax when trained over several datasets including Pile, Wikitext-103, and Text-8. %We used models with 6, 12 and 16 layers, and $p=8$ for power normalization.}
    }
    \label{fig:powerVssoftmaxrWikit}
\end{figure*}

\noindent\textbf{Stability}\quad %
To assess the contribution of our numerically stable variant, we conduct dedicated experiments. In Fig.~\ref{fig:stablityEmpirical}, we provide training curves for models with 32 layers and a hidden dimension size of 1024, trained on 10\% of the Wikitext-103 dataset. We compare two \PowerSoftmax-based transformers with the same training procedure, one with (blue) and one without (red), the stable variant from Eq.~\ref{eq:stableeVariant}. As an additional baseline, we trained a vanilla transformer (black). As shown, the stable variant consistently outperforms the \PowerSoftmax baseline, closing a third of the gap between the \PowerSoftmax and the \Softmax baseline. Additionally, we observe that in more challenging regimes, such as training on the full dataset or other datasets, the stable variant is much more robust to optimization issues and less sensitive for hyperpramtaer tuning.

\begin{figure}[t]
%\begin{wrapfigure}{r}{0.45\textwidth}
    \centering
    \includegraphics[width=0.63\linewidth]{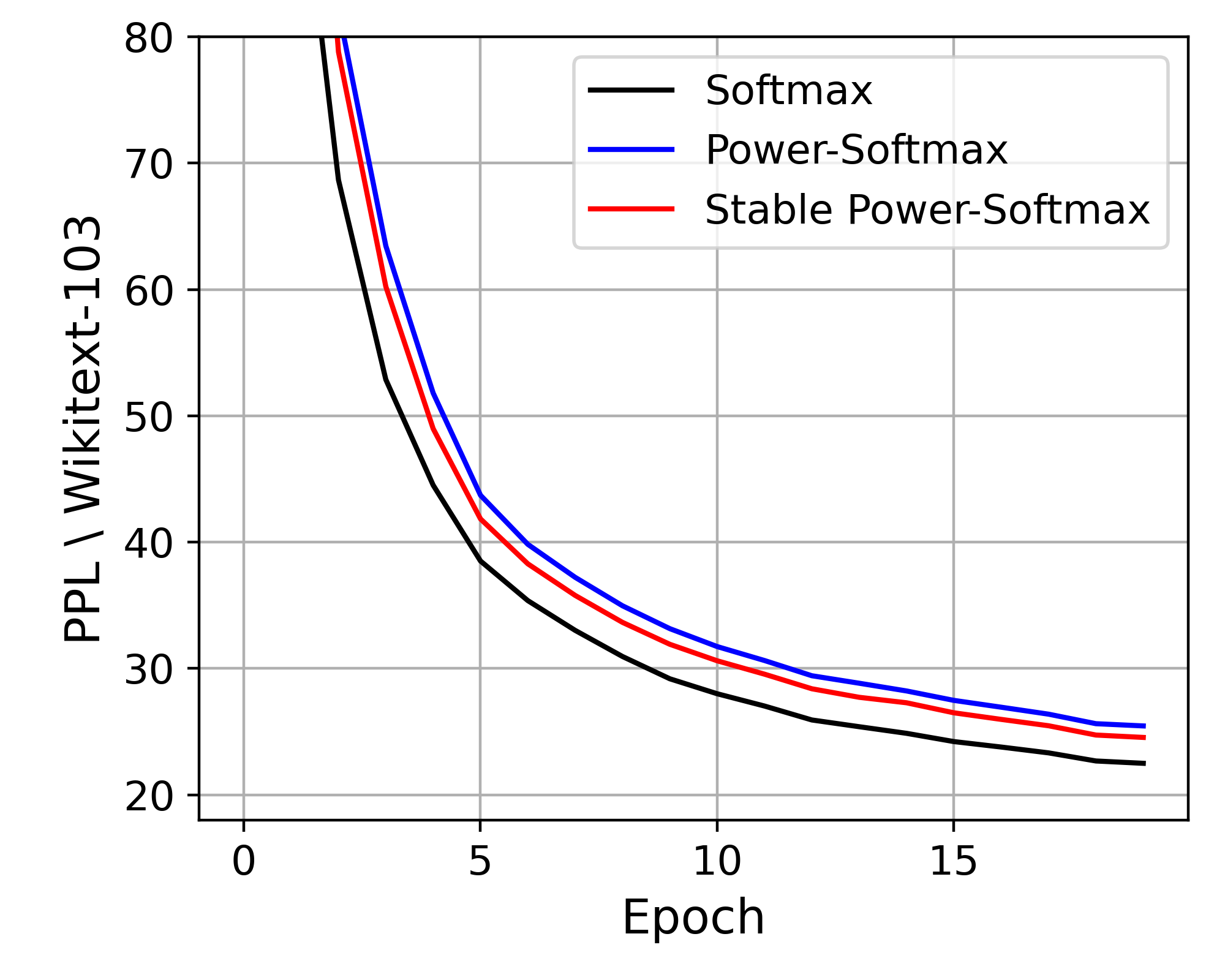}
    \caption{\small\textbf{The Significance of the stable variant.} Training curves for \gls{NTP} on Wikitext for large models. The stable variant (red) consistently outperforms the vanilla \PowerSoftmax (blue). 
    }
    \label{fig:stablityEmpirical}
%\end{wrapfigure}
\end{figure}

\noindent\textbf{$\boldsymbol{\epsilon}$-bounded division for Softmax  }  The HE-friendly attention variant from Eq.~\ref{eq:LipschitzPolyAttention} proposes adding $\epsilon$ to make the approximation problem of division easier, resulting in an approximation of a $\frac{1}{{\epsilon}^2}$-Lipschitz continuous function. Fig.~\ref{fig:epsilon} empirically supports this evidence by showing that the approximation error obtained by the Goldsmith method decreases as $\epsilon$ increases. Additionally, Fig.~\ref{fig:epsilonTraining} in App.~\ref{app:ablations} shows that higher $\epsilon$ values improve the training dynamics.

\subsection{Comparisons with SotA 
Methods\label{subsec:ComparisonSota}}

To the best of our knowledge, only two prior efforts have successfully presented fully polynomial transformers executed over FHE: (i) ~\citet{zimermanconverting}, who employ the \gls{AAT} approach, and (ii)~\citep{zhang2024neural}, who introduced NEXUS and focus on the \gls{PTA} regime. 
We begin by noting that our method exhibits superior scaling properties compared to both methods. This is evidenced by the fact that both methods concentrated on relatively simple text classification tasks, such as those found in the GLUE benchmark, with or without pre-training. In contrast, our models tackle much more complex tasks, including those that require \textit{reasoning and ICL capabilities}, which are %typically
associated with LLMs.

Additionally, when operating over encrypted data, our model is significantly more efficient than both methods. Specifically, Nexus incorporates three high-degree polynomial approximations per attention layer (for the exponential, division, and maximum functions), whereas our approach requires only a single non-polynomial division. Furthermore, \cite{zimermanconverting} exhibits substantially worse scaling properties for large models, and we were unable to successfully train deep transformers with 32 layers using their method. One possible explanation is that their pointwise attention lacks score normalization, resulting in training instability. A comparison is provided in Fig.~\ref{fig:robertaWikitext} (App.~\ref{app:roberta_params}). Although both methods include a single non-polynomial operation per attention head, ours is far more efficient for long contexts: their method applies an activation to each element of the attention matrix, yielding $L^2$ deep polynomial evaluations per head, whereas ours applies division once per row, yielding only $L$ deep polynomials and fewer \gls{HE} bootstrap operations. For additional empirical analysis, see App.~\ref{app:he-param}.

Beyond \gls{FHE}, several \gls{MPC}-based secure \gls{LLM} inference protocols exist, for example,~\citep{PrivMLLM, dong2025puma}, but they typically require gigabytes of communication overhead, making deployment impractical compared to \gls{FHE}-centric approaches.

% Beyond FHE, several MPC‑based secure‑LLM protocols exist \cite{PrivMLLM,dong2025puma}, but they typically impose substantial communication overheads, often multiple gigabytes, making deployment challenging compared to FHE‑centric approaches.

\begin{figure}[t]
    \centering
    %\begin{subfigure}[t]{0.45\textwidth}
    %    \centering
        \includegraphics[width=0.615\linewidth]{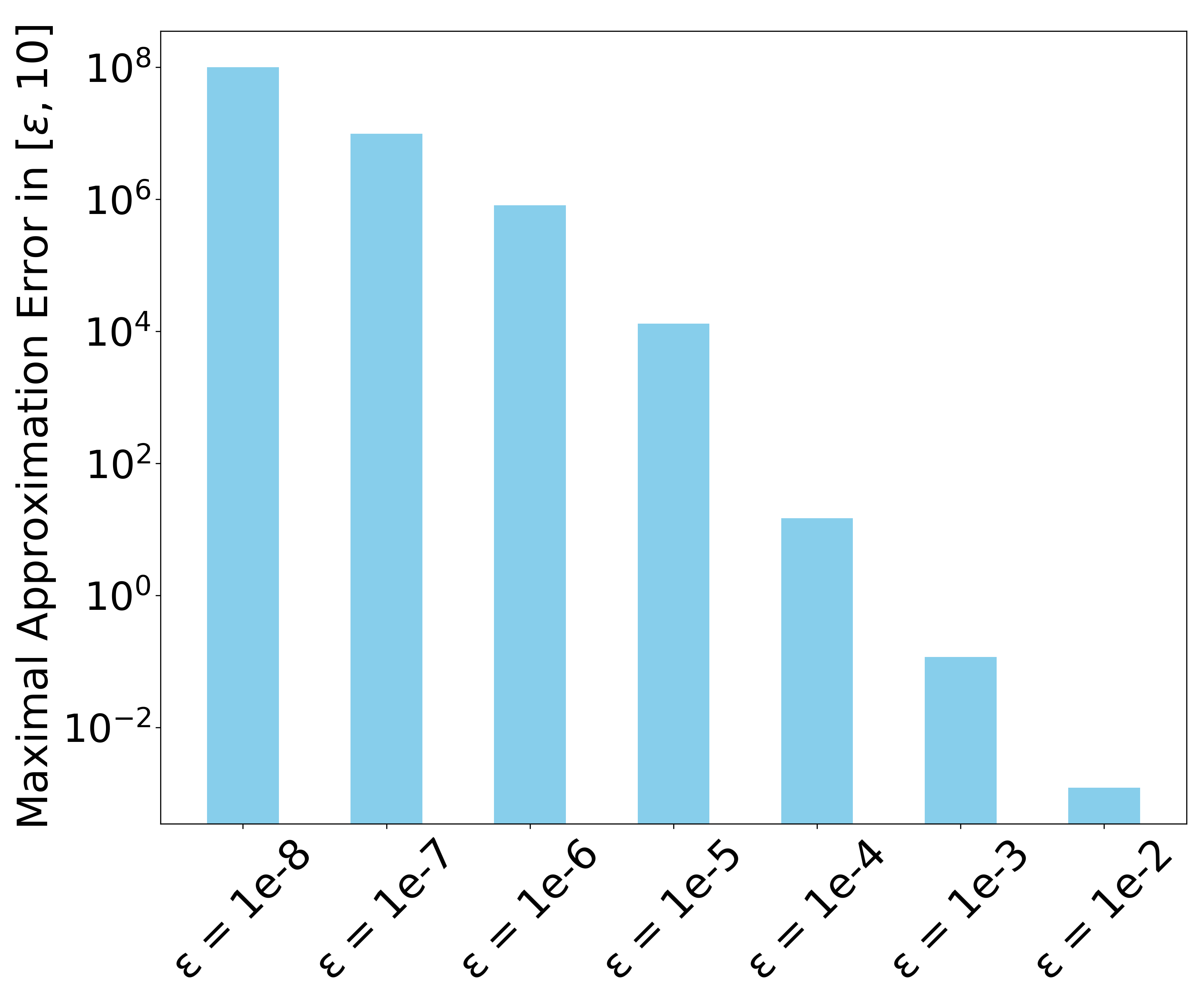}
     %   \caption{\small}
     %   \label{fig:epsilon}
    % \end{subfigure}
    % \hfill
    % \begin{subfigure}[t]{0.45\textwidth}
    %     \centering
    %     \includegraphics[width=0.75\linewidth]{graphs/mech_interp/mean_attention_distance_across_models.png}
    %     \caption{\small}
    %     \label{fig:AttnMeanDistance}
    % \end{subfigure}
    \caption{\small %(a) Polynomial approximation error for different values of $\epsilon$. (b) 
    Attention mean distance for different transformer variants.}
    \label{fig:epsilon}
    \vspace{-15pt}
\end{figure}

\subsection{Analyzing %PowerSoftmax Via
Attention Matrices \label{subsec:AttnMats}}

\PowerSoftmax introduces an important hyperparameter $p$ that differentiates it from the traditional \Softmax function. To better understand its mechanistic behavior, we examine how the attention matrices evolve with varying values of $p$. Our analysis reveals that as $p$ increases, the resulting attention matrices become more localized as depicted in Fig. \ref{fig:AttnMats}.
For instance, by comparing the first column (\PowerSoftmax with $p=4$) with the third column ($p=12$), we observe a significantly stronger diagonal in the latter, whereas the $p=4$ model displays a more uniform attention distribution.
Additionally, we empirically confirm this pattern by analyzing the average of the mean attention distance \citep{vig-belinkov-2019-analyzing} per model (i.e., averaged across all the layers and heads) as illustrated in Fig.~\ref{fig:AttnMeanDistance}.
Moreover, we observe that later layers tend to exhibit more longer-distance relationships  compared to earlier layers in both \PowerSoftmax and \Softmax. This finding is consistent with previous research \citep{vig-belinkov-2019-analyzing}. %Additional analysis can be found in the Figures \ref{fig:AttnMats} and \ref{fig:AttnMatsSamples} in the Appendix.
Additional analysis and visualizations can be found in %the Fig. \ref{fig:AttnMatsSamples} 
in Appendix \ref{app:attn_matrices}.

\section{CONCLUSION AND LIMITATIONS\label{sec:conclustions}}

We presented a method for training polynomial \glspl{LLM} with approximately 1.4 billion parameters, significantly larger than those employed in previous works. For that, we introduced an HE-friendly alternative to self-attention, which we demonstrate performs comparably to the original model. This variant allows us to present the first polynomial \gls{LLM} with zero-shot and reasoning capabilities. Despite the promising results, a full evaluation of the auto-regressive generative abilities of our models in both sequential decoding over plain and encrypted environments has not yet been conducted. For future work, we plan to investigate these aspects further and explore techniques to reduce the model's latency when operating on encrypted data.

\bibliography{ref}
\bibliographystyle{plainnat}

\clearpage

\appendix

\thispagestyle{empty}

\onecolumn

\section{EXPERIMENTAL SETUP AND HYPER-PARAMETERS\label{sec:experimental}}
All training experiments were conducted on public datasets using the PyTorch framework. Results were averaged over three random seeds, with experiments running on two A100 80GB GPUs for a maximum of two days, except for those involving the Pile dataset, which were run for up to three days on eight A100 40GB GPUs.% The hyper-parameters are provided in Tables {\color{black} XXX and XXX}.

% \noindent\textbf
\subsection{GPT} \label{app:pythia_params}
% \quad 
We used the framework of neox-gpt\footnote{\url{https://github.com/EleutherAI/gpt-neox}} \citep{gpt-neox-library} with its configuration of Pythia to train the 70M and 1.4B models. For this process. The replacement process is done as follows:

\begin{enumerate}
    \item Load a checkpoint of the pre-trained model.
    \item Replace \Softmax with \PowerSoftmax, with $p=4$, and employ continual pre-training over the Pile dataset for 100 iterations.
    \item Finetune the model with range-loss to minimize c and the input to GELU. This process takes around 17K iterations.
    \item Apply polynomial approximation.
\end{enumerate}
Table \ref{tab:pythia_params} shows the specific hyperparameters used for this process.

\begin{table}[h!]
    \centering 
      \caption{HE-related configuration for Pythia 1.4B, 70M, and RoBERTa models}
    \begin{tabular}{lccc} 
        \toprule
        \textbf{Parameter} & \textbf{GPT 1.4B} & \textbf{GPT 70M} & \textbf{RoBERTa-Base} \\ \midrule
         Sum power weights epsilon & $1e{-4}$ & $0.001$ & $1e{-4}$\\ 
        \PowerSoftmax loss weight ($c$) & $1e{-4}$ & $1e{-4}$ & $0.01$ \\ 
      
        \GELU loss weight & 0.001 & $1e{-4}$ & $0$ \\
        Learning rate & $4e{-5}$ & $1e{-4}$ &  $1e{-4}$\\ 
        % Iterations & 17k & 18k \\ 
        \hline
    \end{tabular}
  
    \label{tab:pythia_params}
\end{table}

\subsection{RoBERTa} \label{app:roberta_params}

We employed the RoBERTa framework \footnote{\url{https://github.com/facebookresearch/fairseq/blob/main/examples/roberta}} \citep{ott2019fairseq} and configuration to train and fine-tune the base model with 125M parameters for three GLUE tasks: SST-2, QNLI, and MNLI. The process was carried out as follows:

\begin{enumerate}
    \item Load a checkpoint of the pre-trained base model.
    \item Replace \Softmax with \PowerSoftmax, with $p=6$, and continually pre-train the model on the OpenWebText dataset for 1250 iterations.
    \item Fine-tune the model individually for each of the three GLUE tasks for up to 10 epochs. This fine-tuning followed the procedure described in the original RoBERTa paper, except for substituting the Tanh activation function in the classification head with a Sigmoid, which we found to perform better under HE. %improved performance during polynomial evaluation. 
    
    % We fine-tuned for 10 epochs using the configuration from the original paper, applying early stopping based on the accuracy of the development set. 
    \item %After selecting the best-performing models, 
    Perform an additional fine-tuning step using range-loss with \PowerSoftmax loss weight for 10 epochs. The \GELU ranges were narrow enough and did not require tuning. 
    \item Apply polynomial approximation.
\end{enumerate}

We reported accuracy results in Tab. \ref{tab:glue}. See Table \ref{tab:pythia_params} for the specific hyperparameters. %The results of SST-2 and QNLI are reported as an average of three random seeds.
%tasks across three random initializations.

Additionally, we train RoBERTa models with 12 layers from scratch over the Wikitext-103 benchmark for three types of attention: (i) \Softmax (black), (ii) our \PowerSoftmax (blue), and (iii) the Scaled-ReLU (red) attention baseline of~\citet{zimermanconverting}, all using the same training procedure and hyperparameters optimized for the vanilla \Softmax-based transformer. Training curves are averaged over three seeds and presented in Fig.~\ref{fig:robertaWikitext}. As shown, the Scaled ReLU variant is not competitive with the variants that employ proportional normalization. While \Softmax achieves better final results, it converges slightly slower than the \PowerSoftmax variant. With the implementation of early stopping, the models achieved average perplexity of 8.48 for \Softmax, 8.69 for \PowerSoftmax, with the Scaled ReLU lagging behind with 9.12.

\begin{figure}
    \centering
    \includegraphics[width=0.52\linewidth]{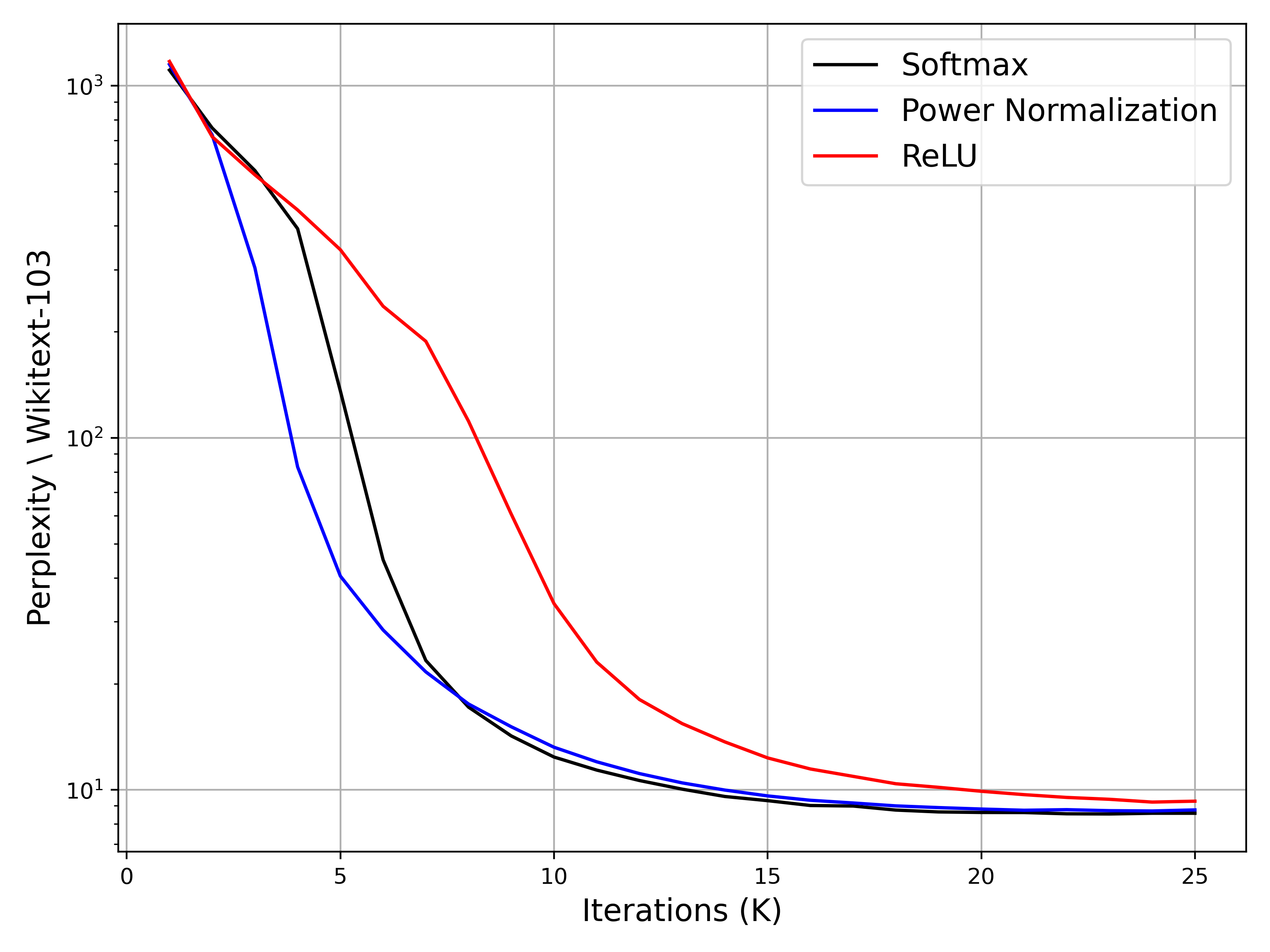}
    \caption{Comparison of training curves for 12-layer RoBERTa models with different attention mechanisms on the Wikitext-103 benchmark. The \PowerSoftmax variant (blue) converges faster than \Softmax (black), while the Scaled-ReLU baseline (red) underperforms. Curves are averaged over three seeds.}
    \label{fig:robertaWikitext}
\end{figure}

\section{ADDITIONAL ABLATION STUDIES\label{app:ablations}}

To gain a clearer understanding of the impact of $\epsilon$ in \PowerSoftmax-based attention models, we trained several models using different values of $\epsilon$. As shown in Figure~\ref{fig:epsilonTraining}, our variants demonstrate robustness across various $\epsilon$ values in terms of training dynamics. However, Figure~\ref{fig:epsilon} shows that for larger values of $\epsilon$, the resulting approximation function for division becomes easier, and we consider these settings (as an example $\epsilon=1e-2$) to be preferred.

\begin{figure}[h]
    \centering
    \includegraphics[width=0.5\linewidth]{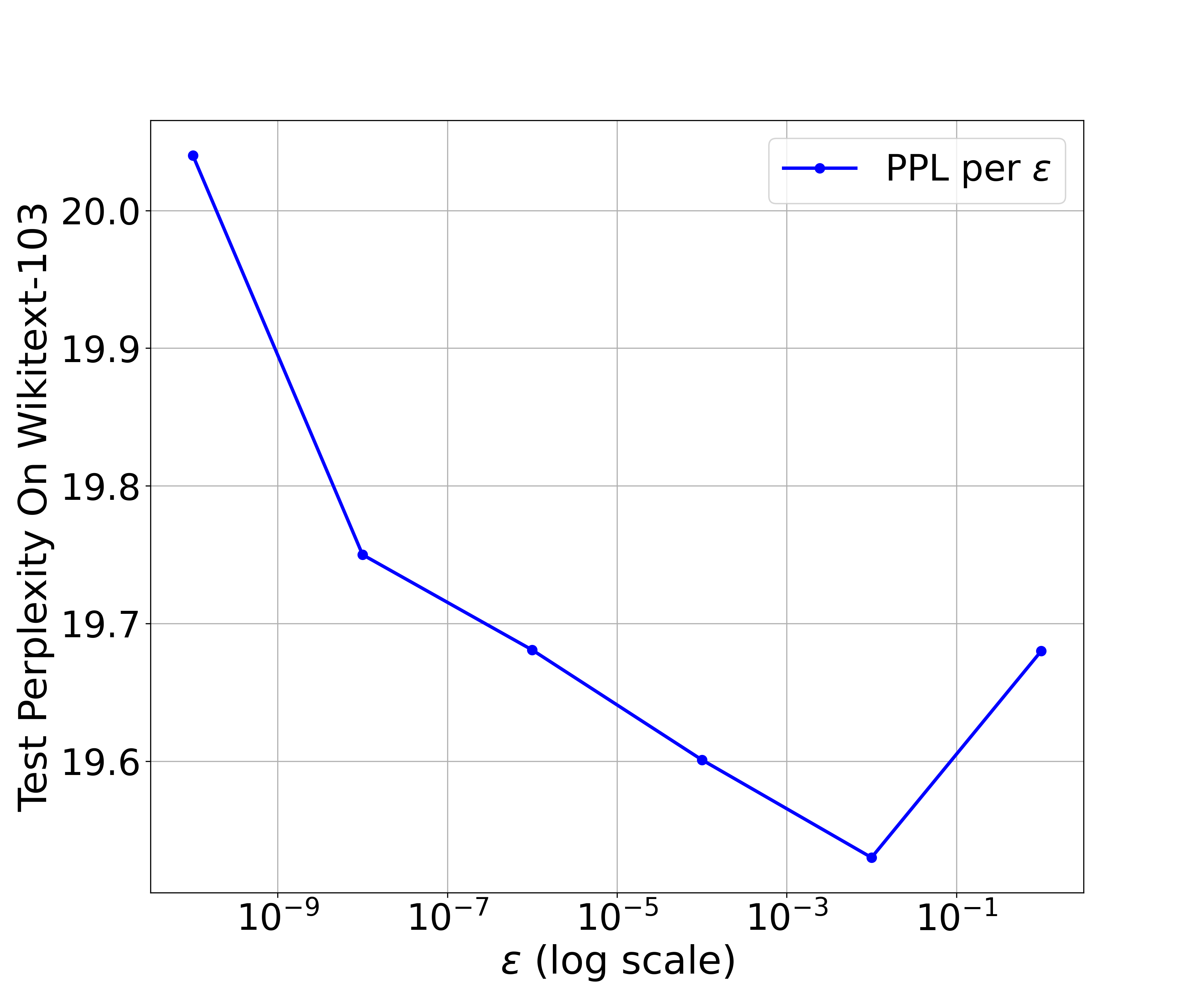}
    \caption{The impact of different values of $\epsilon$ on training dynamics of \PowerSoftmax-based models}
    \label{fig:epsilonTraining}
\end{figure}

\section{OUR POLYNOMIAL APPROXIMATIONS} \label{app:approximations}
Our \PowerSoftmax-based transformers utilize three polynomial approximations. For the division in \PowerSoftmax and the $\frac{1}{\sqrt{x}}$ function in LayerNorm, we apply the Goldschmidt approximation, following previous work in the domain~\citep{zimermanconverting,zhang2024neural}. For the GELU approximation, we use the following identity to reduce the problem to approximating the $\text{Sigmoid}$ function, which has been extensively explored in previous research in the \gls{HE} domain.

\begin{equation}
GELU(x) = x \cdot \text{Sigmoid}(1.702 \cdot x).
\end{equation}

\section{ANALYSIS OF THE PARAMETER $P$}
In this section, we provide a detailed analysis of the parameter $P$ and its role in shaping training dynamics. Our empirical study demonstrates that setting \(4 \leq P \leq 8\) consistently yields the best results across various tasks and datasets. A representative example is illustrated in Figure~\ref{fig:ablateP}, which shows training curves of several \PowerSoftmax-based language models trained for next-token prediction on WikiText-103. The figure depicts perplexity across epochs for different $P$ values.

It can be observed that when $P$ is set too high, training instabilities arise. For instance, models with $P > 8 $ exhibit oscillations in perplexity, indicating numerical issues. While this instability can be mitigated using our stable variant, the results do not outperform those achieved with \(4 \leq P \leq 8\).

Conversely, when $P$ is too low (e.g., $P=2$), the results are sub-optimal. One possible explanation is that the super-linear trend is overly conservative, limiting the model's capacity to capture complex relationships in the data adequately.

Thus, to maintain both performance and computational efficiency with low multiplication depth, we select $P = 4$ for all models.

\begin{figure}[h]
    \centering
    \includegraphics[width=0.5\linewidth]{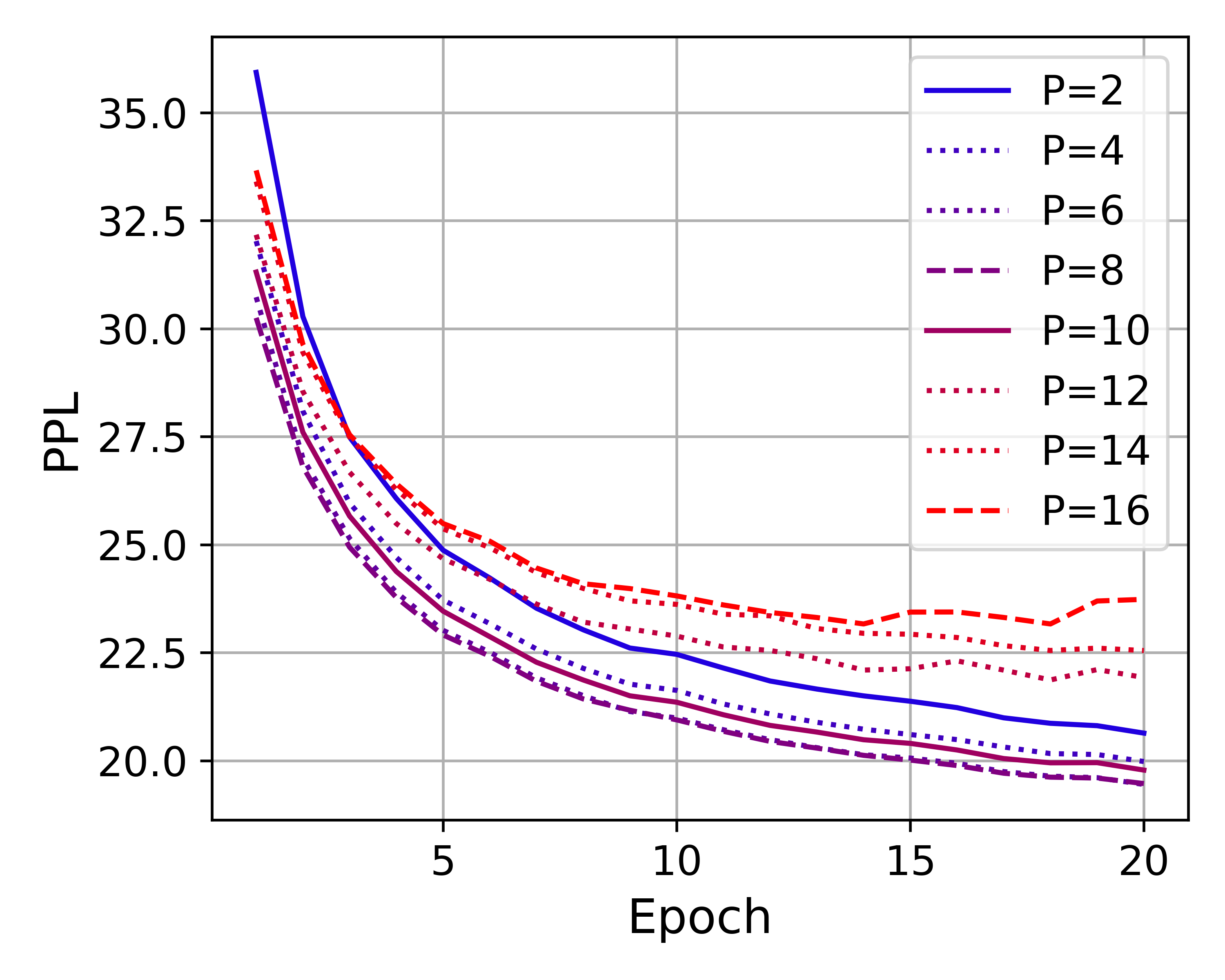}
    \caption{The impact of $P$ during training.}
    \label{fig:ablateP}
\end{figure}

\section{ADDITIONAL EMPIRICAL ANALYSIS\label{app:additionalExpirments}}
In addition to the analysis in Figure~\ref{fig:powerVssoftmaxrWikit}, we further explore the differences between \Softmax and \PowerSoftmax in vision by conducting experiments on Tiny-ImageNet, CIFAR-100, and CIFAR-10. The results are reported in Figure~\ref{fig:powerVsbaseVision}, showing that by the end of training, the models achieve similar performance.
\begin{figure}[h!]
    \centering
    \includegraphics[width=0.82\linewidth]{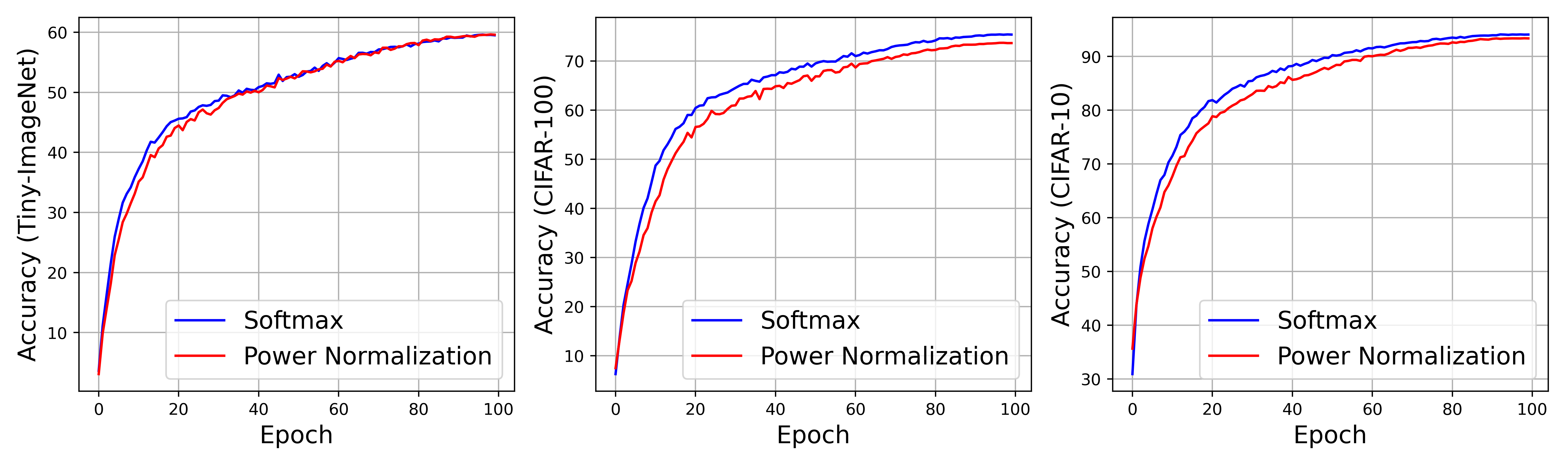}
    
        \caption{\textbf{Results on vision tasks.} Training curves for ViT variants with \PowerSoftmax (red) and the \Softmax baseline (blue). On the left, results are presented for Tiny-ImageNet and on the middle and right for CIFAR-100 and CIFAR-10 accordingly. %As can be seen, both models achieve similar, though not identical, performance across all datasets.
        }
    \label{fig:powerVsbaseVision}
    
\end{figure}

To provide a clear overview of the final performance of the fully trained models, we include Tab.~\ref{tab:comparison} summarizing the results across six datasets and various model sizes. The table consolidates data from Fig.~\ref{fig:powerVssoftmaxrWikit} and Fig.~\ref{fig:powerVsbaseVision}, showcasing perplexity for language modeling tasks and accuracy for image classification tasks. This summary highlights the consistent performance of our \PowerSoftmax variant compared to the baseline \Softmax models, across different configurations and datasets.

\begin{table}[ht!]
    \centering
    \caption{{\color{black}\textbf{Comparison of fully-trained models with \Softmax and \PowerSoftmax variants.} Results include perplexity (PPL, lower is better, $\downarrow$) for language modeling datasets and accuracy (denoted by `Top-1') higher is better, $\uparrow$) for image classification datasets.}}
    \label{tab:comparison}
    \smallskip
    \begin{tabular}{lccc}
        \toprule
        \textbf{Dataset}          & \textbf{Metric} & \textbf{Softmax} & \textbf{PowerSoftmax} \\
        \midrule
        WikiText-103 (6 layers)       & PPL ($\downarrow$)           &    21.18              & 21.27  \\          
        WikiText-103 (12 layers)      & PPL ($\downarrow$)            &    18.60              & 18.76      \\
        WikiText-103 (16 layers)      & PPL ($\downarrow$)            &     17.64             & 17.81       \\
        Text-8 (6 layers)         & PPL ($\downarrow$)            &     2.585             & 2.630       \\
        Text-8 (12 layers)        & PPL ($\downarrow$)            &   2.477               & 2.515       \\
        Text-8 (16 layers)        & PPL ($\downarrow$)            &     2.383             & 2.413         \\
        Pile                      & PPL ($\downarrow$)            &     16.05             & 16.51     \\
        Tiny-ImageNet             & Top-1 ($\uparrow$)       &     59.48             & 59.58       \\
        CIFAR-100                 & Top-1 ($\uparrow$)        &   75.39               & 73.66     \\
        CIFAR-10                  & Top-1 ($\uparrow$)        &   94.07               & 93.36      \\
        \bottomrule
    \end{tabular}

\end{table}

}

\section{IMPACT STATEMENT\label{app:impact}}
Our research introduces the first polynomial LLM, enabling \gls{HE}-based secure inference with transformers with billion parameters over encrypted data and through encrypted weights, and an HE-friendly transformer architecture. This advancement contributes to privacy-preserving deep learning, offering significant implications for data-sensitive sectors like healthcare and finance. This work aligns with the ethical need for responsible AI development by enhancing data privacy.

\section{EXPERIMENTS OVER HE}\label{app:he-param}
Our experiments used HElayers 1.5.4 \citep{helayers}, configured for CKKS with 128-bit security and polynomial degree of $2^{16}$. The underlying CKKS library was HEaaN, using its FGb parameter set with the following specific values: $log(QP) = 1555$, $N = 2^{16}$, $L = 9$, $h = 192$, $\lambda = 128$. 

Figures \ref{fig:tt0} and \ref{fig:tt1} illustrate the packing and chain-index flow of secure inference on the Pythia 70M model compiled with HElayers under FHE. For brevity, we included only the major nodes, such as plaintext matrix multiplication (PMM), ciphertext matrix multiplication (CMM), layer normalization, and the rotary positional embedding (RoPE) layers. The outputs and inputs of nodes include a tuple of information consisting of the tile tensor shape (packing method), the number of tiles (ciphertexts) at this point, and the chain index (ci) of the ciphertexts. Blue arrows indicate direct input/output data, while orange dashed arrows represent skipped trivial low-latency computation nodes in the figure. Whenever bootstrapping was required, a green arrow indicates the number of ciphertexts that needed bootstrapping. We distinguish between sequential (Seq) and parallel (Par) bootstrapping as follows: By writing \textit{4 Seq Bootstraps}, we mean that the same input tiles were consumed through their multiplication depth 4 times, requiring 4 bootstraps at this node. An example of such a scenario is performing the Goldschmidt method with 21 iterations (i.e., a multiplication depth of 22), which requires $22/9 \leq 3$ sequential bootstraps. In contrast, by writing \textit{4 Par Bootstraps}, we refer to bootstrapping 8 ciphertexts in parallel by using complex plaintext numbers (and not evaluating 4 bootstraps in parallel using threads, as we use a single-threaded GPU).

We use the tile tensor language \citep{helayers} to describe the packing method used by HElayers, ensuring consistency with HElayers logs and because we believe it makes the flow easier to read and understand. A detailed description of the tile tensor language is beyond the scope of this paper, and we refer the reader to \citep{he4ds}[Chapters 7-9], for a complete explanation. Alternatively, a good tutorial on tile tensors is provided in \citep{ccs-tutorial}.

For plaintext matrix multiplication, HElayers uses the diagonalization method from \citep{halevi2018faster}, extended to the matrix-matrix case, as described in \citep{he4ds}. For ciphertext matrix multiplication, HElayers uses the method from \citep{jiang2018secure}, which requires a multiplication depth of 3 per operation.

The latency of running one secure instance was 93 seconds. The PMMs took 28 seconds (30\%), multiplying by $W_q$, $W_k$, and $W_v$ requiring approximately 0.32 seconds each for the 18 PMM operations (total: 5.814 seconds). The (h24h/4h2h) PMMs took about 1.88 and 1.242 seconds, respectively (total: 18.732 seconds). The remaining time was spent on the 6 linear PMM operations. The CMM operations took 17 seconds (18\%), GELU and the Goldschmidt Inverse-SQRT took 0.65 seconds each (total: 7.7 seconds or 8\% of the total latency), and \PowerSoftmax took 1.475 seconds each (total: 8.82 seconds or 9.5\%). Other bootstrap operations took 18.563 seconds (19.4\%), while the remaining 15 seconds out of the 93 seconds (16\%) were spent equally on lower-latency operations such as addition, subtraction, rescaling, and repacking. Among the most time-consuming FHE operations, bootstrap took 33 seconds (34\%), encoding took 25 seconds (27\%), and rotation took 13.623 seconds (14\%).

\begin{figure}[ht!]
    \centering
    \includegraphics[width=0.85\linewidth]{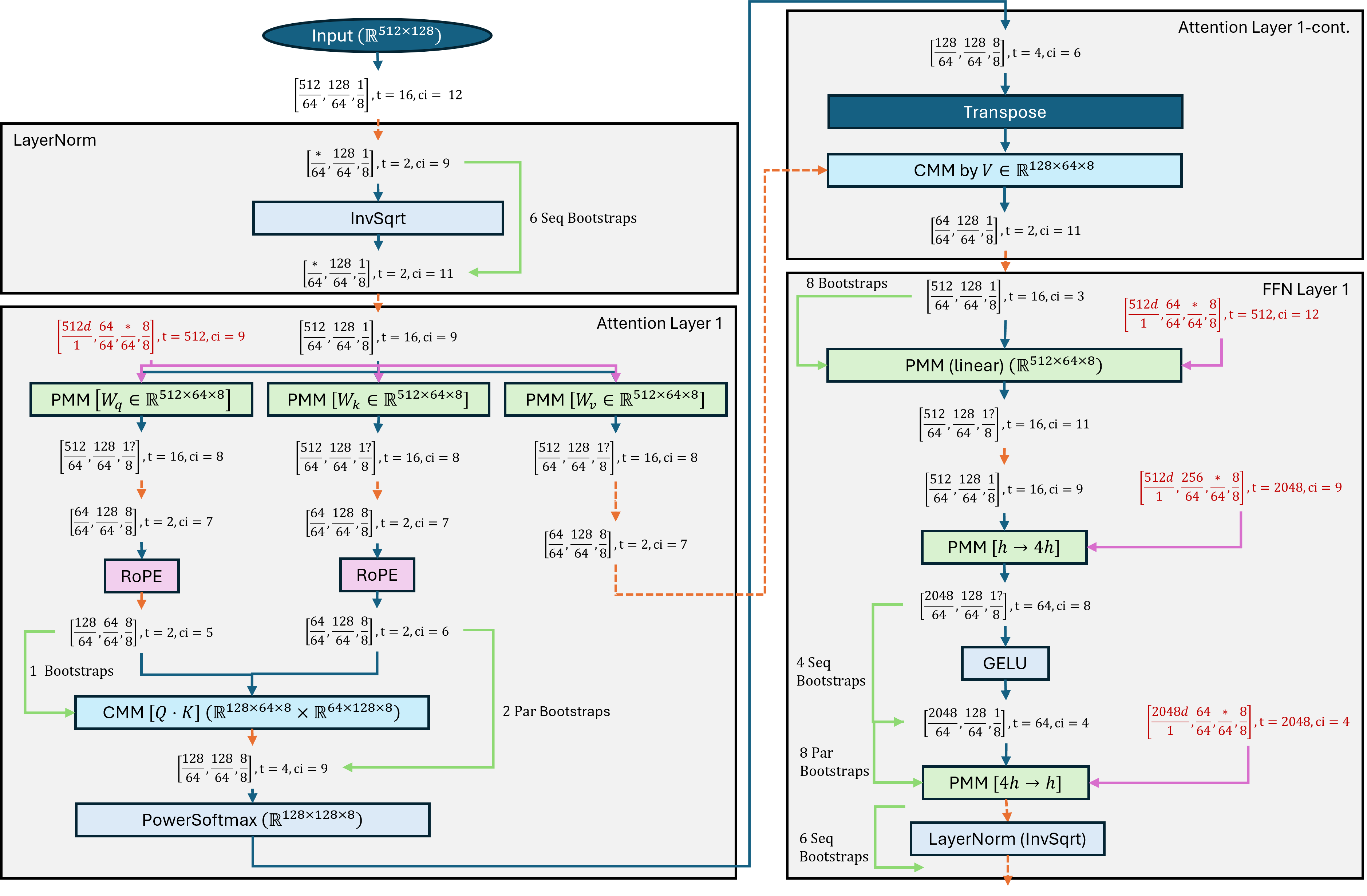}
    \caption{An illustration of the HE-encrypted Pythia 70M model up to the end of the first transformer layer. See the text for details and Figure \ref{fig:tt1} for the rest of the layers.}
    \label{fig:tt0}
\end{figure}

\begin{figure}[ht!]
    \centering
    \includegraphics[width=0.85\linewidth]{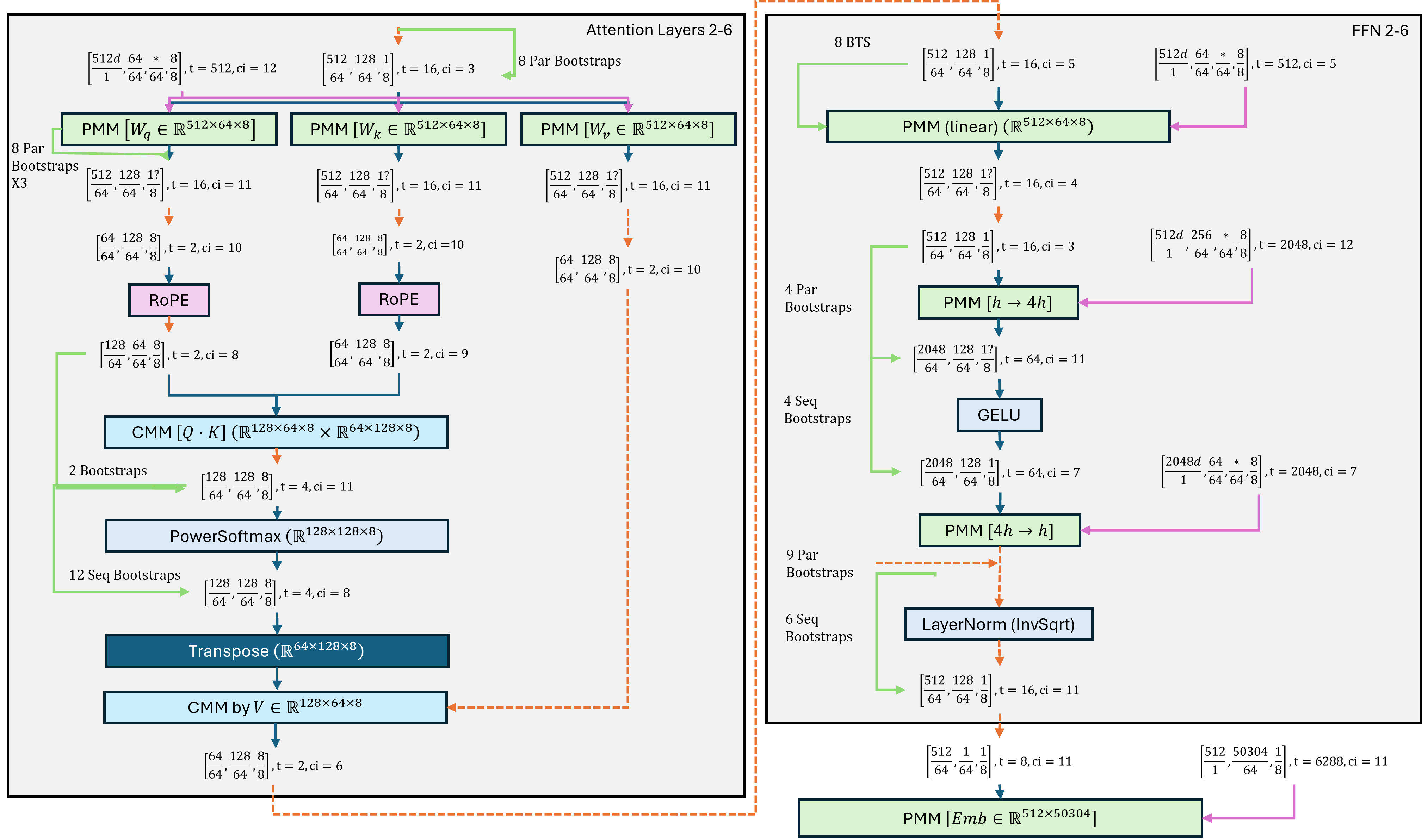}
    \caption{An illustration of the last five transformer layers of the HE-encrypted Pythia 70M model. See the text for details.}
    \label{fig:tt1}
\end{figure}

\paragraph{Comparing PowerSoftmax with Sota}
The latency reported by HElayers on one A100 80GB GPU for our \PowerSoftmax over 32 layers of 128 tokens ($128 \times 128$) took only $16$ seconds (and 0.6 seconds for $1 \times 32768$ input). In contrast, prior SotA -- NEXUS \citep{zhang2024secure}[Table IV] reported that \Softmax using 4 (instead of 1) A100 GPUs over $128 \times 128$ blocks for $12$ (instead of $32$) layers took $1.15$ seconds * $32$ (batch size) = $36.8$ seconds, $\times 2.3$ slower. Moreover, following the publication of this paper on arXiv, another study \citep{cho2024fast} addressing \Softmax was published. This paper introduces a new \Softmax approximation. Compared to \citep{cho2024fast}, our method reduces multiplication depth by at least $23\%$ when replacing \Softmax with our \PowerSoftmax in fine-tuned models. Moreover, the latency reported by \citep{cho2024fast} on an NVIDIA RTX-6000 for $128 \times 128 \times 32$(layers) + $1 \times 32768$ inputs was 90 seconds, slower than when using \PowerSoftmax. 

To further extend our empirical analysis of \PowerSoftmax over HE, we conducted additional dedicated experiments isolating the impact of the algorithm while eliminating the effects of different hardware, other aspects of model architecture, and HE libraries. All models were run under HE layers with the parameters described above, using a single A100 GPU. %
We chose to focus on the following baseline methods, as to the best of our knowledge, these are the only methods that propose algorithms for softmax that have been shown to work with fully polynomial models at relatively large scale: (i) Nexus, by~\citet{zhang2024secure}, which uses an 8-degree Taylor approximation for the exponent in softmax and Goldschmidt's approximation for division without specifying the number of iterations (we use 21, as in PowerSoftmax); and (ii) the method proposed by~\citet{cho2024fast} (see Alg. 2 in their work), which employs the normalize-and-square strategy for softmax evaluation, resulting in a multiplication depth of 48. Results are described in able~\ref{tab:softmax_methods_comparison} shows that in the tested scenarios, our method reduces latency by a factor of 20 compared to the square strategy, and by a factor of 9.7 compared to the Taylor and Goldschmidt approaches, both on Pythia 70M with 1024 tokens. Similar trends are observed in other regimes, such as shorter contexts and different models, highlighting the effectiveness of our method.

\begin{table}[h]
\centering
\small
\setlength{\tabcolsep}{5pt}
\begin{tabular}{lcccccc}
\toprule
\textbf{Model} & \textbf{\#Heads} & \textbf{\#Tokens} & \textbf{\#Layers} & Taylor and Goldschmidt (i) & Square strategy (ii) & PowerSoftmax \\
\midrule
Pythia 70M   & 8  & 128 / 1024 & 6  & 3.6 / 83.9   & 7.42 / 173.4 & 2.7 / 8.6  \\
Pythia 1.4B  & 16 & 128 / 1024 & 24 & 19.4 / 662.3 & 42.1 / 1371.6 & 11.1 / 51.8 \\
Llama 7B     & 32 & 128        & 32 & 46.1         & 84.05         & 16.6       \\
\bottomrule \\ 
\end{tabular}
\caption{Comparison of computation cost (in seconds) for three different polynomial alternatives to Softmax, lower is better.}
\label{tab:softmax_methods_comparison}
\end{table}

\section{THREAT MODEL}\label{app:threat}
We consider a two-party \gls{FHE} scenario where a semi-honest server runs inference and a data-owner submits encrypted queries. Whether the server holds the model weights encrypted or in plaintext depends on the trust relationship with the model owner; our solution is orthogonal to this choice, affecting only latency. We refer the reader to \citep{he4ds}[Chapter 3] for a formal treatment of this security model.

% \section{Threat model}\label{app:threat}
% Sec. \ref{sec:problem} lists two threat-model examples for secure inference for LLMs over HE: 1) using encrypted weights; or 2) using encrypted input samples (queries). When considering a 2-party scenario with FHE, it is commonly assumed that one party is the (untrusted, semi-honest) server who holds the model (encrypted or not) and the other party is the data-owner who performs the query and would like to avoid revealing the query to the server. The decision of whether the server holds the model encrypted or not depends on whether a third-party the model-owner agrees to share the data with the server. Our proposed solution is orthogonal to the above decision, which eventually only affects latency and for brevity we have decided to refer the reader to \citep{he4ds}[Chapter 3] that further describes this security model. 

% \noindent \textbf{Reformulate Attention Mask.\quad}%
\section{MASKED HE-FRIENDLY VARIANT\label{sec:attnMask}}%
Attention masks are a well-known technique used to manipulate self-attention by determining which tokens can attend to each other. Traditional LLMs leverage a binary mask $M$ for various applications. A notable example is the causal mask, employed for training LLMs via \gls{NTP}, a popular self-supervised learning scheme. These standard masking mechanisms are specifically designed for \Softmax-based self-attention (masked values were represented by $-\infty$ and used as an additive term) and should be reformulated for HE-Friendly Attention. Masking is applied via an element-wise product, denoted by $\odot$, as follows:

\
\begin{equation}\label{eq:LengthAgnosticMaskedAttention}
    \operatorname{Masked~HE-Friendly~Attn}(Q,K,V) = %\operatorname{Length-Agnostic~PowerSoftmax}
    \left(\frac{QK^T \odot M}{\sqrt{d_k}}\right), \quad M_{i,j} \in [0,1].
\end{equation}

\section{ADDITIONAL POLYNOMIAL ATTENTION VISUALIZATION\label{app:attn_matrices}}
In Fig.~\ref{fig:AttnMats} and Fig.~\ref{fig:AttnMatsSamples}, we present a visual analysis of attention matrices obtained from both the vanilla \Softmax-based models and the corresponding polynomial HE-friendly variants across different layers. Fig.~\ref{fig:AttnMats} depicts the attention matrices averaged over 3 seeds, all attention heads at a layer, and 1,000 examples. Additionally, to provide a comprehensive view of the attention matrices, Fig.~\ref{fig:AttnMatsSamples} contains random samples of attention matrices. All models rely on a BERT-like 12-layer causal model with a context length of 512, trained on Wikitext-103 for next-token prediction with the same training procedure. We use examples from the test set of Wikitext-103 as input samples.%{\color{black}As can be seen in figures, TBD. Some conclusion, or reference to main paper.}

Our analysis reveals that as $p$ increases, the resulting attention matrices become more localized as depicted in Fig. \ref{fig:AttnMats}.
For instance, by comparing the first column (\PowerSoftmax with $p=4$) with the third column ($p=12$), we observe a significantly stronger diagonal in the latter, whereas the $p=4$ model displays a more uniform attention distribution.
Additionally, we empirically confirm this pattern by analyzing the average of the mean attention distance \citep{vig-belinkov-2019-analyzing} per model (i.e., averaged across all the layers and heads) as illustrated in Fig.~\ref{fig:AttnMeanDistance}.

\begin{figure}[h]
    \centering
    \includegraphics[width=0.4\linewidth]{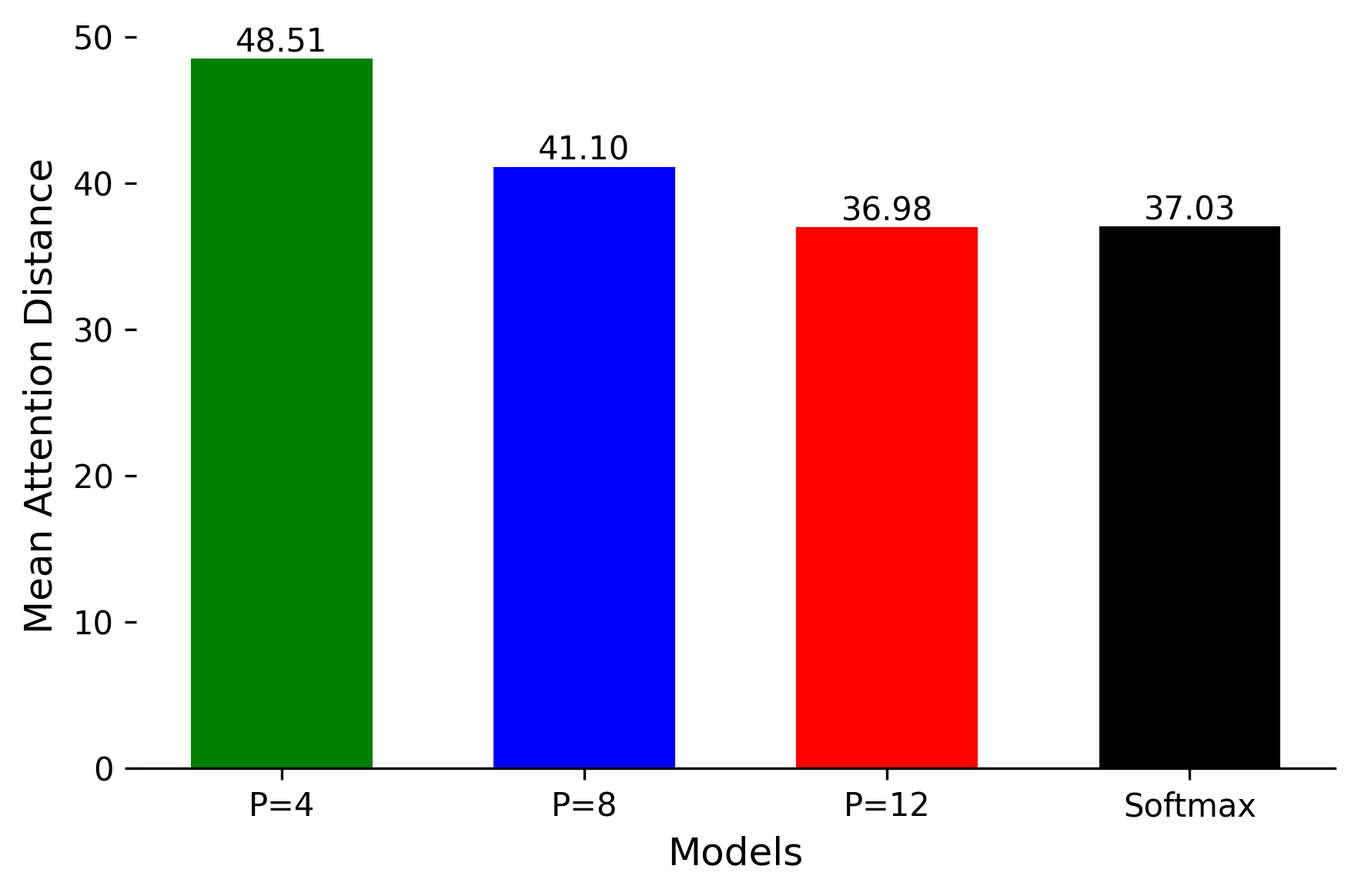}
    \caption{\small Attention mean distance for different transformer variants.}
    \label{fig:AttnMeanDistance}
\end{figure}

\begin{figure*}[h]
\centering
\begin{tabular}{cccc} 
  $p=4$ & $p=8$ & $p=12$ & $ \textbf{Softmax}$ \\
  \cmidrule(lr){1-4}
  %\midrule
    \includegraphics[width=0.16\textwidth]{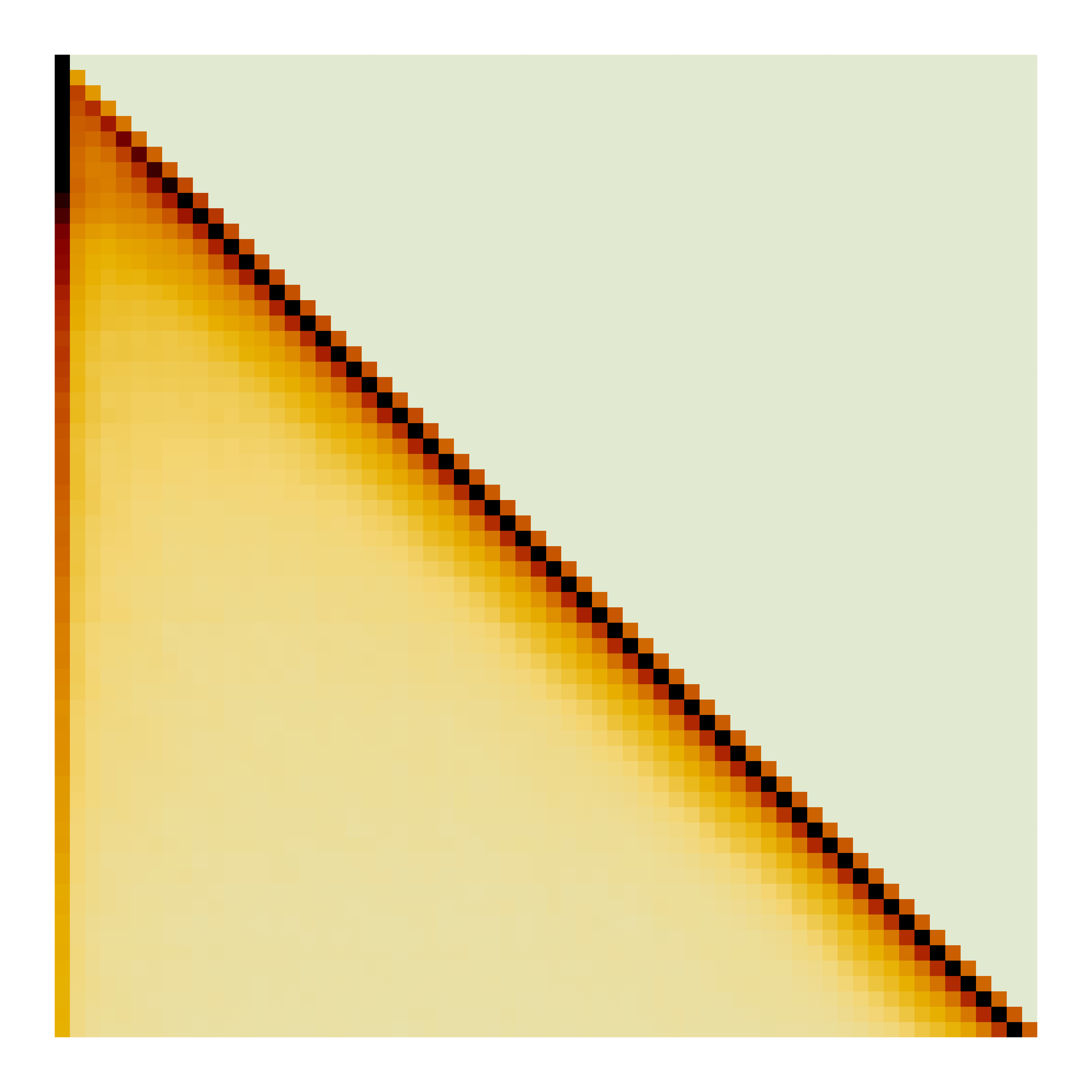} & \includegraphics[width=0.16\textwidth]{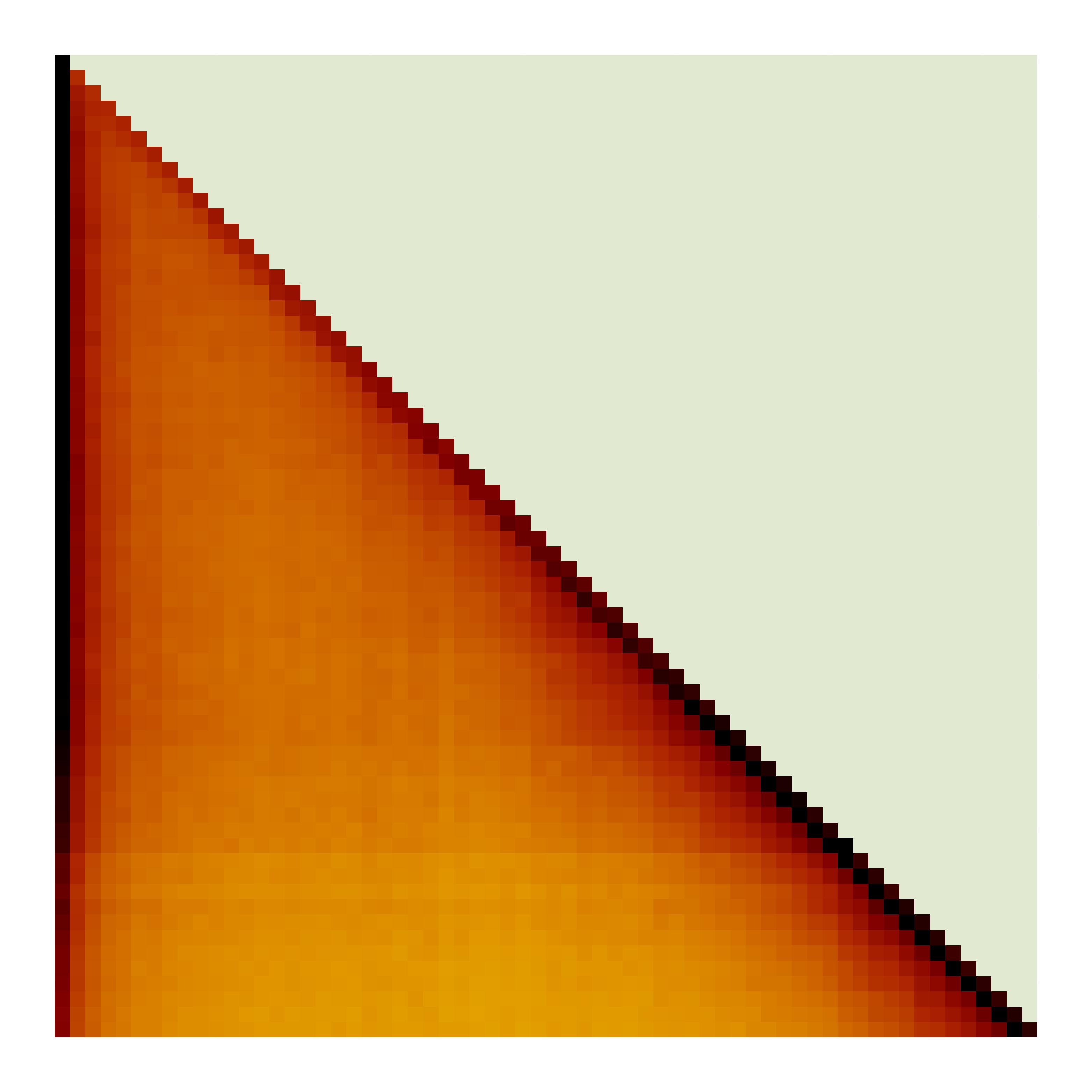} &
    \includegraphics[width=0.16\textwidth]{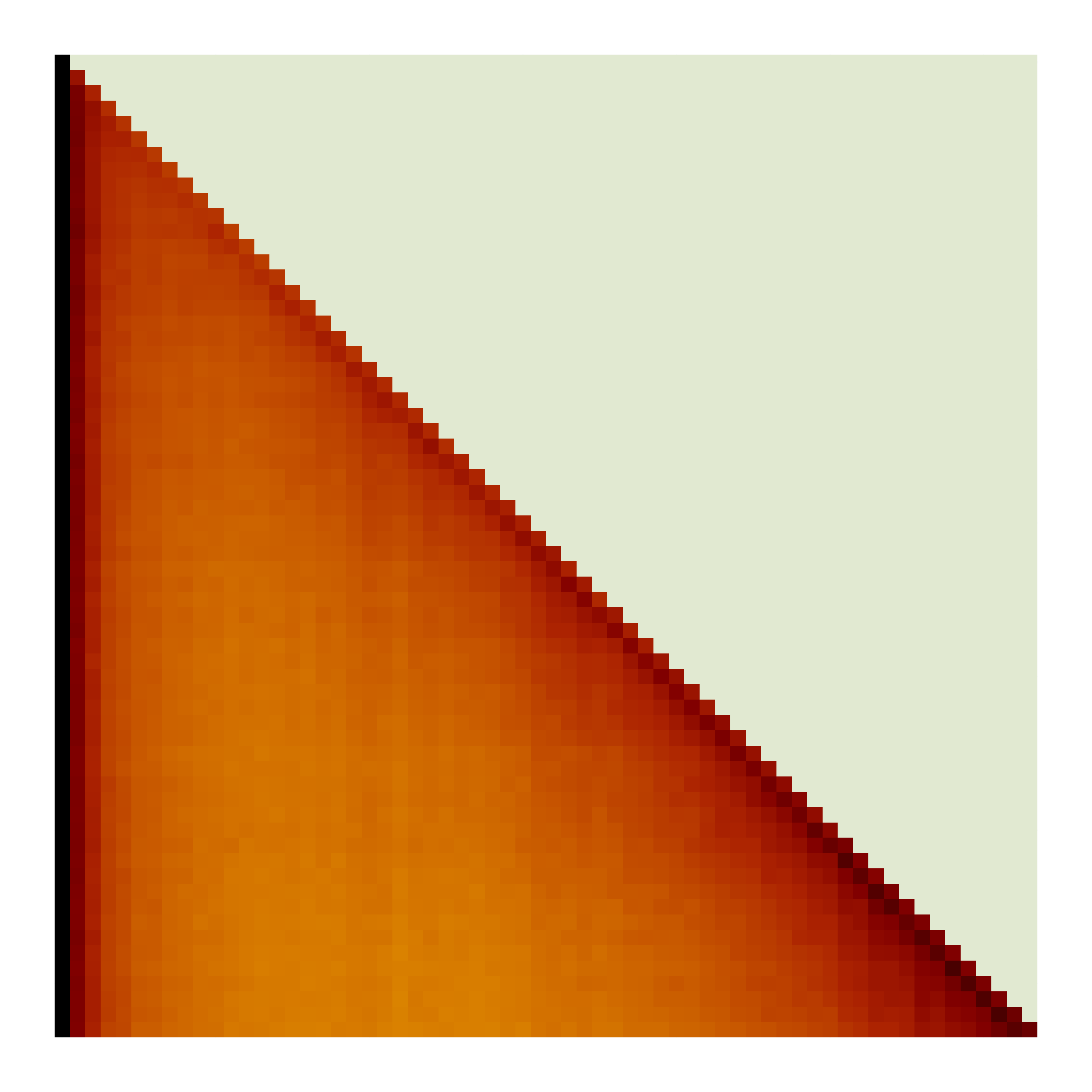} & \includegraphics[width=0.16\textwidth]{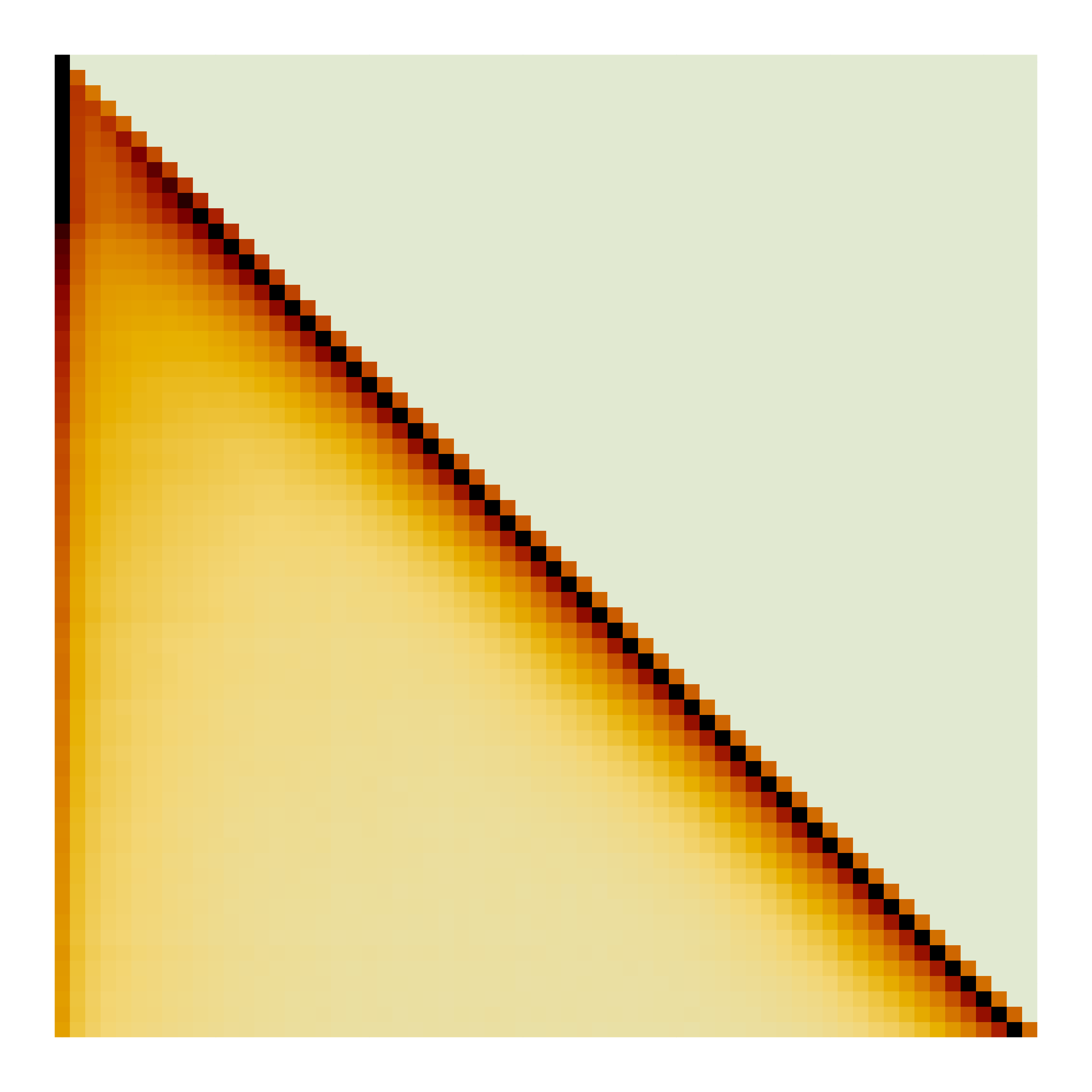}\\
    \multicolumn{4}{c}{Attention Layer 2} 
    \\   
     
  \includegraphics[width=0.16\textwidth]{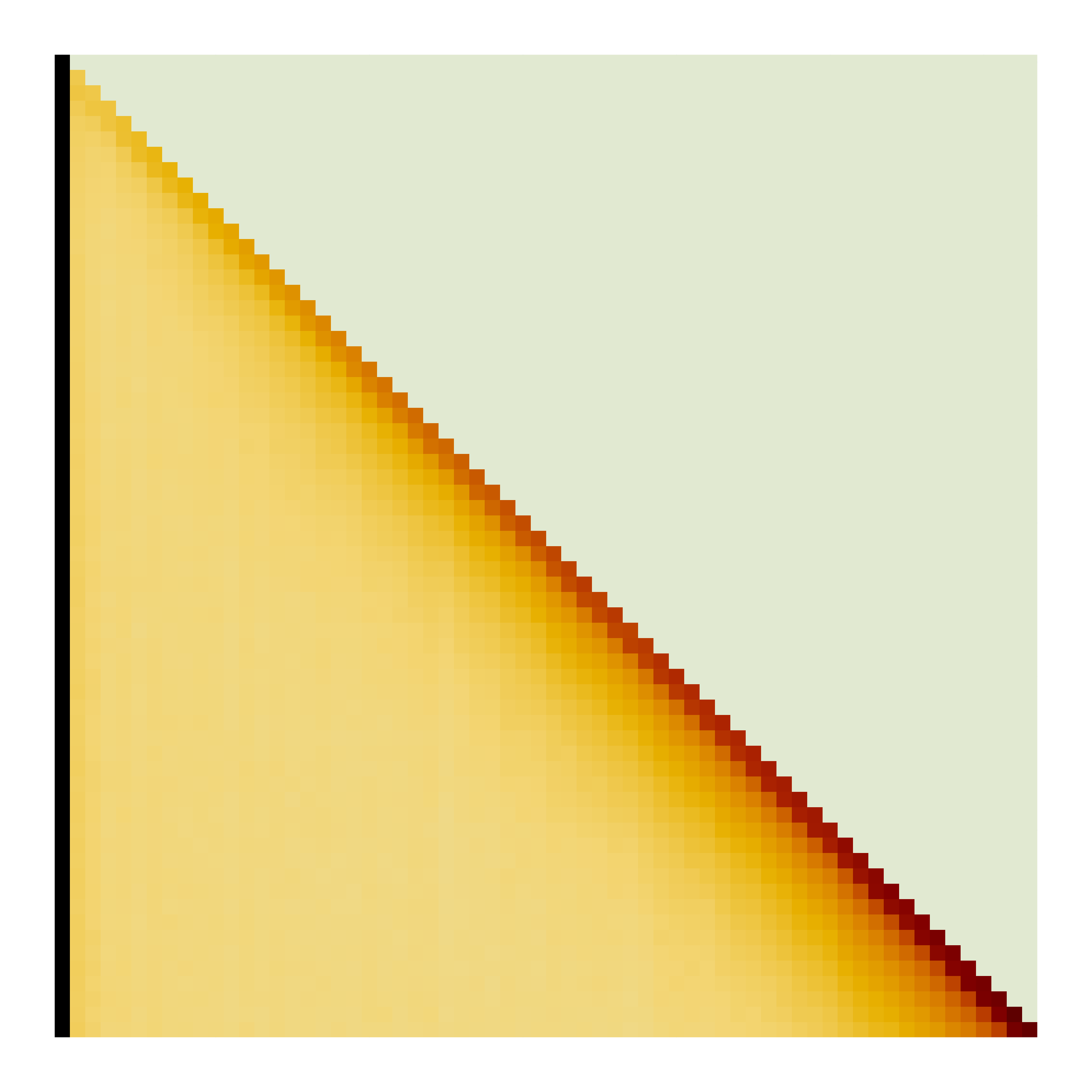} & \includegraphics[width=0.16\textwidth]{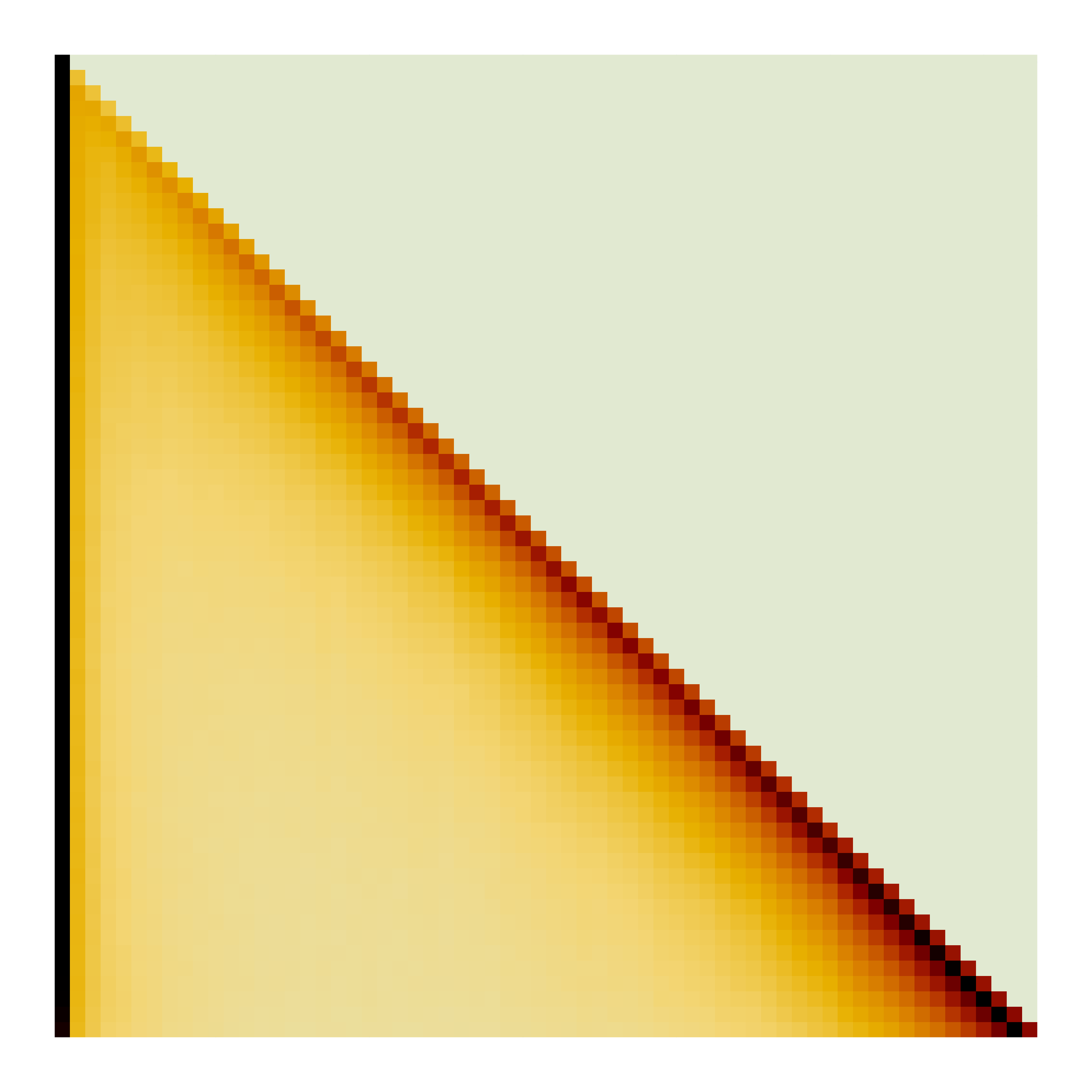} &
    \includegraphics[width=0.16\textwidth]{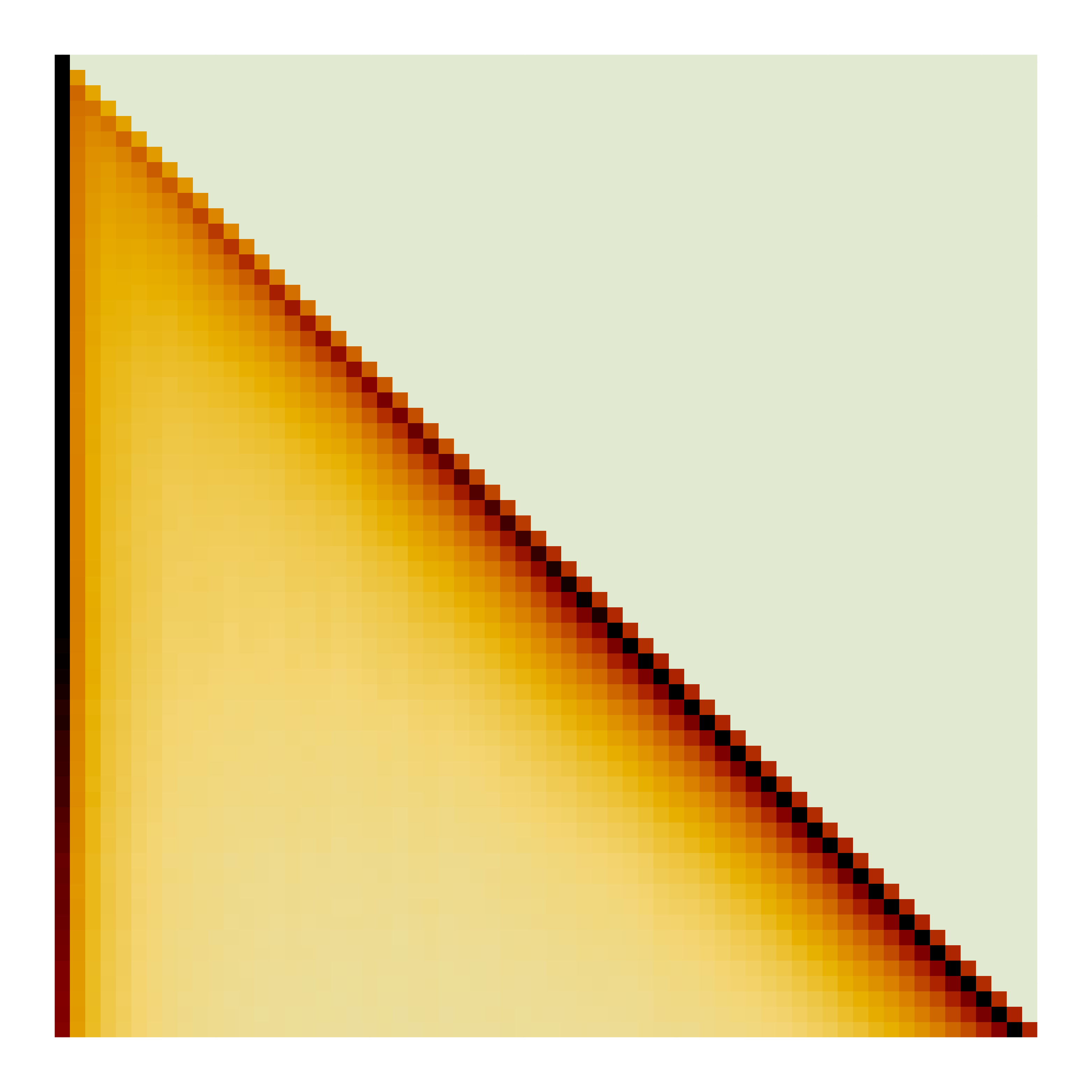} & \includegraphics[width=0.16\textwidth]{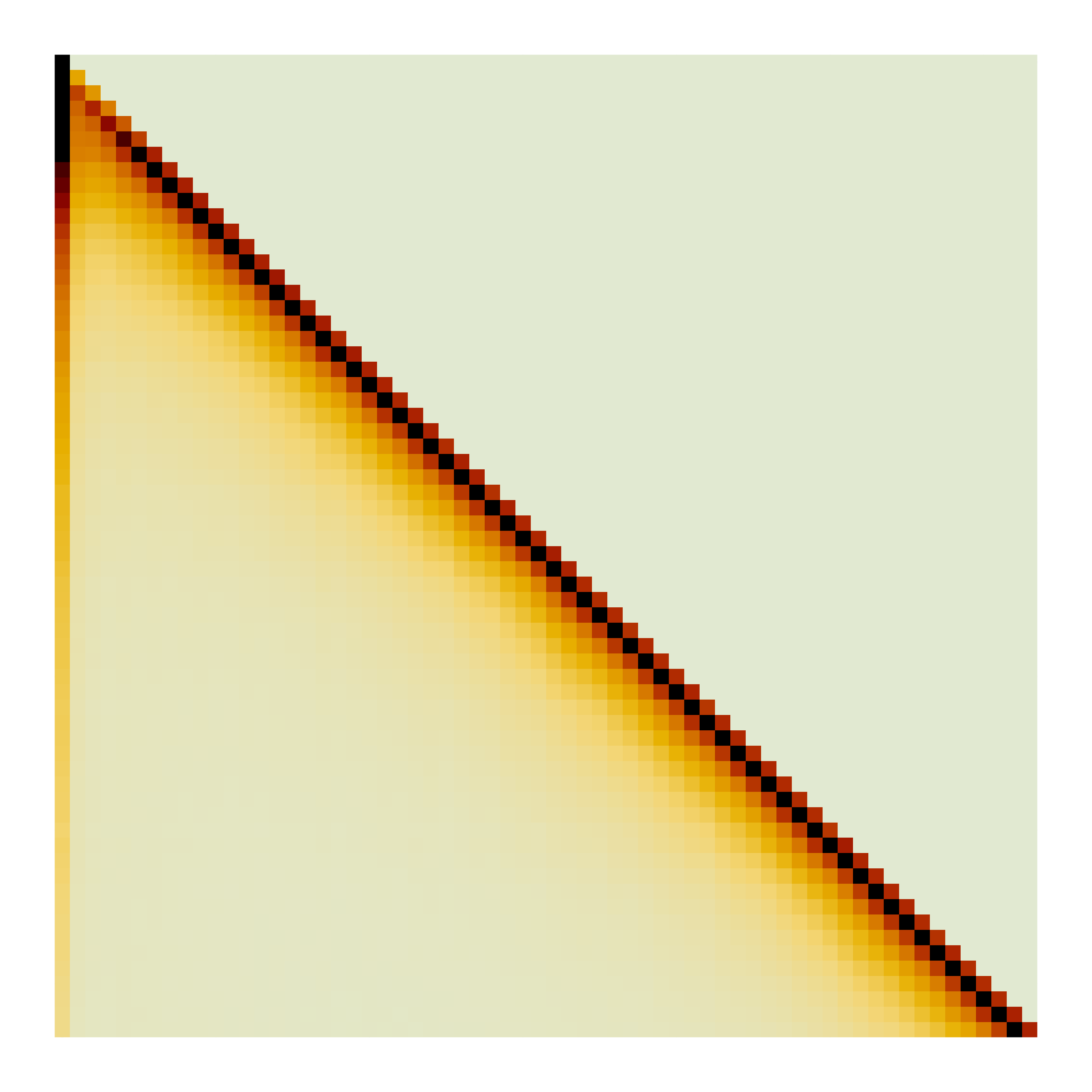}\\
    \multicolumn{4}{c}{Attention Layer 4}
 \\
  \includegraphics[width=0.16\textwidth]{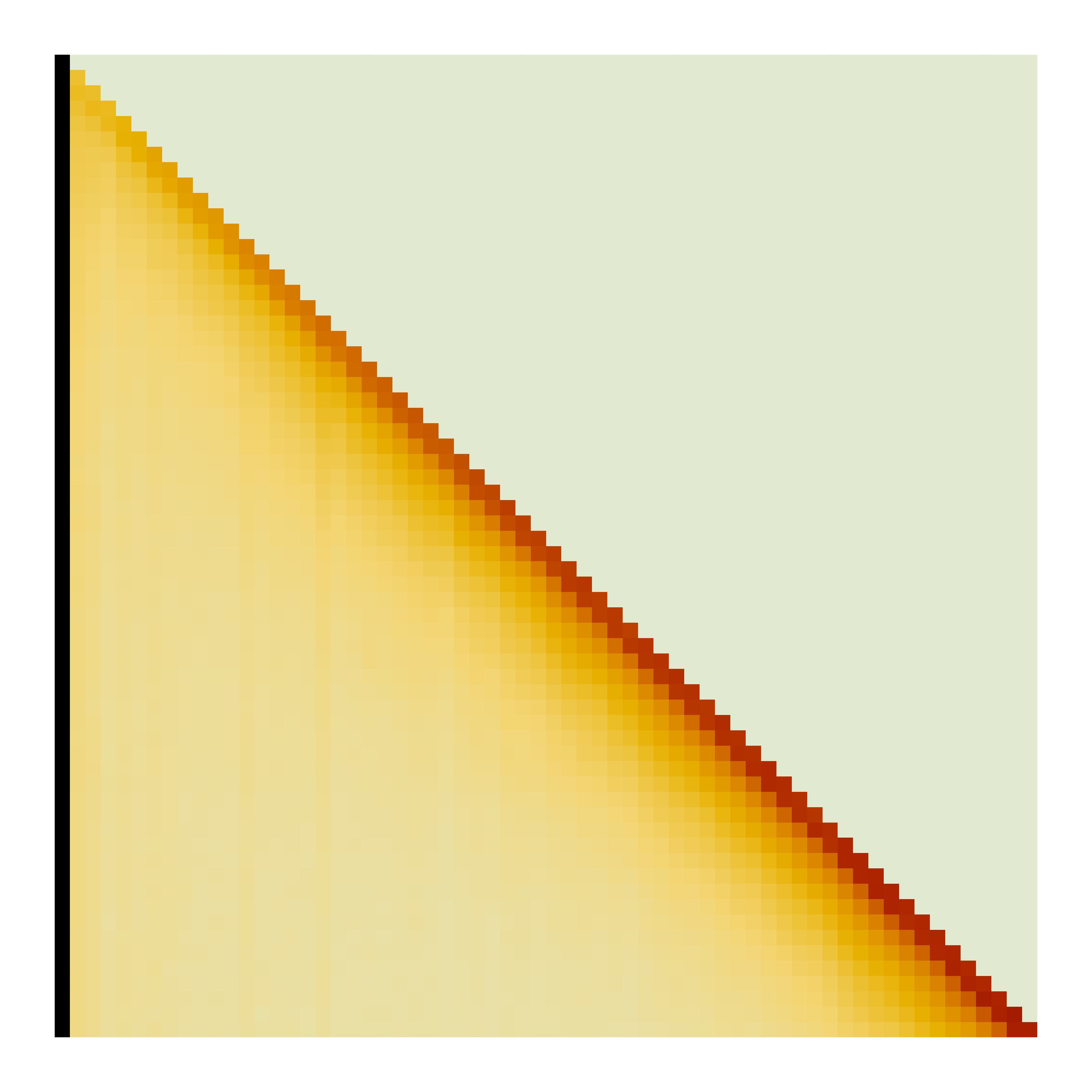} & \includegraphics[width=0.16\textwidth]{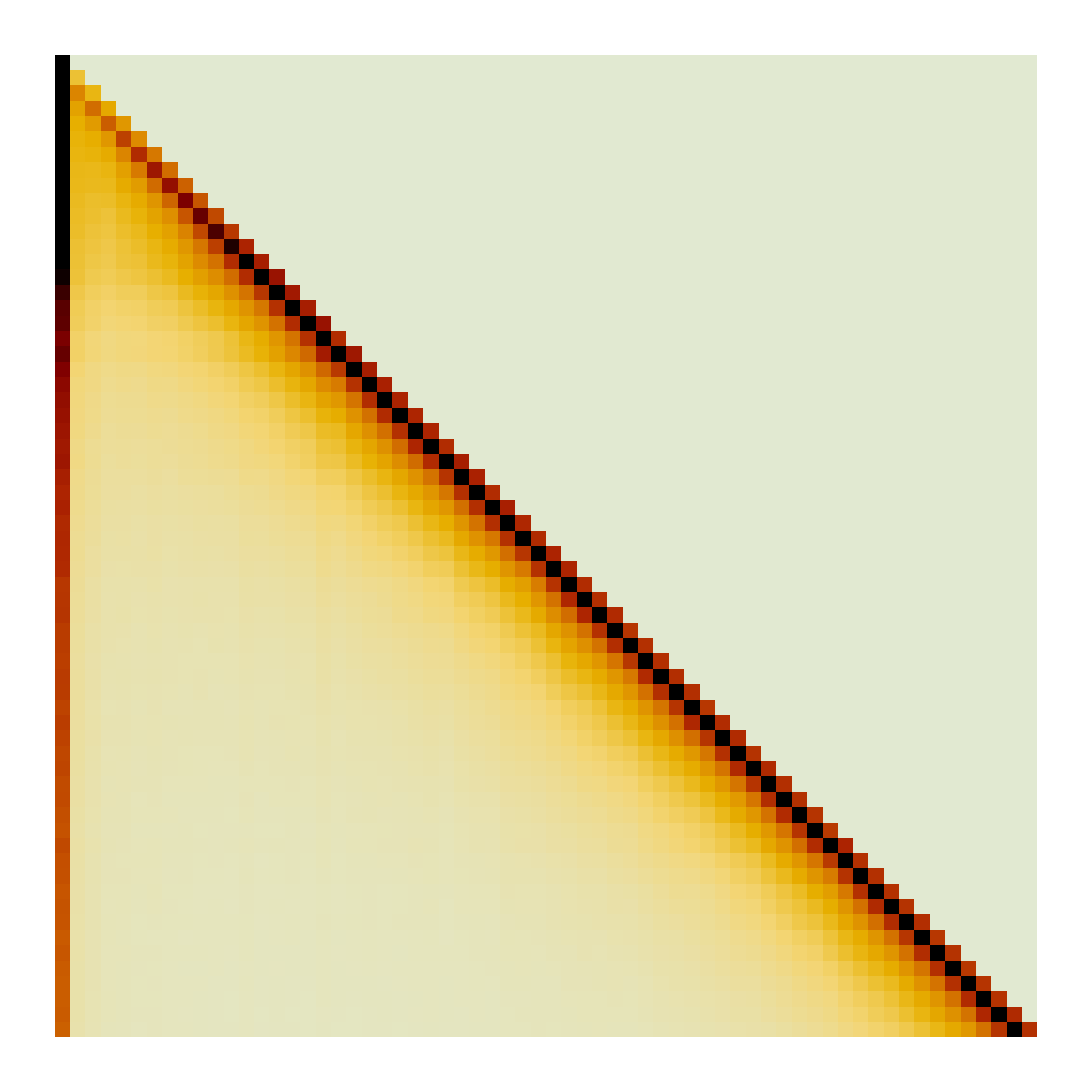} &
    \includegraphics[width=0.16\textwidth]{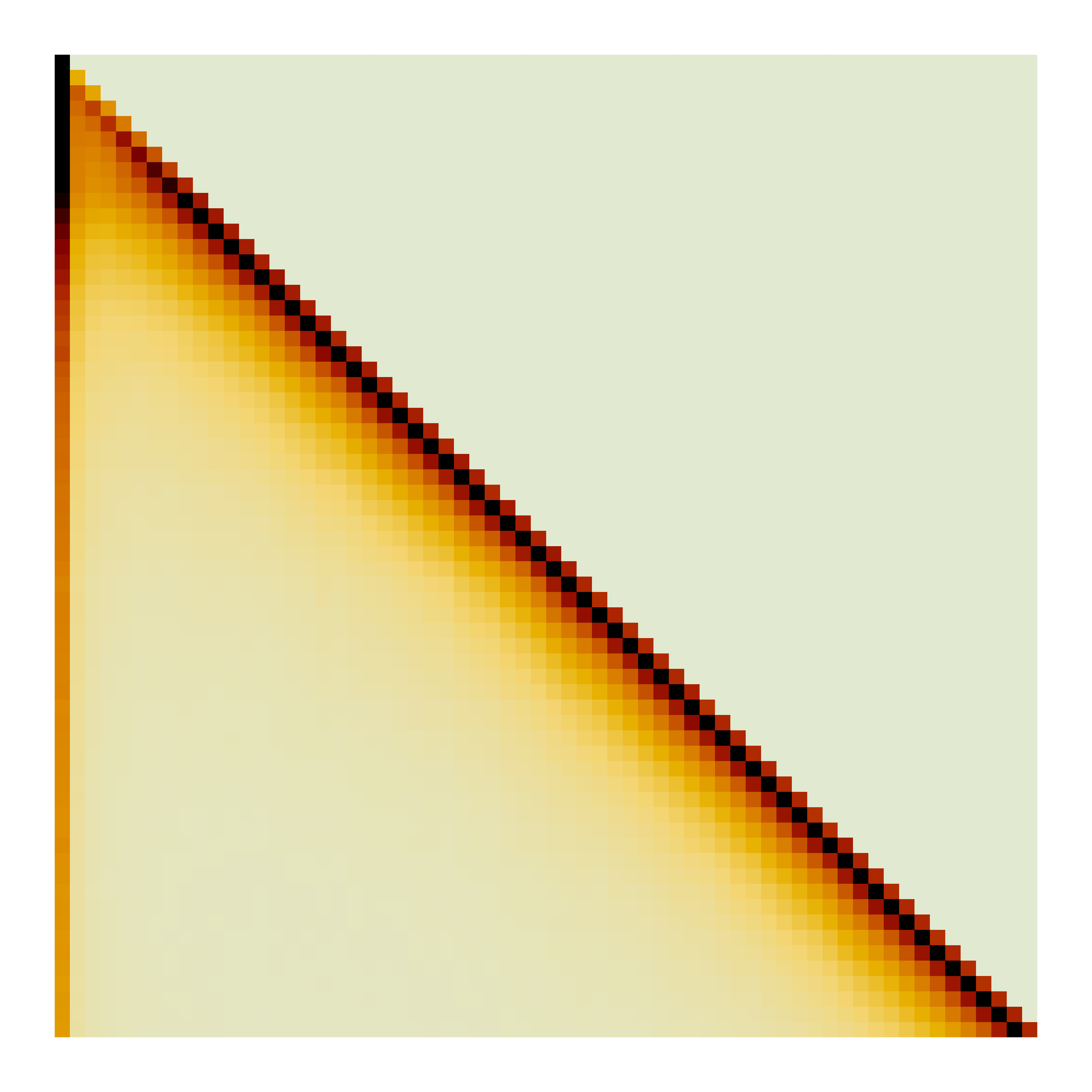} & \includegraphics[width=0.16\textwidth]{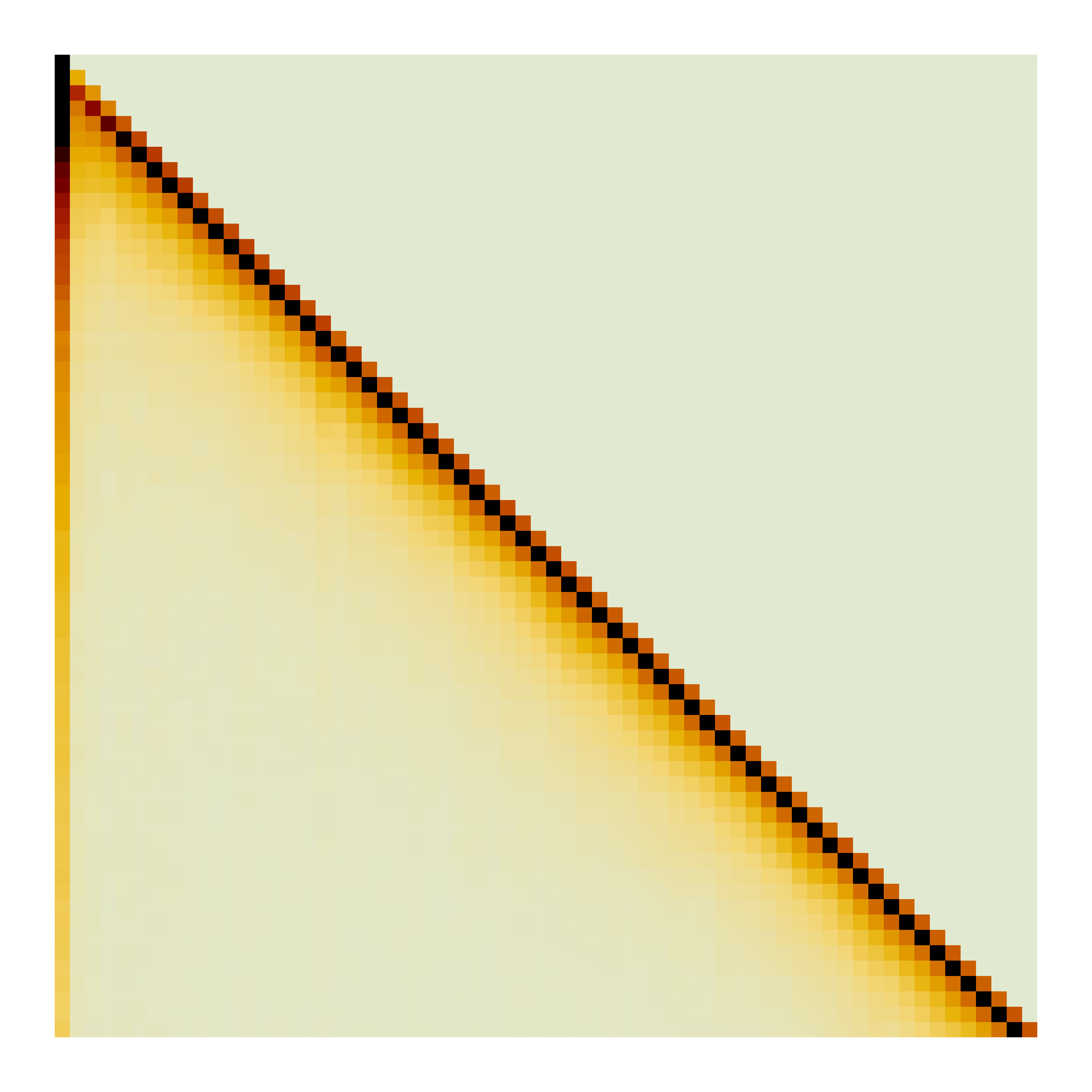}\\
    \multicolumn{4}{c}{Attention Layer 6}    
 \\
\includegraphics[width=0.16\textwidth]{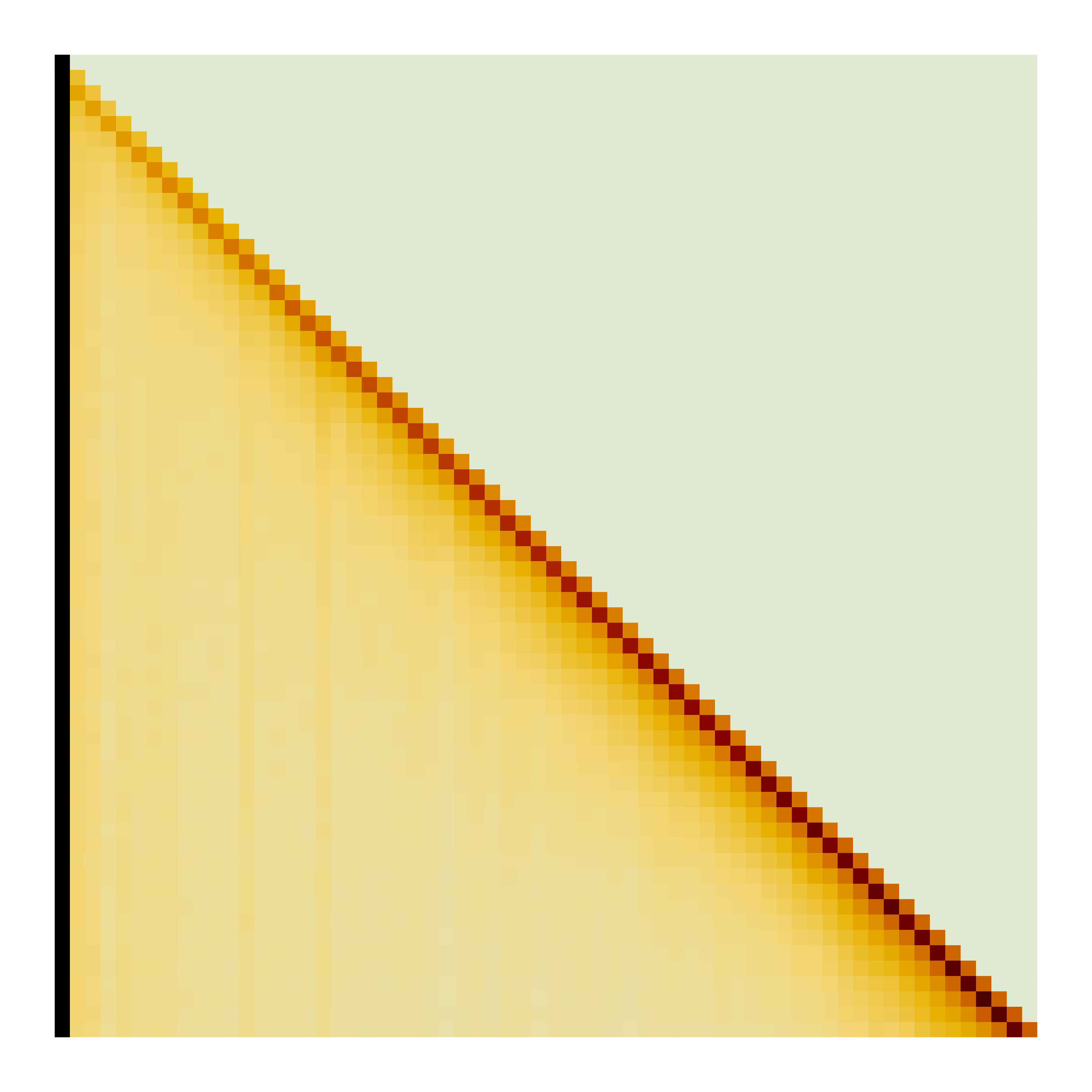} & \includegraphics[width=0.16\textwidth]{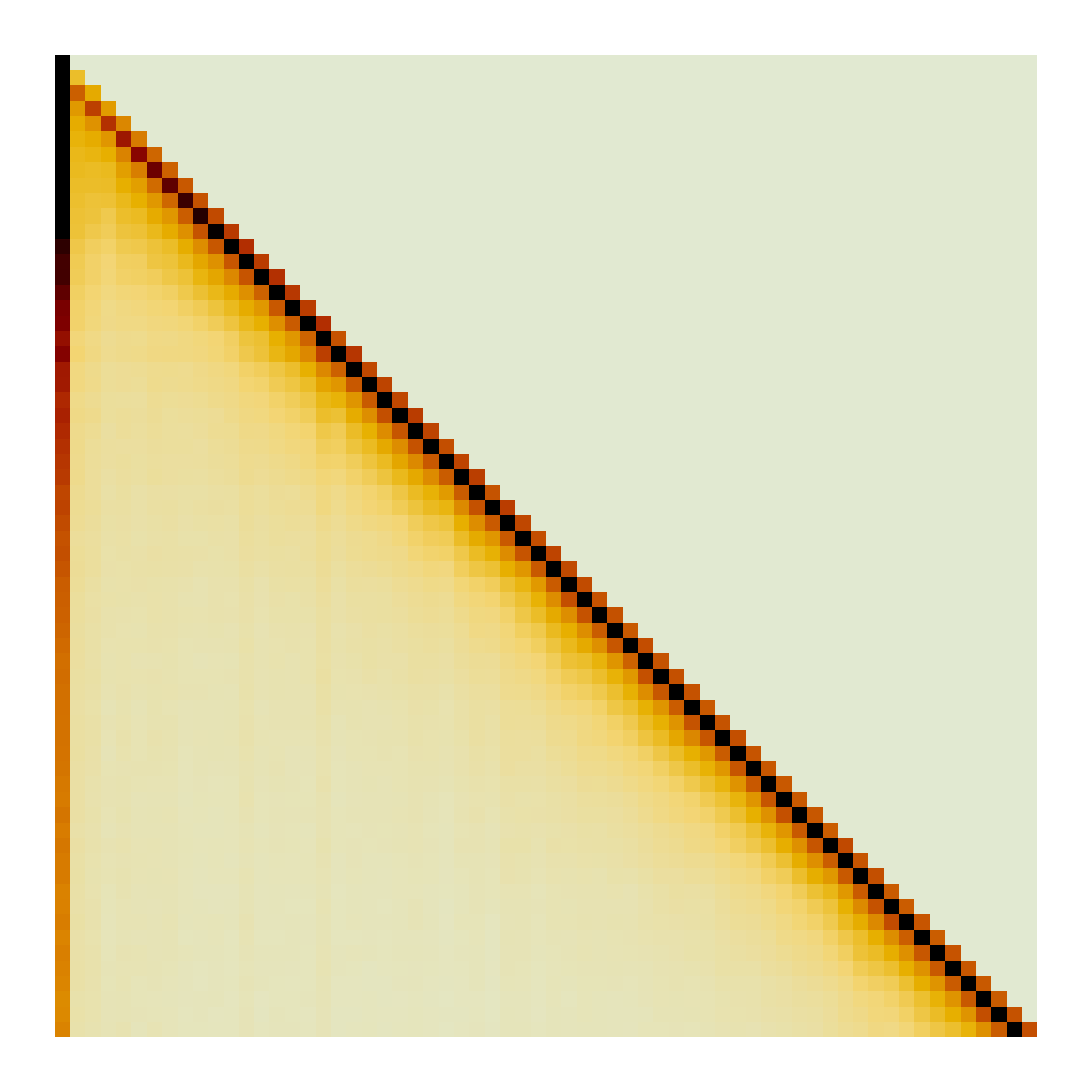} &
    \includegraphics[width=0.16\textwidth]{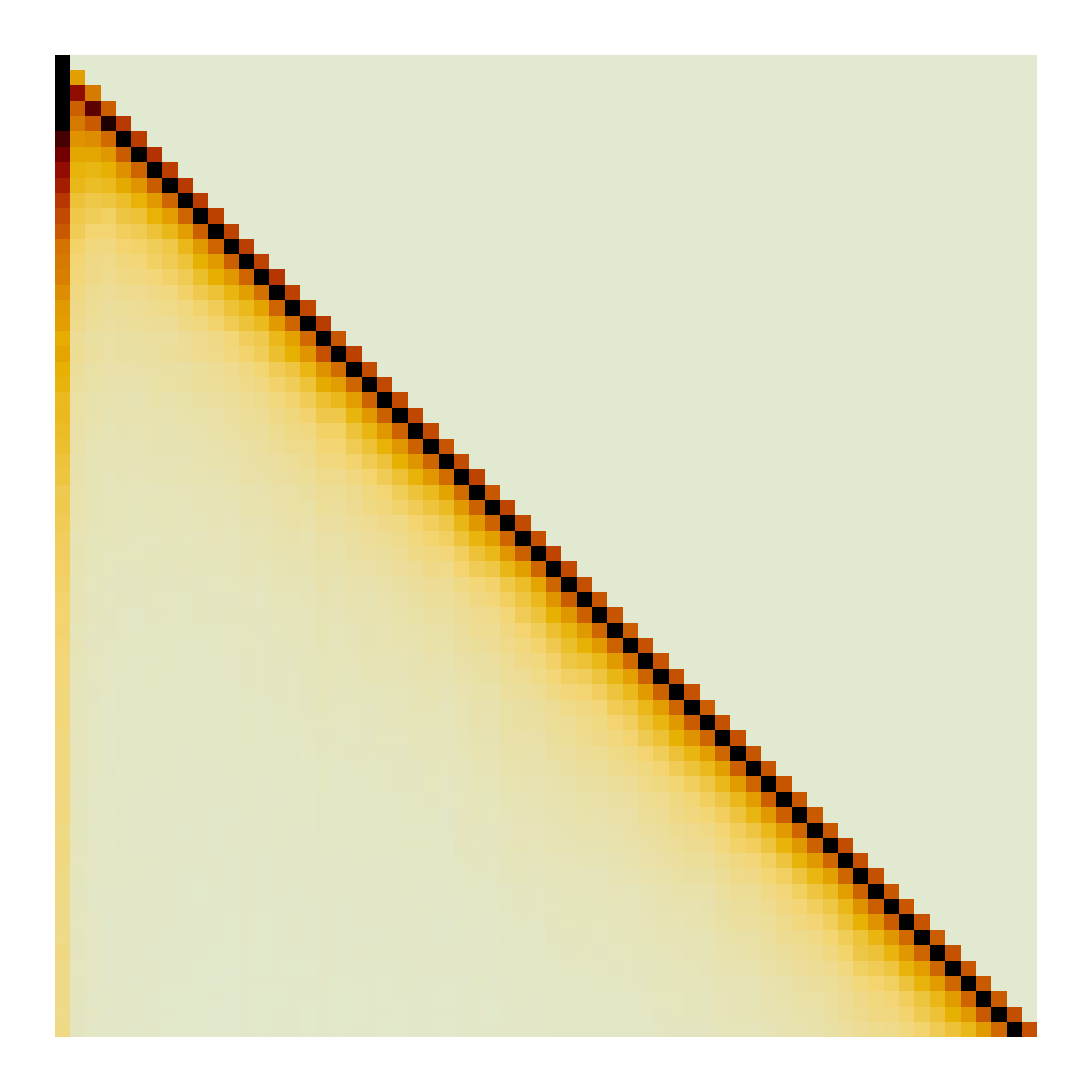} & \includegraphics[width=0.16\textwidth]{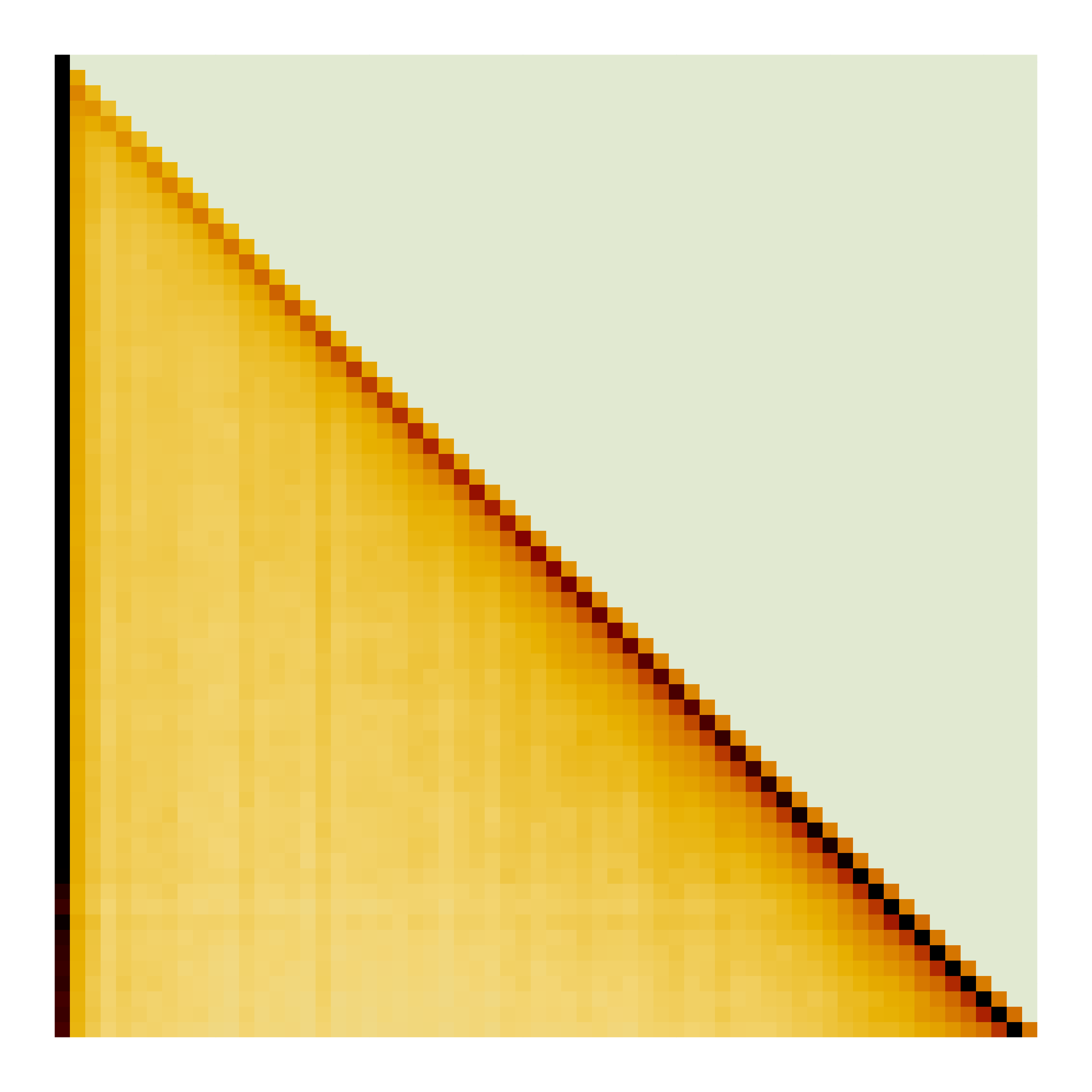}\\
    \multicolumn{4}{c}{Attention Layer 8}      
 \\
\includegraphics[width=0.16\textwidth]{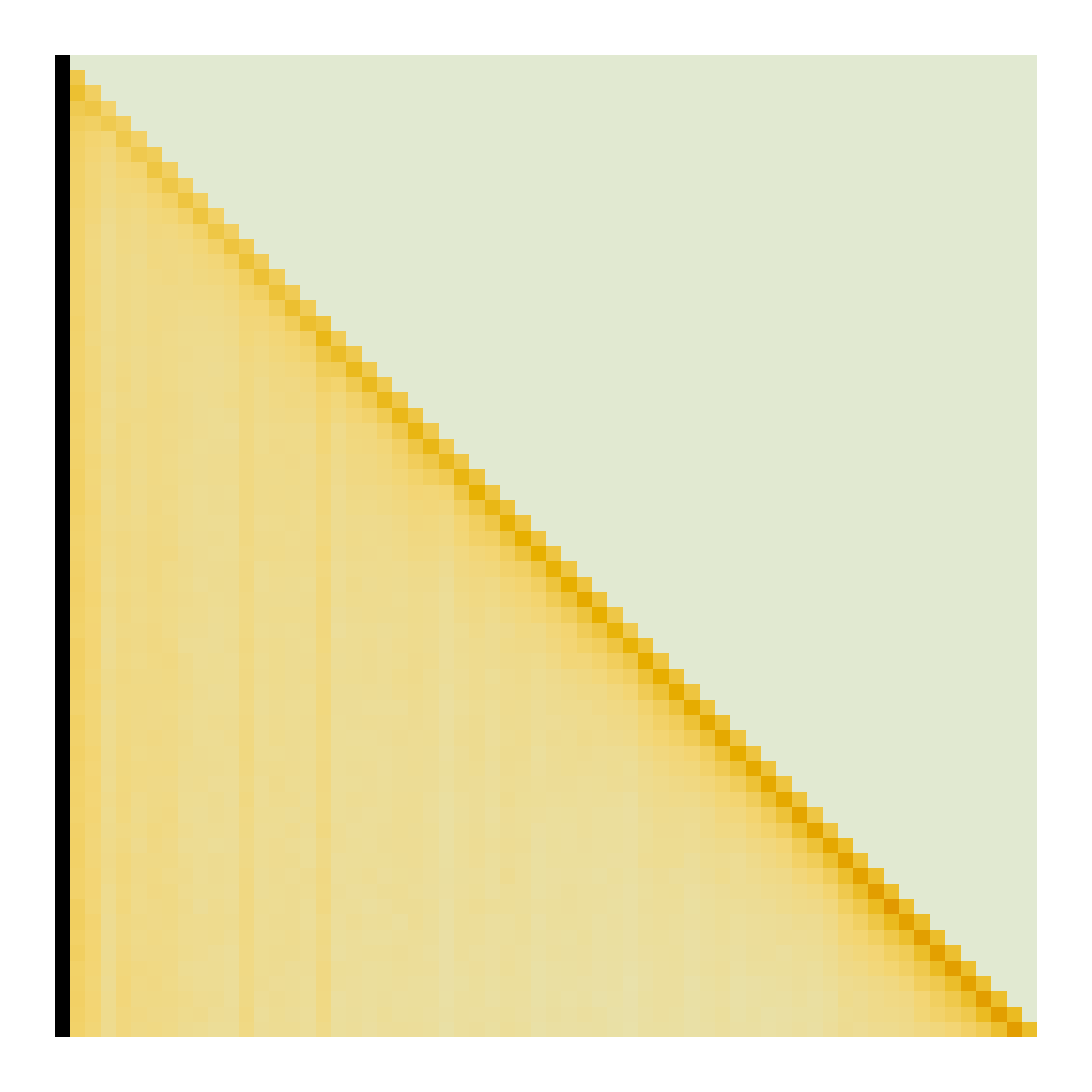} & \includegraphics[width=0.16\textwidth]{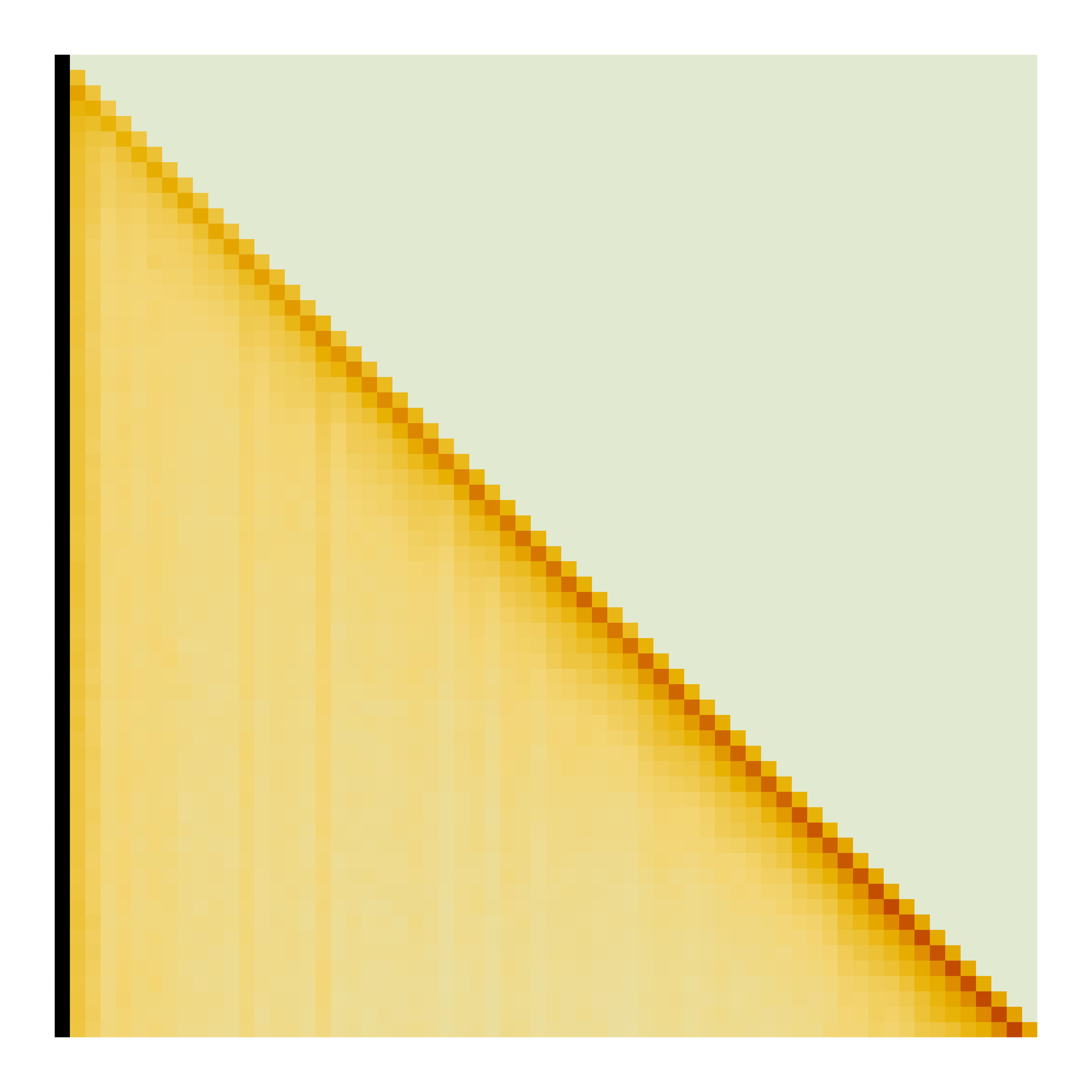} &
    \includegraphics[width=0.16\textwidth]{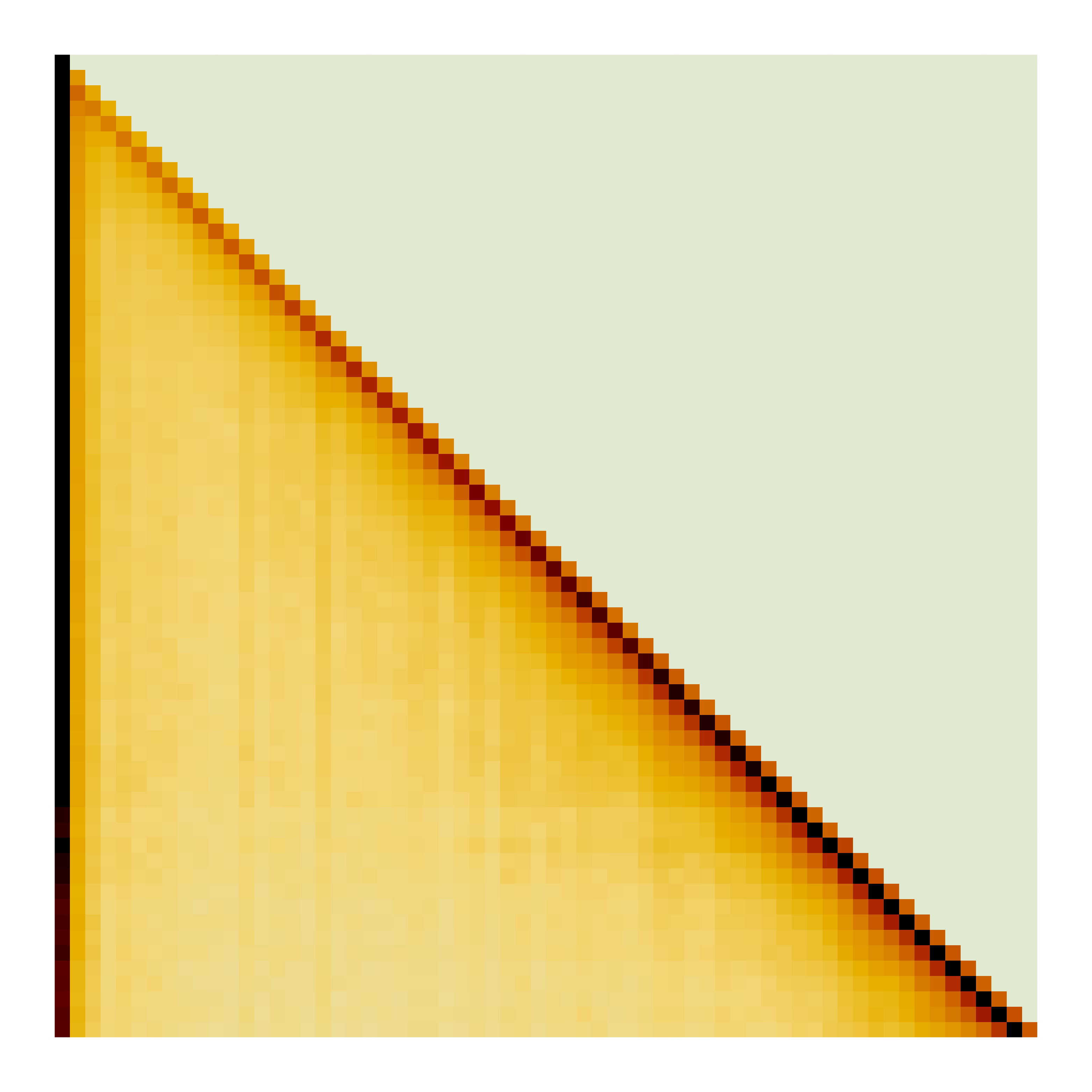} & \includegraphics[width=0.16\textwidth]{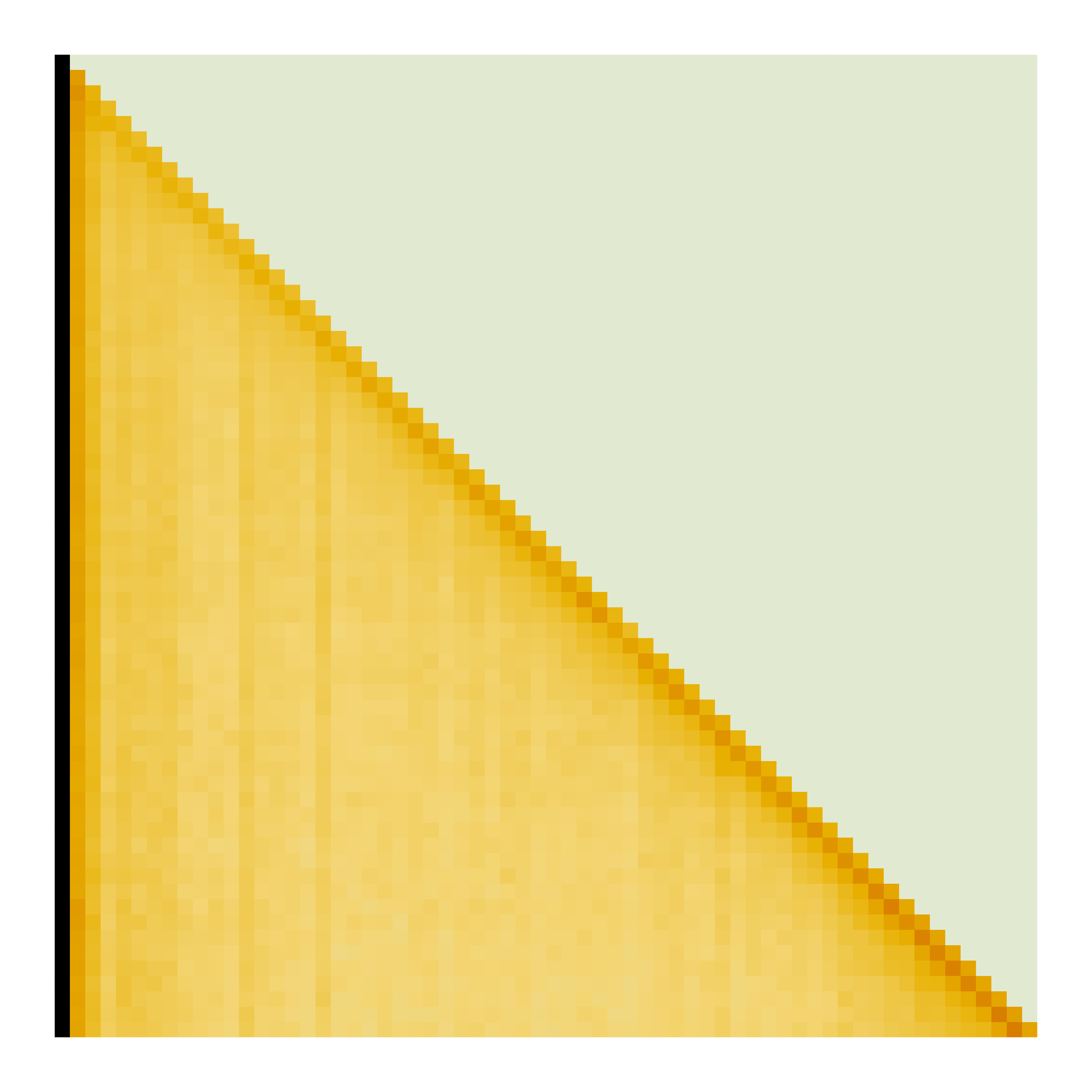}\\
    \multicolumn{4}{c}{Attention Layer 10} 
 \\
  \includegraphics[width=0.16\textwidth]{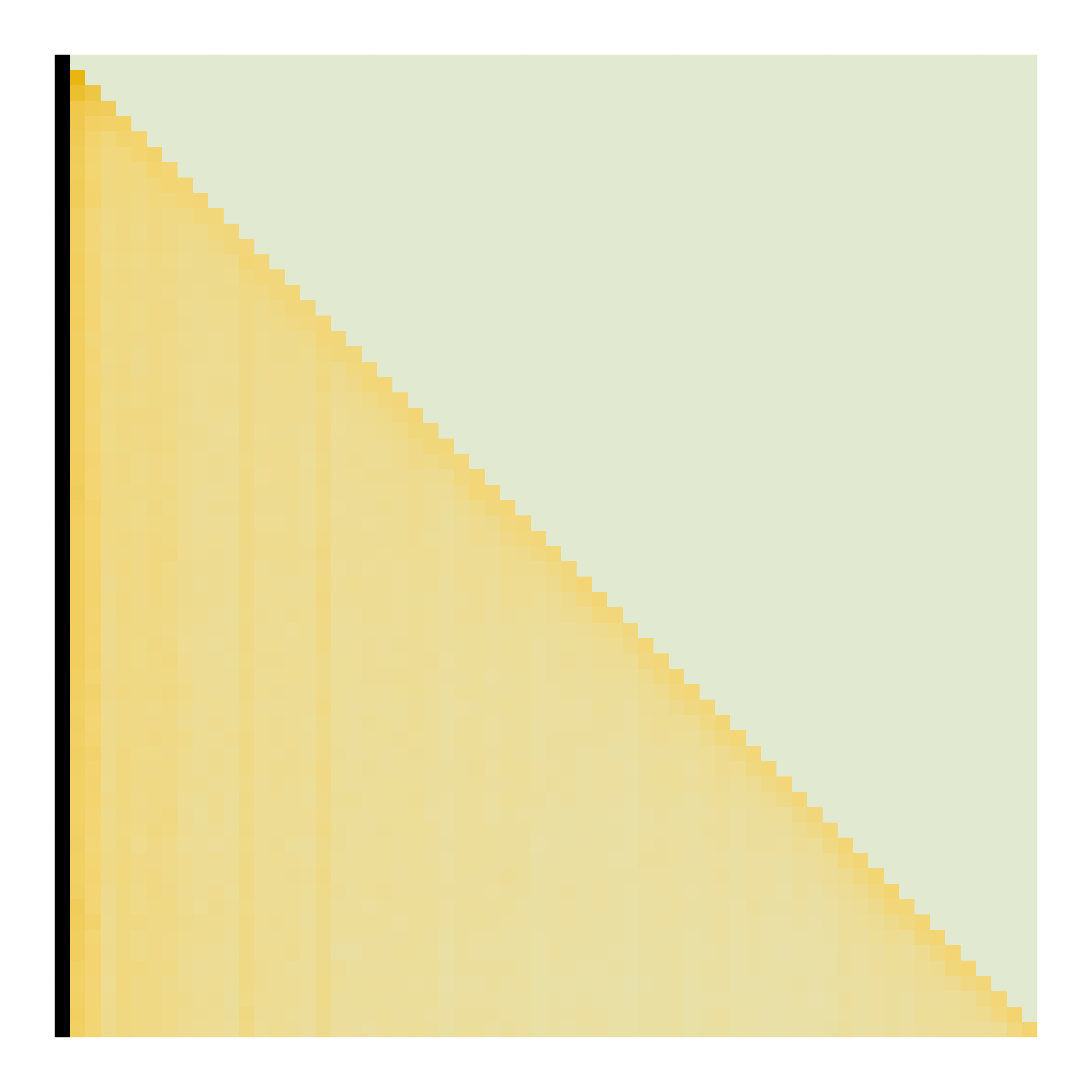} & \includegraphics[width=0.16\textwidth]{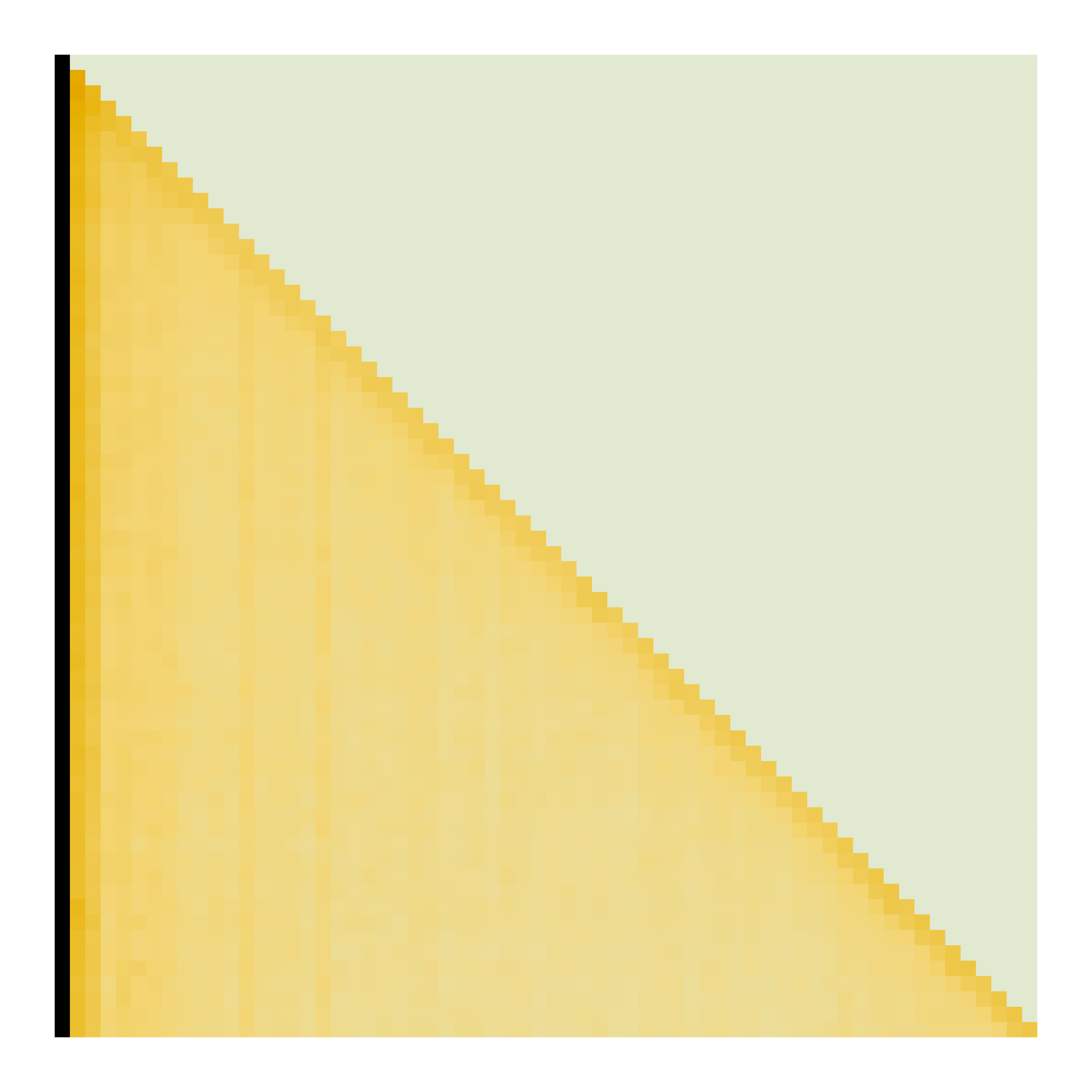} &
    \includegraphics[width=0.16\textwidth]{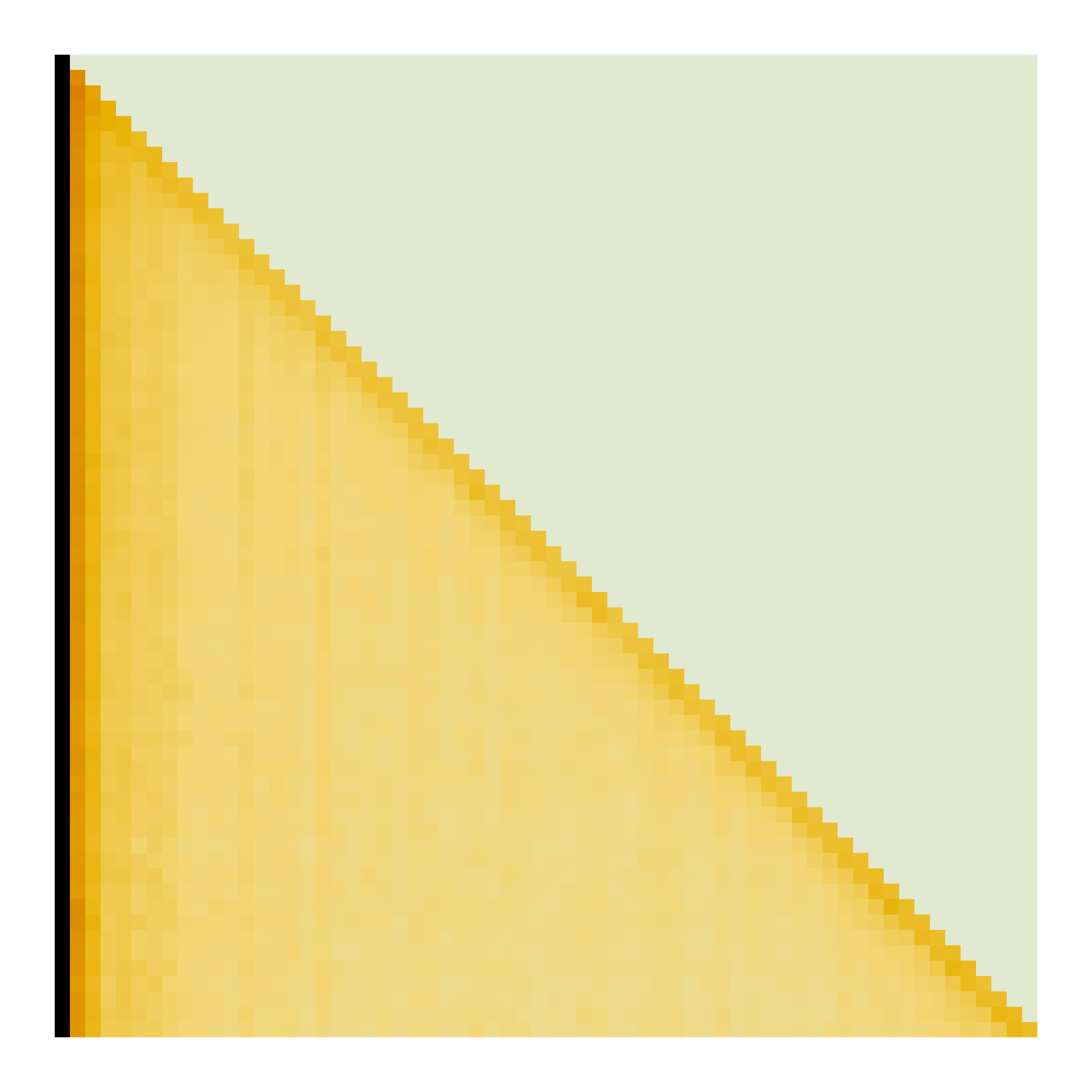} & \includegraphics[width=0.16\textwidth]{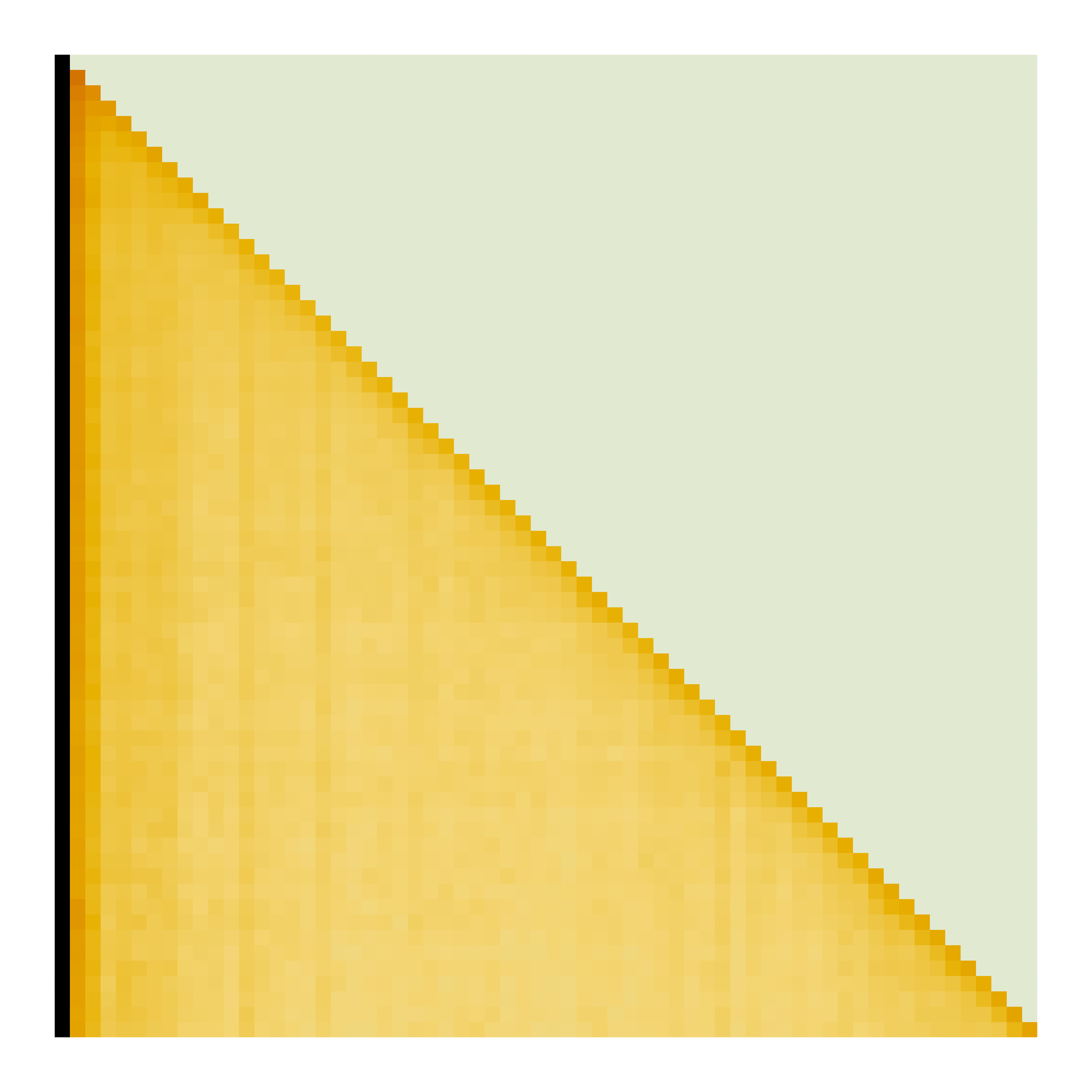}\\
    \multicolumn{4}{c}{Attention Layer 12} 
 \\
 
 \end{tabular}
 \caption{\textbf{Visualisation of polynomial average attention matrices:} Models with $P=4$ (first column) generate more local attention matrices, with reduced mass near the diagonal compared to models with $P=8$ or $P=12$, particularly in layers 4-10. In all models, the final layers (rows at the bottom) display more global attention patterns than the middle layers.}\label{fig:AttnMats}
\end{figure*}

\newpage
\begin{figure*}[h]
\centering
\begin{tabular}{cccc} 
  $p=4$ & $p=8$ & $p=12$ & $ \textbf{Softmax}$ \\
  \cmidrule(lr){1-4}
  %\midrule
    \includegraphics[width=0.16\textwidth]{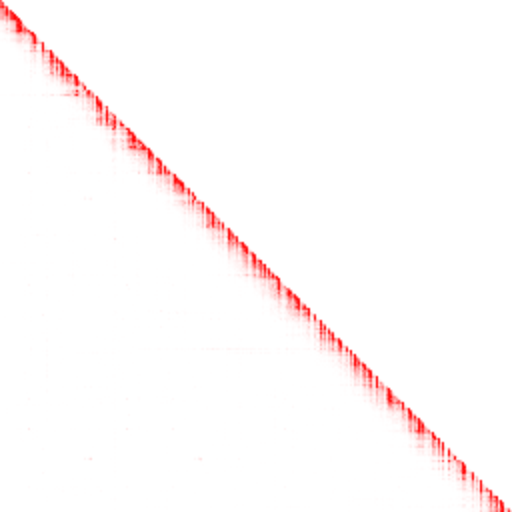} & \includegraphics[width=0.16\textwidth]{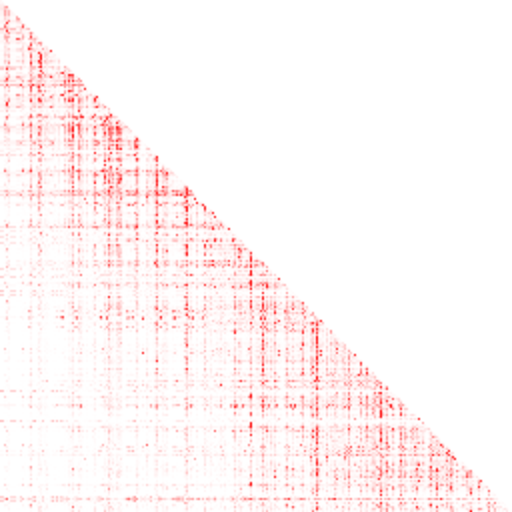} &
    \includegraphics[width=0.16\textwidth]{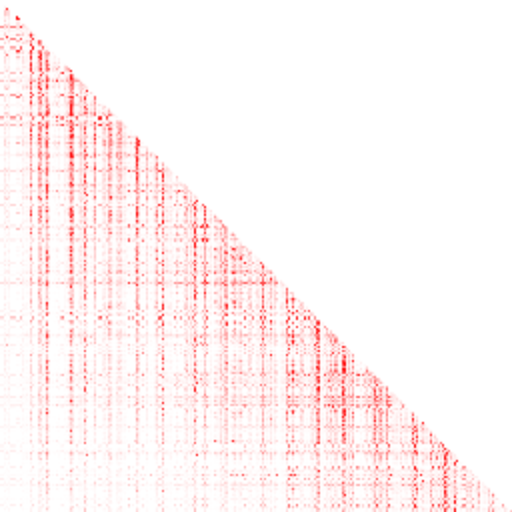} & \includegraphics[width=0.16\textwidth]{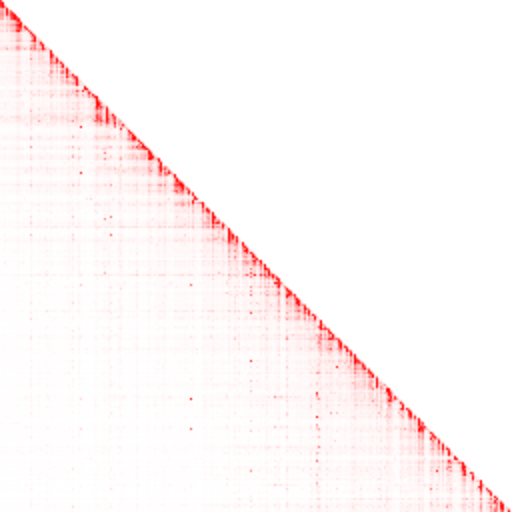}\\
    \multicolumn{4}{c}{Attention Layer 2} 
    \\   
     
    \includegraphics[width=0.16\textwidth]{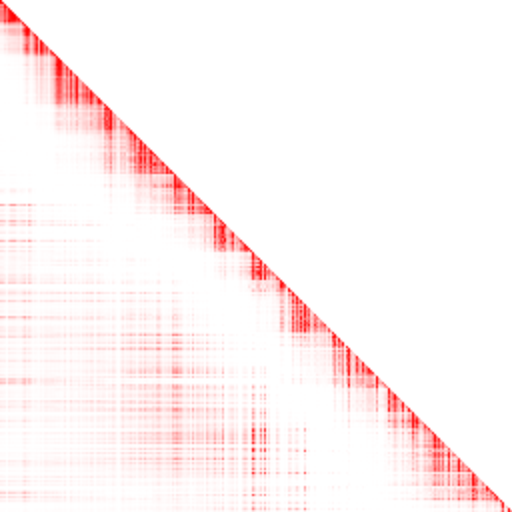} & \includegraphics[width=0.16\textwidth]{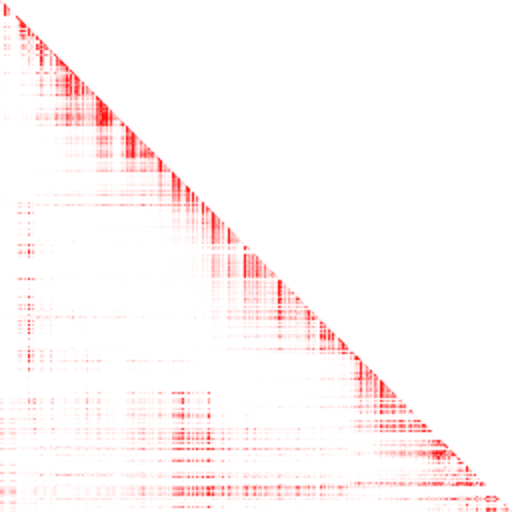} &
    \includegraphics[width=0.16\textwidth]{graphs/mech_interp/attn_mat_samples/P8/image_layer_2_andbatch_idx_0.png} & \includegraphics[width=0.16\textwidth]{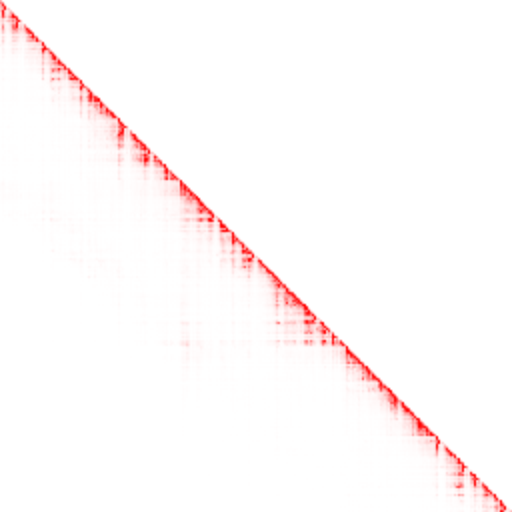}\\
    \multicolumn{4}{c}{Attention Layer 4}
 \\
    \includegraphics[width=0.16\textwidth]{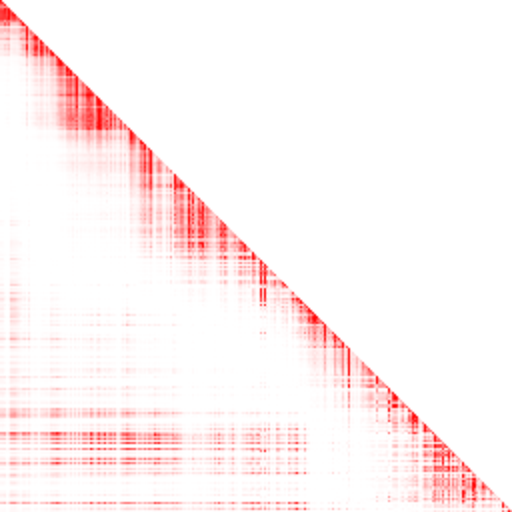} & \includegraphics[width=0.16\textwidth]{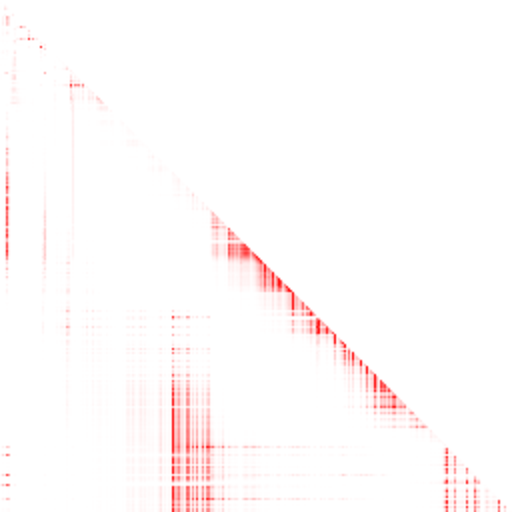} &
    \includegraphics[width=0.16\textwidth]{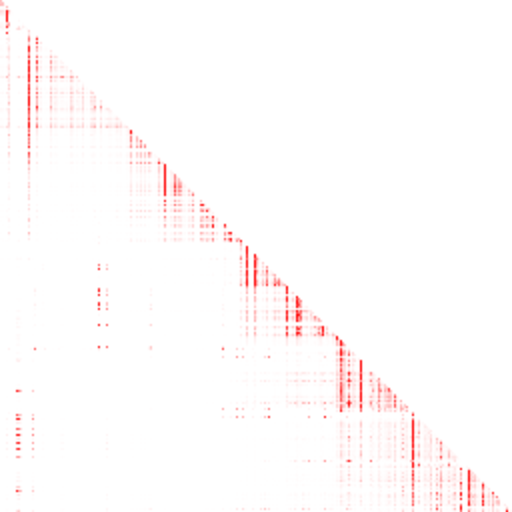} & \includegraphics[width=0.16\textwidth]{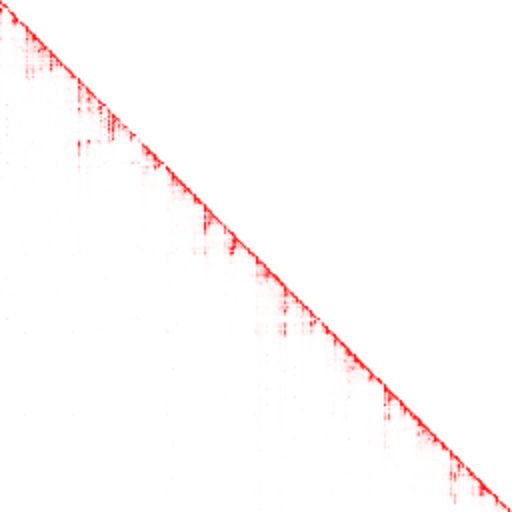}\\
    \multicolumn{4}{c}{Attention Layer 6}    
 \\
       \includegraphics[width=0.16\textwidth]{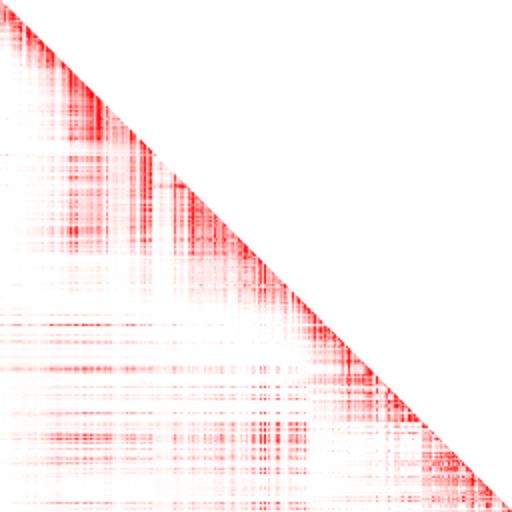} & \includegraphics[width=0.16\textwidth]{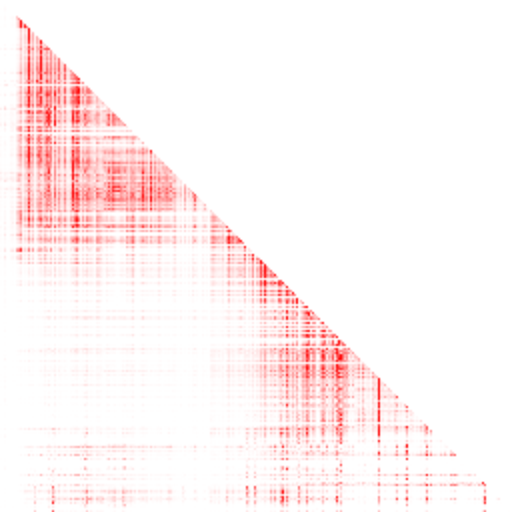} &
    \includegraphics[width=0.16\textwidth]{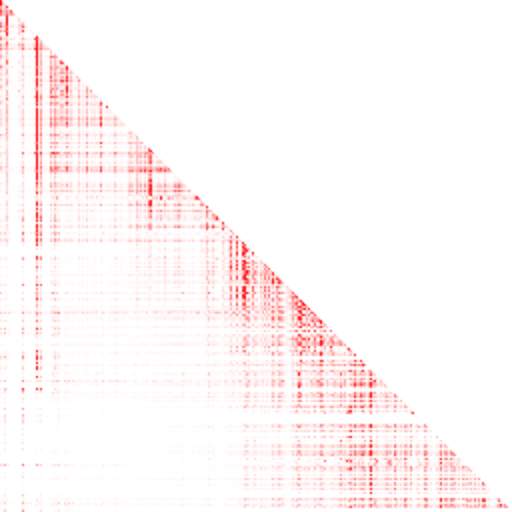} & \includegraphics[width=0.16\textwidth]{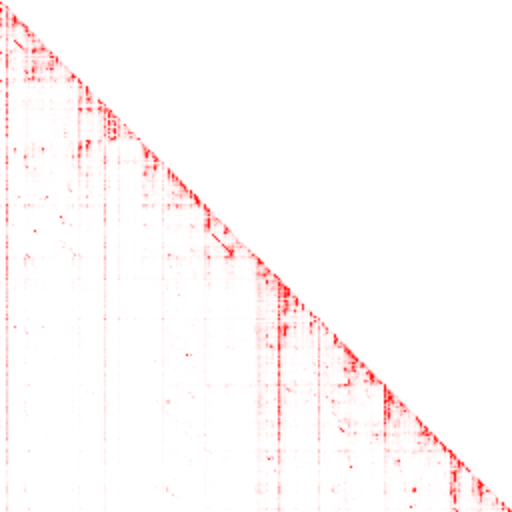}\\
    \multicolumn{4}{c}{Attention Layer 8}      
 \\
      \includegraphics[width=0.16\textwidth]{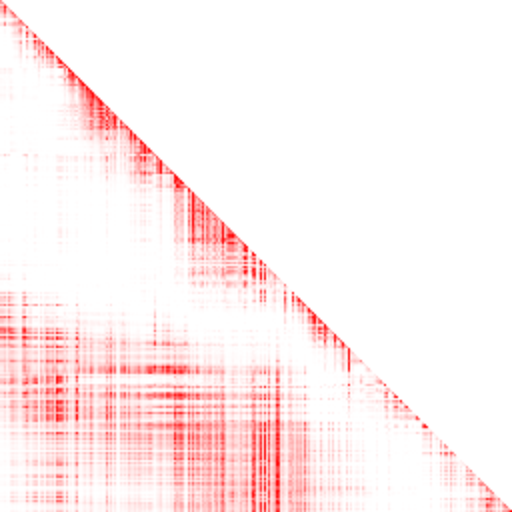} & \includegraphics[width=0.16\textwidth]{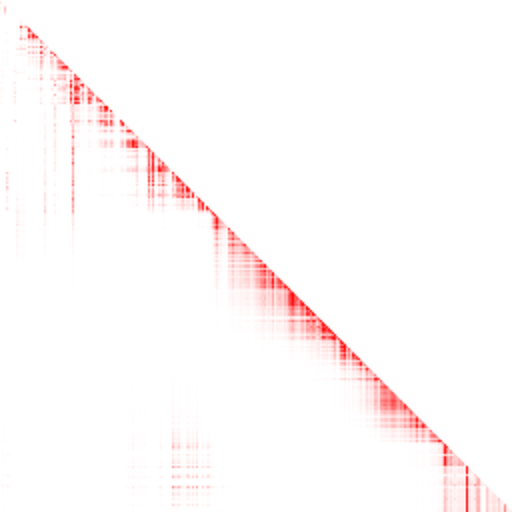} &
    \includegraphics[width=0.16\textwidth]{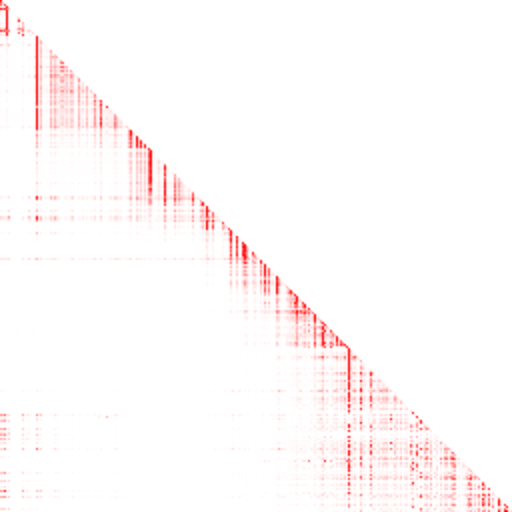} & \includegraphics[width=0.16\textwidth]{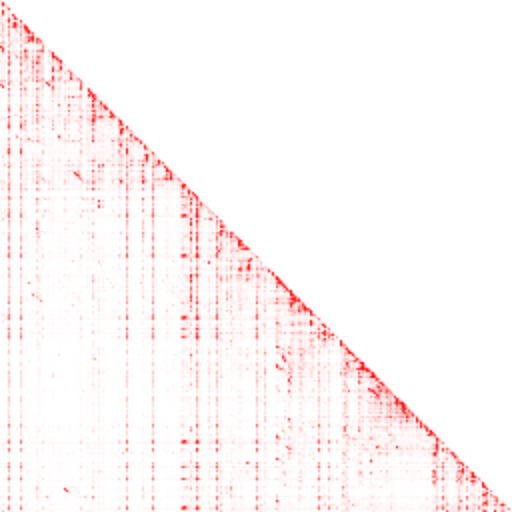}\\
    \multicolumn{4}{c}{Attention Layer 10} 
 \\
    \includegraphics[width=0.16\textwidth]{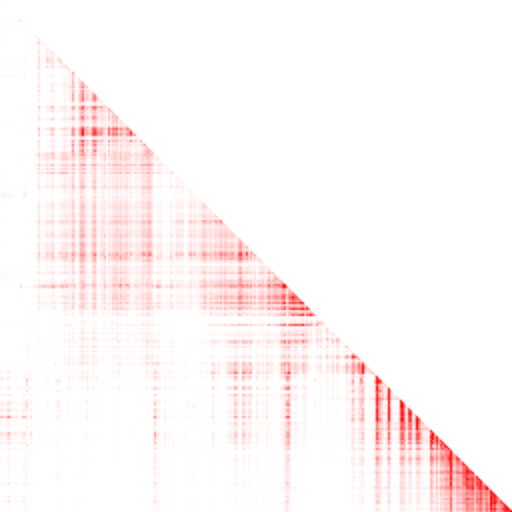} & \includegraphics[width=0.16\textwidth]{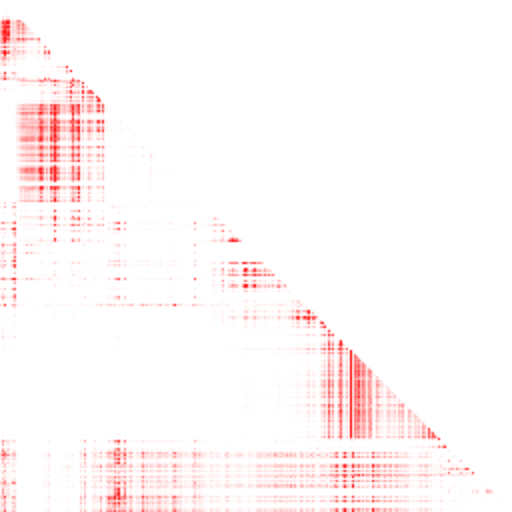} &
    \includegraphics[width=0.16\textwidth]{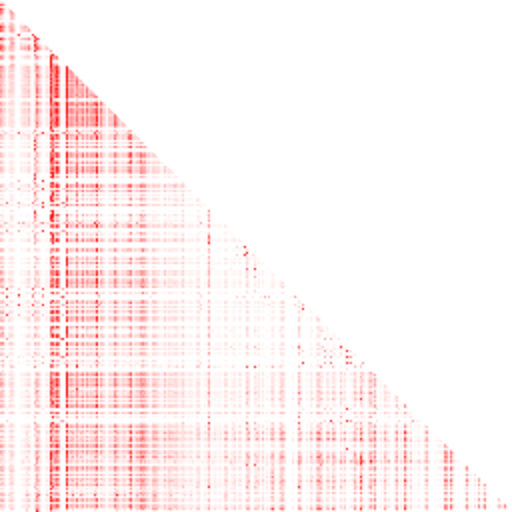} & \includegraphics[width=0.16\textwidth]{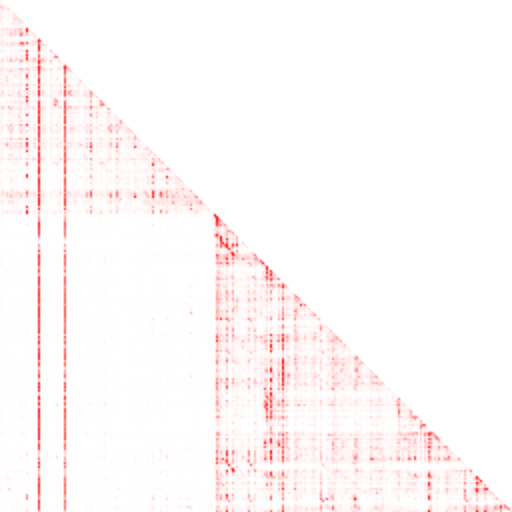}\\
    \multicolumn{4}{c}{Attention Layer 12} 
 \\
 \end{tabular}
 \caption{\textbf{Visualisation of random samples of polynomial attention matrices:} Although the attention matrices are noisy and a small number of samples may not capture the full distribution trend, the \PowerSoftmax-based models (first three columns) show behavior similar to the original \Softmax (last column). Notably, our attention layers can dynamically adjust focus across different parts of the input, allowing attention heads to freely learn both local and global patterns.} \label{fig:AttnMatsSamples}
\end{figure*}

\newpage
\clearpage

%%%%%%%%%%%%%%%%%%%%%%%%%%%%%%%%%%%%%%%%%%%%%%%%%%%%%%%%%%%%

\end{document}